\documentclass[acmsmall]{acmart}

\AtBeginDocument{%
  }

\setcopyright{acmlicensed}
\copyrightyear{2018}
\acmYear{2018}
\acmDOI{XXXXXXX.XXXXXXX}

\acmJournal{JACM}
\acmVolume{37}
\acmNumber{4}
\acmArticle{1}
\acmMonth{8}

\usepackage{algorithm}
\usepackage{algorithmic}
\usepackage{subfigure}
\usepackage{multirow}
\usepackage{multicol}
\usepackage{bm}
\usepackage{amsmath}
\usepackage{bm}
\usepackage{diagbox}
\usepackage{booktabs}

\begin{document}

\title{Bayesian Inverse Transfer in Evolutionary Multiobjective Optimization}

\thanks{This research is supported partially by Distributed Smart Value Chain programme which is funded under the Singapore RIE2025 Manufacturing, Trade and Connectivity (MTC) Industry Alignment Fund-Pre-Positioning (Award No: M23L4a0001); partially by the Ramanujan Fellowship from the Science and Engineering Research Board, Government of India (Grant No. RJF/2022/000115 ); partially by the Center for Frontier AI research center at the Institute of High Performance Computing, A*STAR.}
\author{Jiao Liu}
\affiliation{%
	\institution{College of Computing \& Data Science, Nanyang Technological University}
	\streetaddress{50 Nanyang Ave}
	\city{Singapore}
	\country{Singapore}}
\email{jiao.liu@ntu.edu.sg}

\author{Abhishek Gupta}
\authornote{*Corresponding author}
\affiliation{%
	\institution{School of Mechanical Sciences, Indian Institute of Technology Goa}
	\streetaddress{Goa Engineering College Campus, Farmagudi, Ponda}
	\city{Ponda}
	\country{India}}
\email{abhishekgupta@iitgoa.ac.in}

\author{Yew-Soon Ong}
\affiliation{%
	\institution{College of Computing \& Data Science, Nanyang Technological University}
	\streetaddress{50 Nanyang Ave}
	\city{Singapore}
	\country{Singapore}}
\email{asyong@ntu.edu.sg}


\begin{abstract}
  Transfer optimization enables data-efficient optimization of a target task by leveraging experiential priors from related source tasks. This is especially useful in multiobjective optimization settings where a set of trade-off solutions is sought under tight evaluation budgets. In this paper, we introduce a novel concept of \textit{inverse transfer} in multiobjective optimization. Inverse transfer stands out by employing Bayesian inverse Gaussian process models to map performance vectors in the objective space to population search distributions in task-specific decision space, facilitating knowledge transfer through \emph{objective space unification}. Building upon this idea, we introduce the first \textit{Inverse Transfer Evolutionary Multiobjective Optimizer} (invTrEMO). A key highlight of invTrEMO is its ability to harness the common objective functions prevalent in many application areas, even when decision spaces do not precisely align between tasks. This allows invTrEMO to uniquely and effectively utilize information from heterogeneous source tasks as well. Furthermore, invTrEMO yields high-precision inverse models as a significant byproduct, enabling the generation of tailored solutions on-demand based on user preferences. Empirical studies on multi- and many-objective benchmark problems, as well as a practical case study, showcase the faster convergence rate and modelling accuracy of the invTrEMO relative to state-of-the-art evolutionary and Bayesian optimization algorithms. The source code of the invTrEMO is made available at https://github.com/LiuJ-2023/invTrEMO.
\end{abstract}




\maketitle

\section{Introduction}
A plethora of optimization problems in the real-world, from engineering design~\cite{osyczka1978approach} and manufacturing operations~\cite{li2018efficient} to finance~\cite{wang2022multi} or even scientific discovery~\cite{macleod2022self}, often seek to minimize/maximize a set of conflicting objectives. These are known as multiobjective optimization problems (MOPs)~\cite{branke2008multiobjective}. Typically, there is no single solution that minimizes/maximizes all of the objectives in an MOP~\cite{deb2011multi, ehrgott2005multicriteria}. Therefore, the goal is to find a representative \emph{set} of optimized solutions---referred to as nondominated or Pareto optimal solutions---that map to the Pareto front (PF) of best achievable performance trade-offs in the objective space. Evolutionary algorithms (EAs), by virtue of the implicit parallelism of their population-based search strategy, have emerged as a method of choice for converging toward such a set of solutions in a single optimization run~\cite{coello2007evolutionary}. Mechanisms for maintaining population diversity in an EA facilitate the discovery of solutions that are well-distributed along the PF, thereby allowing a decision-maker (DM) to visualize the performance trade-offs and select
the most preferred solution(s) \textit{a posteriori}~\cite{9005525, deb2023learning}.

Despite many success stories of EAs for MOPs, these algorithms are typically data-hungry, consuming tens of thousands of function evaluations to approach the Pareto optimal solution set. The emerging topic of \textit{transfer evolutionary optimization}~\cite{gupta2017insights} is tailored to address this issue. It builds on the observation that real-world problems seldom exist in isolation, so that experiential priors in the form of data or learnt models from previously solved tasks (also known as \textit{source tasks}) could be reused for data-efficient optimization of a related \textit{target task}. The majority of existing transfer optimization algorithms can be broadly grouped into two categories. The first relies on knowledge contained in the set of high-quality solutions evolved in an EA, as demonstrated in works such as~\cite{louis2004learning,feng2017autoencoding, scott2023first}. The second category utilizes learnt models---in the form of probability distributions of evolved populations \cite{shakeri2022scalable} or discriminative surrogate models trained to map points from the search to the objective space \cite{wistuba2018scalable, 8207638}---as building-blocks of knowledge that can be transferred from the source to the target. In the latter, the transfer is usually performed by aggregating the models drawn from the respective tasks, made readily possible in homogeneous transfer settings where decision spaces of the tasks are the same. In situations of source-target heterogeneity, additional domain adaptation steps are needed to unify their search/decision spaces before knowledge transfer can be initiated~\cite{min2020generalizing,lim2021non}. This adaptation may be carried out through linear \cite{bali2017linearized, xue2020affine} or nonlinear \cite{lim2021solution} solution transformations schemes, raising the non-trivial question of ascertaining a scheme that offers the most favourable balance between computational complexity and adaptation proficiency for the given problem at hand.

In this paper, we introduce a new concept of \emph{inverse transfer multiobjective optimization} through objective space unification. In conjunction, a first \textit{Inverse Transfer Evolutionary Multiobjective Optimizer} (invTrEMO) is proposed {for MOPs with expensive black-box objectives}. The algorithm centres around the construction of Bayesian inverse Gaussian process (GP) models that map performance vectors from the objective space to search distributions in task-specific decision space \cite{cheng2015multiobjective,lin2022pareto}, from which populations of candidate solutions are sampled. The model is iteratively and preferentially updated during the course of a target optimization run with the nondominated solutions found so far, thereby increasing the probability of sampling a higher-quality population in subsequent iterations. \emph{The invTrEMO stands out from other transfer optimizers by conducting knowledge transfers through its inverse model}. {A key motivation behind this idea is to address the challenge of heterogeneous decision spaces. In practice, it is possible that decision spaces may not always align across different optimization tasks~\cite{fan2023transfer}. For instance, datasets may originate from experiments exploring various robot skills, each characterized by distinct, potentially overlapping sets of control parameters~\cite{wang2021learning}. Likewise, in the realm of machine learning model tuning, data sourced from a commercial Bayesian optimization tool~\cite{golovin2017google,balandat2020botorch} may exhibit variability due to differences in the hyperparameters tuned by different users and the varied terminologies employed. Notwithstanding such differences in decision space, it is worth noting that the objective functions of interest often coincide in MOPs within the same application domain~\cite{tan2023pareto}.} Examples abound in science and engineering design applications, among others, where although design variables may vary from one task to the next (e.g., new variables/features are added to a design and obsolete ones removed), objective functions such as overall performance efficiency, cost, robustness or design safety, recur across tasks. {Building upon this observation, the invTrEMO treats this \textit{common objective space} as a unifying bridge, enabling a connection between heterogeneous source-target task pairs as well.} In our implementation, this connection is established through the use of principled \emph{inverse Transfer GP} (invTGP) models.

The salient features of the invTrEMO bring distinctive advantages over existing transfer optimization algorithms. First, even in the absence of aligned decision spaces, invTrEMO is still able to exploit transferrable information that may reside in common objective spaces prevalent in many application areas. Second, conducting knowledge transfer in the inverse mode leads to the creation of high-precision inverse models as a significant byproduct of the invTrEMO. These models can play a pivotal role in generating Pareto optimal solutions on-demand that comply with user preferences, thus supporting multi-objective decision-making processes.

The main contributions and experimental analyses of this paper are summarized below.
\begin{itemize}    
	\item We present a novel \textit{inverse transfer} paradigm in multiobjective optimization that melds probabilistic inverse modelling (via invTGPs in our implementation) with evolutionary variation and selection. A key highlight of the paradigm is its ability to facilitate knowledge transfers even across heterogeneous source-target task pairs (with non-identical decision spaces), by harnessing the unification provided by common objective spaces. {To the best of our knowledge, this is the first technology to achieve heterogeneous transfer in the context of MOPs with expensive black-box objectives.}
	
	\item The probabilistic inverse models derived from invTrEMO enable on-demand generation of Pareto optimal solutions with preferred performance trade-offs, in response to simple DM queries. Incorporating source data significantly enhances the accuracy of the inverse models, promising to augment future multiobjective decision-making processes.
	
	\item The effectiveness of our approach is assessed using multi- and many-objective benchmarks from the DTLZ suite~\cite{deb2005scalable}, the DTLZ$^{-1}$ suite~\cite{ishibuchi2016performance}, and practical vehicle crashworthiness design examples~\cite{liao2008multiobjective}. To demonstrate data-efficiency gains, we consider optimization under limited evaluation budgets. Comparisons against state-of-the-art evolutionary and Bayesian optimization algorithms, that are specifically crafted for problems with expensive black-box objectives, confirm invTrEMO's capacity to boost convergence rates and produce high-precision inverse models in the process.
\end{itemize}
The remainder of the paper is organized as follows. Section II presents related work from the literature and an overview of important background concepts. Section III introduces the invTrEMO. The experimental studies are presented in Section IV. Section V concludes the paper.

\section{Preliminaries}

This section provides an overview of related work and essential background concepts in multiobjective optimization, decomposition-based optimization methods, GP modeling and inverse machine learning in MOPs.

\subsection{Related Work}
Comprehensive reviews of the topic of transfer optimization are available in the literature \cite{tan2021evolutionary, osaba2022evolutionary}. Hence, this subsection offers a brief overview of the subject. We begin with some of the latest developments in transfer optimization where the goal is to boost convergence rates on a target task by leveraging prior optimization experiences~\cite{gupta2017insights, tan2021evolutionary}. Emphasis is placed on transfer algorithms for expensive problems, as this is a primary use-case for the invTrEMO implemented in this work. It is worth stating that unlike conventional transfer learning which only focuses on enhancing predictive accuracy~\cite{niu2020decade}, transfer optimization bridges the gap between machine learning for predictive modeling and optimization for prescriptive decision-support.

\textit{Population-centric transfer optimization} with EAs has been extensively studied lately. This methodology includes two widely employed knowledge aggregation strategies: \textit{genetic transfer}~\cite{louis2004learning} and \textit{model-based transfer}~\cite{da2018curbing}. Genetic transfer involves the injection of elite solutions from source tasks into the search space of the target task~\cite{salih2022promoting, scott2023first}, whereas examples of the latter have made adaptive use of probability distribution models of elite solutions from the source to sample high-quality offspring in the target~\cite{shakeri2022scalable, lim2021non}. \textit{Operator-based transfer} is yet another promising strategy \cite{friess2019learning,friess2020improving, friess2020representing}, but one that to our knowledge has received relatively less research attention to date. While significant performance gains through transfer have been reported in the literature, with a variety of applications in combinatorial optimization problems \cite{ardeh2022knowledge, ardeh2021genetic, singh2022study}, dynamic optimization \cite{guo2022knowledge}, evolutionary machine learning \cite{lin2023amtea,jiao2022benefiting}, engineering process design \cite{yao2023piggybacking}, to name just a few, many of these methods still call for a significant computational budget, as they entail the evaluation of populations of candidate solutions in every iteration of an optimization run. Consequently, they may not be the ideal choice for addressing problems involving costly black-box objective functions.

In expensive problems where evaluating a solution requires time-consuming computer simulations or costly real-world procedures, transfer via surrogate models is potentially a more viable approach \cite{8207638, li2023data, li2023evolutionary}. The basic idea is to harness  computationally efficient discriminative models, learnt from data acquired in both the source and target tasks, as substitutes to guide the optimization process~\cite{swersky2013multi}. The majority of such methods aim to speedup convergence by enhancing the predictive accuracy of the discriminative model through knowledge transfer~\cite{8207638,bai2023transfer,gupta2022tightening}. Optimization performance gains have been achieved via different variations of this general strategy~\cite{tighineanu2022transfer, luo2022expensive,fan2020surrogate}. However, it is noteworthy that these approaches typically presume that the decision spaces of the source and target tasks share the same features and dimensions, which limits their applicability in handling heterogeneous source-target pairs whose decision spaces may be non-identical. Although Min \textit{et al.}~\cite{min2020generalizing} introduced an algorithm for heterogeneous transfer in single-objective optimization regimes, the applicability of associated methods to MOPs---where the challenge is to discover a set of Pareto optimal solutions that spread across the PF---remains largely unexplored.

The goal of multiobjective optimization is to simultaneously converge to and diversify along the PF.  For enhancing PF coverage, one promising avenue is to build machine learning models that learn to map points from the PF to the decision space, either in a post-hoc manner \cite{suresh2023machine} or by online integration with the multiobjective optimization algorithm~\cite{cheng2015multiobjective}. Online integration has yielded promising outcomes lately, improving convergence speeds and increasing the density of the PF approximation~\cite{lin2022pareto}. However, the training of the models is still restricted to the data derived from the target task alone, with no mechanism to take advantage of related source tasks. The potential of inverse transfer in MOPs was only recently considered by Tan \textit{et al}.~\cite{tan2023pareto} in a post-hoc setup to improve the predictive accuracy of the learnt model, but without explicitly prioritizing transfer optimization as a means to achieve faster convergence.

This paper is thus the first to present \emph{inverse transfer} for data-efficient multiobjective optimization. The uniqueness of our proposal lies in enabling effective utilization of data even from heterogeneous source MOPs, under the practical observation of problems with common objective spaces. The proposed invTrEMO not only accelerates convergence rates, but also helps DMs efficiently arrive at nondominated solutions that comply with their preferences.

\subsection{Background Concepts}

\subsubsection{Multiobjective Optimization}
An unconstrained MOP can be formulated, without loss of generality, as:
\begin{equation}\label{Eq:MOP}
\begin{aligned}
\min \ & \textbf{f}(\textbf{x}) = [f_1(\textbf{x}), f_2(\textbf{x}), \ldots,f_m(\textbf{x})] \\
\text{s.t.}\ & \textbf{x} \in \mathcal{X} \in \mathbb{R}^d \\
\end{aligned}
\end{equation}
where $f_i(\textbf{x}),(i \in \{1,\ldots,m\})$ is the $i$th objective function, $m$ is the number of objectives, $\textbf{x}$ represents a candidate solution (decision vector), $\mathcal{X} = \{\textbf{x} = (x_1,\ldots,x_d)|L_j \leq x_{j} \leq U_j, \ j = 1,\ldots,d\}$ is the decision space of all possible solutions, $d$ is the dimensionality of the decision space/vector, and $L_j$ and $U_j$ are the lower and upper bounds of the $j$th decision variable. 

Given the formulation in (\ref{Eq:MOP}), we revisit definitions of some basic concepts in multiobjective optimization below.
\begin{definition}[\textbf{Pareto Dominance}]
	Solution $\textbf{x}_a$ {Pareto dominates} $\textbf{x}_b$, if $\forall i \in \{1,2,\ldots,m\}$, $f_{i}(\textbf{x}_a) \leq f_{i}(\textbf{x}_b)$ and $\exists i' \in \{1,2,\ldots,m\}$ such that $f_{i'}(\textbf{x}_a) < f_{i'}(\textbf{x}_b)$.
\end{definition}	
\begin{definition}[\textbf{Pareto Optimal Solution}]
	Solution $\textbf{x}_a$ is said to be {Pareto optimal} if there exists no other solution in $\mathcal{X}$ that Pareto dominates $\textbf{x}_a$.
\end{definition}

\begin{definition}[\textbf{Pareto Set}]
	The set of all Pareto optimal solutions constitutes the Pareto set (PS) in decision space.
\end{definition}

\begin{definition}[\textbf{Pareto Front}]
	The image of the Pareto set in the objective space is referred to as the Pareto front (PF).
\end{definition}

\subsubsection{Decomposition-based Multiobjective Optimization}\label{sec:mop_decomposition} A widely-adopted technique for MOPs, decomposition-based methods involve scalarizing an MOP into a collection of single-objective optimization subproblems~\cite{zhang2007moea}. By solving these subproblems separately or in tandem, a finite set of nondominated solutions can be obtained to approximate the  PS and PF. Popular approaches like the weighted-sum or Tchebycheff scalarization require the specification of a set of \textit{preference vectors} denoted as $\mathcal{W} = \{\textbf{w} = (w_1,\ldots,w_m)| w_i \geq 0, \sum_{i=1}^{m} w_i=1 \}$, using which weighted combinations of the objective functions are derived~\cite{ehrgott2005multicriteria}. We opt for the augmented Tchebycheff scalarization here due to its simplicity and known compatibility with non-convex PFs. This method aggregates the $m$ objective function values, which are assumed to be normalized to (0, 1], as~\cite{lin2022pareto}:
\begin{equation}\label{eqn:techebycheff}
\begin{aligned}
f^{tch}(\textbf{x}|\textbf{w}) = \max_{i \in \{1,\ldots,m\}} \{ w_i f_i(\textbf{x}) \} + \eta \sum_{i=1}^{m} w_i f_i(\textbf{x}),
\end{aligned}
\end{equation}
where $\eta$ is a small positive value (set to 0.05 in our implementation). 

\subsubsection{Gaussian Processes}
The GP is a well-established probabilistic machine learning model frequently employed in multiobjective optimization \cite{binois2022survey, de2021greed}. In this paper, we use the GP in two modes. In a \emph{forward mode}, the GP approximates the mapping from the decision space to the augmented Tchebycheff scalarized output. We assume $f^{tch} \sim \mathcal{GP}(\mu_{gp},k_{gp}(\textbf{x},\textbf{x}'))$, where $\mu_{gp}$ denotes the mean (commonly set to a constant or zero), and $k_{gp}(\textbf{x},\textbf{x}')$ is any valid kernel that captures the similarity between the pair of outputs $f^{tch}(\textbf{x}|\textbf{w})$ and $f^{tch}(\textbf{x}'|\textbf{w})$.

Consider dataset $\mathcal{D} = \{ (\textbf{x}^{(l)}, y^{(l)}) \}_{l=1}^{N}$, where $y^{(l)} = f^{tch}(\textbf{x}^{(l)}|\textbf{w}) + \epsilon^{(l)}$ is deemed to be a noisy observation of the scalarized output corresponding to $\textbf{x}^{(l)}$; $\epsilon^{(l)}$ is the additive Gaussian noise with zero mean. Then, the conditional predictive distribution of the GP for an unseen query point $\textbf{x}^{(*)}$ is also a Gaussian $\mathcal{N} ({\mu}_{gp}(\textbf{x}^{(*)}), {\sigma}_{gp}^2(\textbf{x}^{(*)}))$. The values of ${\mu}_{gp}(\textbf{x}^{(*)})$ and ${\sigma}_{gp}^2(\textbf{x}^{(*)})$ are calculated as:
\begin{equation}\label{eqn:forward_gp_mu}
\begin{aligned}
\mu_{gp}(\textbf{x}^{(*)}) = {\textbf{k}}_{*,gp}^\intercal ({\textbf{K}}_{gp} + \sigma^2\textbf{I}_{N})^{-1} \textbf{y}, \\ 
\end{aligned}
\end{equation}
\begin{equation}\label{eqn:forward_gp_sigma}
\begin{aligned}
\sigma_{gp}^2(\textbf{x}^{(*)}) = k_{gp}(\textbf{x}^{(*)},\textbf{x}^{(*)}) - {\textbf{k}}_{*,gp}^\intercal ({\textbf{K}}_{gp} + \sigma^2\textbf{I}_{N})^{-1} {\textbf{k}}_{*,gp},
\end{aligned}
\end{equation}
where $\textbf{k}_{*,gp}$ is the kernel vector between $\textbf{x}^{(*)}$ and the data in $\mathcal{D}$, $\textbf{K}_{gp}$ is the kernel matrix of the data in $\mathcal{D}$, $\textbf{y} = (y^{(1)},\ldots,y^{(N)})^\intercal$, $\sigma^2$ is the additive noise variance, and $\textbf{I}_{N}$ is a $N \times N$ identity matrix.

The (hyper-)parameters of the GP, which includes $\sigma^2$ and the parameters of the kernel function $k_{gp}(\cdot,\cdot)$, are to be estimated from the data. This is achieved by an empirical Bayes approach maximizing the following log marginal likelihood function:
\begin{equation}\label{eqn:likelihood_forward}
\begin{aligned}
-\frac{1}{2} \textbf{y}^{\intercal} ({\textbf{K}}_{gp} + \sigma^2\textbf{I}_{N})^{-1} \textbf{y}
- \frac{1}{2} \log (|{\textbf{K}}_{gp} + \sigma^2\textbf{I}_{N}|).
\end{aligned}
\end{equation} 
Any gradient-based optimizer, such as L-BFGS~\cite{zhu1997algorithm}, could be used to maximize (\ref{eqn:likelihood_forward}) to arrive at an optimized GP model. The training complexity scales as $\mathcal{O}(N^3)$.

\subsubsection{Inverse Gaussian Processes in MOPs}\label{sec:PE}
Existing multiobjective optimization algorithms are mostly designed to search for a finite set of nondominated samples that offer a good approximation to the PF. However, given objectives that are costly to evaluate, finding a sufficient number of such samples to cover the entire PF is a challenge. Consequently, the obtained nondominated samples may not adequately represent the solutions preferred by a DM. Inverse models emerge as a valuable tool for enhancing the density of PF approximation in such scenarios~\cite{lin2022pareto,giagkiozis2014pareto}. They learn to map points from the PF to nondominated solutions in the decision space. By querying sufficiently accurate inverse models, it becomes possible to generate nondominated solutions on-demand along any preferred subregion of the PF. 

The aforementioned idea is premised on the assumption that an inverse function from the PF to the PS can be defined, which is difficult to confirm in black-box optimization settings. However, prior research has shown that if the Karush-Kuhn-Tucker conditions hold in a given problem, then the PF and PS are both ($m$-1)-dimensional piecewise continuous manifolds, under certain mild conditions \cite{cheng2015multiobjective}. This makes it reasonable (and common) to assume that the mapping between the PS and the PF is bijective, implying that an inverse function would exist between them. In practice, the inverse modeling methodology in MOPs is found to be useful for populating PFs even if the bijectivity condition is not verified \cite{9005525, giagkiozis2014pareto}. 

The inputs to an inverse model are the preference vectors in objective space (as defined in Section \ref{sec:mop_decomposition}), associated with points along the PF. The approach thus enables DMs to query the model by specifying their desired trade-off preferences, and the model returns a prediction of what the corresponding candidate solution could be. For a principled probabilistic modeling setting that offers predictive uncertainties, this can be achieved by defining a Gaussian distribution over inverse functions as $\Psi_{j}^{-1} \sim \mathcal{GP}( \mu_j, {k}_j(\textbf{w},\textbf{w}') )$ \cite{suresh2023machine}. Here, $\Psi^{-1}_j(\textbf{w})$ represents the mapping from the preference vector $\textbf{w}$ in objective space to the $j$th decision variable of the associated nondominated solution in decision space; $\mu_j$ is the mean (typically set to zero or a constant) and ${k}_j(\cdot,\cdot)$ is the $j$th kernel function. The complete inverse prediction is then $[\Psi_{1}^{-1}(\textbf{w}), \Psi_{2}^{-1}(\textbf{w}), \ldots, \Psi_{d}^{-1}(\textbf{w})]$. Note that each component of this vector is a univariate Gaussian distribution, providing a probability distribution over possible solutions in the decision space.

For training the inverse GP, a dataset $\mathcal{D}^{inv} = \{ (\textbf{w}^{(l)},\textbf{x}^{(l)}) \}_{l=1}^{N^{nd}}$ of preference vectors $\textbf{w}^{(l)} = (w_{1}^{(l)},\ldots,w_{m}^{(l)})$ mapping to nondominated solutions $\textbf{x}^{(l)}$ is obtained from a multiobjective optimization run. The multiobjective optimizer used may not be decomposition-based, and hence well-defined subproblems with associated preference vectors may not be specified beforehand. As a general strategy, the formula in (\ref{Eq:transform_w}) can be employed to derive $\textbf{w}^{(l)}$ for $\textbf{x}^{(l)}$ based on objective function values normalized to (0, 1]:
\begin{equation}\label{Eq:transform_w}
\begin{aligned}
{w}_{i}^{(l)} = \frac{c_i^{(l)}}{\sum_{i=1}^{m}c_i^{(l)}}, i\in\{1,\ldots,m\}
\end{aligned}
\end{equation}
where $c_i^{(l)} = \frac{\sum_{v=1}^{m} f_{v}(\textbf{x}^{(l)})}{f_{i}(\textbf{x}^{(l)})}$. The rationale behind this formula is grounded in the following proposition.

\begin{proposition}
	
	Assuming $\eta = 0$ and $\textbf{x}^{ps}$ to be a Pareto optimal solution of the multiobjective optimization task, we have $f^{tch}(\textbf{x}^{ps}) = \min_{\textbf{x}} f^{tch}(\textbf{x})$ when the corresponding preference vector $\textbf{w}^{ps}$ is calculated as shown in \eqref{Eq:transform_w}.
\end{proposition}

A simple proof is provided in the Appendix. \textbf{Proposition 1} establishes that when $\eta = 0$, a nondominated solution minimizes the augmented Tchebycheff scalarized function with the preference vector calculated as per \eqref{Eq:transform_w}. This proposition remains approximately valid even for very small values of $\eta$ as in our implementation.

\section{The Proposed Method}\label{Section3}
This section introduces the proposed invTrEMO. We begin by outlining the problem statement under consideration. The process of constructing invTGP models for knowledge transfer across heterogeneous source-target task pairs is elucidated. Finally, we provide an overview of the overall framework of the algorithm.

\subsection{Problem Setup}
We consider the optimization of a target MOP with $d_T$ decision variables and $m$ objective functions that are expensive to evaluate. Constraint functions are not explicitly considered here, but constraint violations could be included as additional objectives to minimize \cite{coello2002theoretical}.  Taking a cue from the observation that real-world problems seldom exist in isolation, we assume there to exist data from a previously solved source MOP related to the target, but with $d_S$ decision variables. It is possible that $d_S \neq d_T$, thus rendering the transfer heterogeneous and most existing model-based transfer optimization algorithms ostensibly inapplicable. 

Taking a cue from the real-world, both tasks belonging to any given application area are considered to possess a common objective space---i.e., they share the same $m$ objective functions with the same physical meaning. Their decision spaces, on the other hand, may only be partially overlapping, with a nonempty subset of $Q$ decision variables bearing the same physical interpretation in both the source and target tasks. One can imagine new variables/features being added in the target relative to the source, or redundant features being removed; some concrete examples of use-cases will be showcased in Section \ref{sec:rwp}. The overlapping variables are indexed as $q \in \{1, \ldots, Q\}$.  \emph{The inverse transfer mechanism operates on these overlapping decision variables only}. The non-overlapping decision variables unique to the target task are indexed as $p \in \{Q+1, \ldots, d_T\}$. 

The invTrEMO takes advantage of dataset $\mathcal{D}_{S}^{inv} = \{ \textbf{W}_{S}, \textbf{X}_{S} \} = \{ (\textbf{w}_S^{(l)},\textbf{x}_S^{(l)})\}_{l=1}^{N_S^{nd}}$ from the source by melding it with $\mathcal{D}_{T}^{inv} = \{\textbf{W}_{T}, \textbf{X}_{T} \} =\{ (\textbf{w}_T^{(l)},\textbf{x}_T^{(l)})\}_{l=1}^{N_T^{nd}}$ from the target. $\textbf{x}_S^{(l)} \in \mathcal{X}_S$ and $\textbf{x}_T^{(l)} \in \mathcal{X}_T$ represent nondominated solutions for the source and target tasks, respectively, while $\textbf{w}_S^{(l)}$ and $\textbf{w}_T^{(l)}$ denote preference vectors calculated according to \eqref{Eq:transform_w} corresponding to $\textbf{x}_S^{(l)}$ and $\textbf{x}_T^{(l)}$, respectively. $N_S^{nd}$ and $N_T^{nd}$ are the total number of nondominated solutions that comprise the source and target datasets. Bear in mind that the preference vectors $\textbf{w}_S^{(l)},\textbf{w}_T^{(l)} \in \mathcal{W}$ fall within a common objective space. This space provides the necessary unification for invTGP models to be built, making it possible for knowledge transfer to take place even when the decision spaces of the source and the target do not exactly match. In what follows, we present details of the invTGP.

\subsection{Inverse Transfer Gaussian Processes}

\subsubsection{The invTGP Model}
In the invTrEMO, the invTGP models shall be updated repeatedly during the course of the optimization run. Hence, for simplicity, the invTGPs are proposed to be built independently for each (overlapping) target decision variable. Modeling the interactions between the target decision variables using a multi-output GP is possible, but this becomes computationally demanding due to the cubic training complexity with respect to the number of outputs \cite{bonilla2007multi}. For the $q$th overlapping decision variable, source dataset $\mathcal{D}_{S,q}^{inv} = \{ \textbf{W}_{S}, \textbf{X}_{S,q} \} = \{ (\textbf{w}_{S}^{(l)},{x}_{S,q}^{(l)})\}_{l=1}^{N_S^{nd}}$ and target dataset $\mathcal{D}_{T,q}^{inv} = \{ \textbf{W}_{T}, \textbf{X}_{T,q} \} = \{ (\textbf{w}_T^{(l)},{x}_{T,q}^{(l)})\}_{l=1}^{N_T^{nd}}$ are extracted from $\mathcal{D}_S^{inv}$ and $\mathcal{D}_T^{inv}$, where ${x}_{S,q}^{(l)}$ and ${x}_{T,q}^{(l)}$ are the $q$th component of the vectors $\textbf{x}_{S}^{(l)}$ and $\textbf{x}_{T}^{(l)}$, respectively. 

Extending from the vanilla inverse GP, an invTGP is able to leverage source-target relationships through the use of a \textit{transfer kernel} expressed as~\cite{cao2010adaptive}: 
\begin{equation}\label{eqn:transfer_kernel}
\begin{aligned}
\tilde{k}_q(\textbf{w},\textbf{w}') = 
\begin{cases}
{k}_q(\textbf{w},\textbf{w}'), & \textbf{w}, \textbf{w}' \in \textbf{W}_S \\
& \vee \;\textbf{w}, \textbf{w}' \in \textbf{W}_T \\
\lambda_{q} \cdot k_q(\textbf{w},\textbf{w}'), & \text{otherwise}. \\
\end{cases}
\end{aligned}
\end{equation}
Here $\lambda_q \in [-1,1]$ is a measure of source-target correlation that can be estimated from data; no manual tuning of $\lambda_q$ is required, precluding the need for any prior knowledge of source-target relations. If $\lambda_q$ is estimated to be close to $+1$ or $-1$, then there is expected to be high correlation between the source and target tasks, such that the data from the former is likely to be useful in the optimization of the latter. If $\lambda_q$ is close to $0$, the source and target tasks are expected to be unrelated. Note that in the geostatistics literature, the model in (\ref{eqn:transfer_kernel}) corresponds to the \emph{intrinsic coregionalization model}, a specific case of \emph{co-kriging} that uses a single (scalar) $\lambda_{q}$ to represent the inter-task relationship \cite{alvarez2012kernels}. The \emph{linear model of coregionalization}, on the other hand, could offer much greater flexibility in accurately capturing inter-task relationships by using multiple kernels \cite{wei2018uncluttered}, but at a cost of higher complexity of model training and inference.

For an arbitrary preference vector $\textbf{w}^{(*)} \in \mathcal{W}$, the posterior prediction of an invTGP model is then a Gaussian distribution $\mathcal{N}(\mu_{itgp,q}(\textbf{w}^{(*)}),\sigma_{itgp,q}^{2}(\textbf{w}^{(*)}))$, where $\mu_{itgp,q}(\textbf{w}^{(*)})$ and $\sigma_{itgp,q}^{2}(\textbf{w}^{(*)})$ are calculated as:
\begin{equation}\label{eqn:mu_tgp}
\begin{aligned}
\mu_{itgp,q}(\textbf{w}^{(*)}) = \tilde{\textbf{k}}_{*,q}^\intercal (\tilde{\textbf{K}}_q + \boldsymbol{\Lambda}_q)^{-1}
\begin{bmatrix}
\textbf{X}_{S,q} \\ 
\textbf{X}_{T,q} \\
\end{bmatrix},
\end{aligned}
\end{equation}
\begin{equation}\label{eqn:sigma_tgp}
\begin{aligned}
\sigma_{itgp,q}^2(\textbf{w}^{(*)}) = {k}_{q}(\textbf{w}^{(*)},\textbf{w}^{(*)}) - \tilde{\textbf{k}}_{*,q}^\intercal (\tilde{\textbf{K}}_q + \boldsymbol{\Lambda}_q)^{-1} \tilde{\textbf{k}}_{*,q}.
\end{aligned}
\end{equation}
In \eqref{eqn:mu_tgp} and \eqref{eqn:sigma_tgp}, $\textbf{k}_{*,q}$ is the kernel vector between $\textbf{w}^{(*)}$ and $\textbf{W} = \{\textbf{W}_S,\textbf{W}_T \}$, $\boldsymbol{\Lambda}_q = 
\begin{bmatrix} \sigma_{S,q}^2 \textbf{I}_{N_{S}^{nd}
}& 0 \\ 0 & \sigma_{T,q}^2 \textbf{I}_{N_T^{nd}}\\ \end{bmatrix}$, $\sigma_{S,q}^2$ and $\sigma_{T,q}^2$ are the additive noise variances of the source and target tasks, respectively, $\textbf{I}_{N_S^{nd}}$ and $\textbf{I}_{N_T^{nd}}$ are $N_S^{nd} \times N_S^{nd}$ and $N_T^{nd} \times N_T^{nd}$ identity matrices, respectively, $\tilde{\textbf{K}}_q = \begin{bmatrix} {\textbf{K}}_{SS,q} &\tilde{\textbf{K}}_{ST,q}\\ \tilde{\textbf{K}}_{TS,q} &{\textbf{K}}_{TT,q} \end{bmatrix}$ is the overall covariance matrix of the $q$th invTGP model, ${\textbf{K}}_{SS,q}$ and ${\textbf{K}}_{TT,q}$ are the kernel matrices of the data in the source and target tasks, respectively, and $\tilde{\textbf{K}}_{ST,q}$ ($=\tilde{\textbf{K}}_{TS,q}^\intercal$) is the kernel matrix across $\mathcal{D}_{S,q}^{inv}$ and $\mathcal{D}_{T,q}^{inv}$. 

\subsubsection{invTGP Model Training}
The invTGP model consists of several trainable (hyper-)parameters, including $\sigma_{S,q}$, $\sigma_{T,q}$, $\lambda_q$ and the parameters of $k_q(\cdot,\cdot)$. These are typically estimated by training the invTGP model using the joint distribution of the source and target tasks~\cite{bonilla2007multi}. However, this approach can bias the model towards the source rather than the target, especially when $N_{T}^{nd} \ll N_{S}^{nd}$ at the initial stages of the target optimization. To address this issue, we propose a two step training process that reflects our primary interest in the inverse model's performance in addressing the target task. In the first step, we learn $\sigma_{T,q}$ and the parameters of $k_q(\cdot,\cdot)$  based on $\mathcal{D}_{T,q}^{inv}$ alone, by maximizing the following log marginal likelihood function:
\begin{equation}\label{eqn:likelihood_1}
\begin{aligned}
-\frac{1}{2} \textbf{X}_{T,q}^\intercal ({\textbf{K}}_{TT,q} + \sigma_{T,q}^2 \textbf{I}_{N_T^{nd}})^{-1} \textbf{X}_{T,q}
- \frac{1}{2} \log (|{\textbf{K}}_{TT,q} + \sigma_{T,q}^2 \textbf{I}_{N_T^{nd}}|).
\end{aligned}
\end{equation}
The second step keeps the trained kernel parameters and $\sigma_{T,q}$ fixed, and optimizes $\lambda_{q}$ and $\sigma_{S,q}$ by maximizing the following log marginal likelihood jointly on $\mathcal{D}_{S,q}^{inv}$ and $\mathcal{D}_{T,q}^{inv}$:
\begin{equation}\label{eqn:likelihood_2}
\begin{aligned}
-\frac{1}{2} [\textbf{X}_{S,q}^\intercal,\textbf{X}_{T,q}^\intercal] (\tilde{\textbf{K}}_q + \boldsymbol{\Lambda}_q)^{-1} 
\begin{bmatrix}
\textbf{X}_{S,q} \\
\textbf{X}_{T,q} \\ 
\end{bmatrix}
- \frac{1}{2} \log (|\tilde{\textbf{K}}_q + \boldsymbol{\Lambda}_q|).
\end{aligned}
\end{equation}

\subsection{Overall Framework of the invTrEMO}

\begin{algorithm}[t]
	\caption{\textit{The invTrEMO}}\label{Alg:ITEMO}
	\renewcommand{\algorithmicrequire}{\textbf{Input:}} 
	\renewcommand{\algorithmicensure}{\textbf{Output:}}
	\begin{algorithmic}[1]
		\REQUIRE Target MOP objective functions $\textbf{f}_T(\textbf{x})$; decision space $\mathcal{X}_T$; \emph{evaluation budget}; offspring population size $N^{u}$; predefined preference vectors $\mathcal{W}$; source dataset $\mathcal{D}^{inv}_{S}$; 
		\ENSURE Inverse models and the obtained nondominated target solutions;
		\STATE $t \leftarrow 0$;
		\STATE Sample a population of $N^{init}$ solutions $\textbf{x}^{(1)}_{T},\ldots,\textbf{x}^{(N^{init})}_{T}$ in $\mathcal{X}_T$;
		\STATE $\mathcal{D}^M_T \leftarrow \{(\textbf{x}^{(l)}_{T},\textbf{f}_T(\textbf{x}_{T}^{(l)}))\}_{l=1}^{N=N^{init}+t}$
		\WHILE {$N < $ {\emph{evaluation budget}}}
		\STATE Sample a preference vector $\textbf{w}^{(t)}$ from $\mathcal{W}$;
		\STATE Scalarize $\mathcal{D}^M_{T}$ based on \eqref{eqn:techebycheff} and $\textbf{w}^{(t)}$ to generate dataset $\mathcal{D}_T= \{(\textbf{x}^{(l)}_{T},y_T^{(l)})\}_{l=1}^{N}$;
		\STATE Update the forward GP model based on $\mathcal{D}_T$ to get predictions $\mathcal{N} ({\mu}_{gp}(\textbf{x}^{(*)}), {\sigma}_{gp}^2(\textbf{x}^{(*)}))$;
		\STATE Generate $\mathcal{D}^{inv}_{T}$ based on \eqref{Eq:transform_w} and the nondominated samples in $\mathcal{D}^M_{T}$;
		\STATE Update invTGP models for overlapping decision variables based on $\mathcal{D}^{inv}_{S}$ and $\mathcal{D}^{inv}_{T}$ to make predictions $\mathcal{N} ({\mu}_{itgp,q}(\textbf{w}^{(*)}), {\sigma}_{itgp,q}^2(\textbf{w}^{(*)})), q \in \{1,\ldots,Q\}$;
		\STATE Update inverse GP models for non-overlapping target decision variables based on $\mathcal{D}^{inv}_{T}$ to make predictions $\mathcal{N} ({\mu}_{igp,p}(\textbf{w}^{(*)}), {\sigma}_{igp,p}^2(\textbf{w}^{(*)})), p \in \{Q+1,\ldots,d_T\}$;
		\STATE Query the inverse models with $\textbf{w}^{(t)}$ to get a predictive distribution over solutions in $\mathcal{X}_T$;
		\STATE Sample offspring population $\mathcal{U}$ of $N^{u}$ solutions from the predicted distribution;
		\STATE Select solution $\textbf{x}_{T}^{(t)} = \arg\max_{\textbf{u}\in\mathcal{U}}(-{\mu}_{gp}$ $(\textbf{u}) + \beta \cdot {\sigma}_{gp}(\textbf{u}))$;
		\STATE $\mathcal{D}^M_T \leftarrow \mathcal{D}^M_T \cup \{(\textbf{x}_{T}^{(t)},\textbf{f}_T(\textbf{x}_{T}^{(t)}))\}$; 
		\STATE $t \leftarrow t + 1$;
		\ENDWHILE
		\STATE Rank samples in $\mathcal{D}_T^M$ based on Pareto dominance;
		\RETURN Inverse models and nondominated solutions in $\mathcal{D}_T^M$.
	\end{algorithmic}
\end{algorithm}

A pseudocode of the invTrEMO is presented in \textbf{Algorithm 1}. The steps are detailed below. Throughout the algorithm it is assumed that the cost of function evaluations is sufficiently high to ameliorate the cost of updating the forward and inverse GPs in every iteration. This is a reasonable assumption for many  MOPs where evaluating a solution calls for time-consuming computer simulations or costly real-world procedures to be carried out.

\begin{itemize}
	\item \textbf{Initialization} (steps 1-3): The iteration counter $t$ is set to 0, and $N^{init}$ initial solutions are generated either randomly or using Latin square sampling in the target decision space. These initial solutions are evaluated using the objectives of the target task, creating the initial dataset $\mathcal{D}^M_{T} =\{(\textbf{x}^{(l)}_{T},\textbf{f}_T(\textbf{x}_{T}^{(l)}))\}_{l=1}^{N}$, where $N = N^{init} + t$.
	
	\item \textbf{Problem Decomposition} (steps 5-6): In any iteration $t$ of the invTrEMO, we randomly select a preference vector $\textbf{w}^{(t)}$ from the predefined set $\mathcal{W}$ of uniformly distributed preference vectors in the common objective space. This is used to produce a scalarized target dataset $\mathcal{D}_{T}$ based on the augmented Tchebycheff function defined in \eqref{eqn:techebycheff}, $\mathcal{D}^M_{T}$, and $\textbf{w}^{(t)}$. 
	
	\item \textbf{Gaussian Process Modeling} (step 7): A GP model is built based on the dataset $\mathcal{D}_{T}$ for cheap GP evaluations of the augmented Tchebycheff scalarized function. 
	
	\item \textbf{Inverse Modeling and Transfer} (steps 8-10): A dataset $\mathcal{D}_T^{inv}$ for inverse modeling is created by applying \eqref{Eq:transform_w} to the nondominated solutions found so far in $\mathcal{D}_T^{M}$. For those decision variables that overlap in the source and target tasks, invTGP models using both $\mathcal{D}_S^{inv}$ and $\mathcal{D}_T^{inv}$ are built. For the remaining target decision variables,  independent inverse GPs using only $\mathcal{D}_T^{inv}$ are built. During training of the inverse models, we constrain the additive noise term's standard deviation to be at least $\sigma_0$, i.e., $\sigma_{T,q} \geq \sigma_0$. This is important because samples in $\mathcal{D}_T^{inv}$ may not represent the true Pareto optimal of the target task, especially at the early stages of the invTrEMO, violating the requirements of \textbf{Proposition 1}. Including a minimum additive noise level allows the search algorithm to sample and evaluate new data, reducing the chance of getting trapped by overconfident but inaccurate predictions. 
	
	\item \textbf{Evolutionary Variation and Selection} (steps 11-14): Given the selected preference vector $\textbf{w}^{(t)}$ at iteration $t$, an offspring population $\mathcal{U} = \{\textbf{u}^{(l)}\}_{l=1}^{N^{u}}$ is generated by sampling the multivariate Gaussian distribution  $[\Psi_{1}^{-1}(\textbf{w}^{(t)}), \Psi_{2}^{-1}(\textbf{w}^{(t)}), \ldots, \Psi_{d_T}^{-1}(\textbf{w}^{(t)})]$ in decision space, formed by querying the inverse models with $\textbf{w}^{(t)}$. Thereafter, the upper confidence bound (UCB)~\cite{srinivas2012information} acquisition function is employed as the figure of merit to select\footnote{In this procedure, $\textbf{x}_{T}^{(t)}$ is selected by utilizing the predictions provided by the forward GP model. Given that the evaluation cost associated with this predictive process is substantially lower than that of the true expensive evaluations, we operate under the assumption that this process does not incur any significant computational overhead.} one solution $\textbf{x}_{T}^{(t)}$ from $\mathcal{U}$. For minimization problems, the UCB is defined as $-{\mu}_{gp}(\textbf{x}^{(*)}) + \beta \cdot {\sigma}_{gp}(\textbf{x}^{(*)})$, where $\beta$ is an appropriate constant. The selected $\textbf{x}_{T}^{(t)}$ is evaluated using the true objective functions, following which $\textbf{x}_{T}^{(t)}$ and $\textbf{f}_T(\textbf{x}_{T}^{(t)})$ are appended to $\mathcal{D}_T^{M}$.
\end{itemize}

Steps 5 to 14 are repeated until the termination condition, e.g., evaluation budget, is met. The algorithm then returns the inverse models and the nondominated solutions contained in $\mathcal{D}^{M}_T$. 

\subsection{Summarizing the Transfer Mechanism of invTrEMO}

\begin{figure}[!t]
	\begin{center}
		\includegraphics[width=0.85\columnwidth]{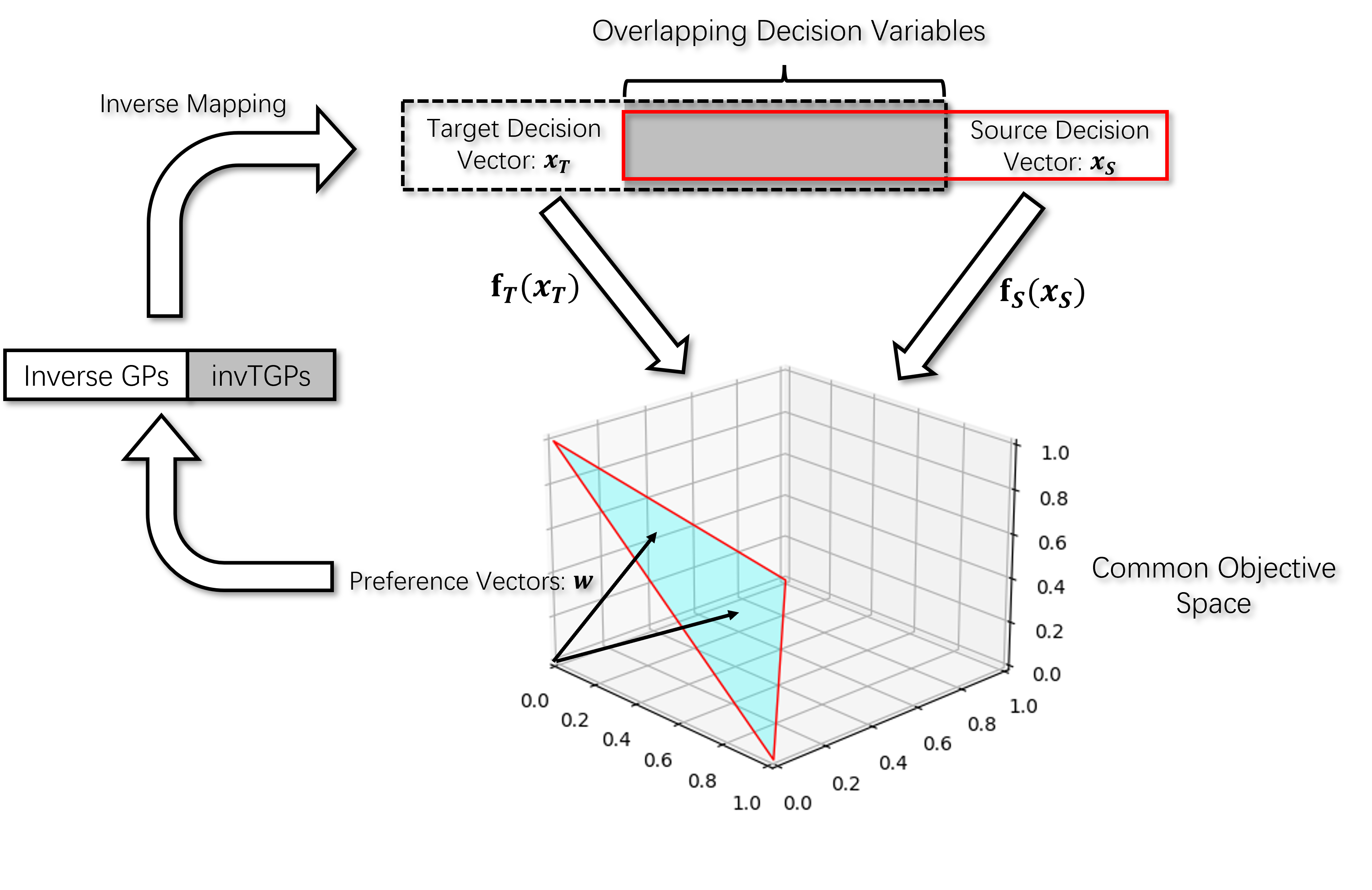}
		\caption{An illustration of the assumed relationship between the source and the target tasks in our problem setup. The common objective space provides the unification through which knowledge transfer could take place, in the inverse setting, between the subset of variables that overlap. Decision variables in distinct tasks are said to overlap if they bear the same physical interpretation in the application domain of interest.}\label{Fig:Source_Target}
	\end{center}
\end{figure}

As illustrated in Fig.~\ref{Fig:Source_Target}, during the forward mapping, the objective functions of both the source and target tasks map solutions from the decision space to a common objective space. Conversely, the inverse mapping translates points on the PF of the common objective space back to the target task's decision space given a preference vector as input. In invTrEMO, we approximate this inverse mapping using $Q$ independent invTGPs and $(d_T - Q)$ vanilla inverse GPs. Knowledge transfer occurs through the invTGPs, across the $Q$ decision variables that overlap between the source and the target tasks. Unlike a very recent inverse model-based algorithm PSL-MOBO~\cite{lin2022pareto}, these invTGPs are trained utilizing both the source dataset $\mathcal{D}_{S,q}^{inv}$ and the target dataset $\mathcal{D}_{T,q}^{inv}$, thereby encapsulating knowledge from both domains.

In comparison to conventional forward surrogate model-based optimization, knowledge transfer is expected to quickly guide the target search toward performant solutions. The idea is to harness data from source tasks to curtail the need for excessive and expensive exploration of the target search space. This point is experimentally analyzed in Section S-II.1 of the supplementary file, where we show that the offspring sampled from the inverse models' predicted distribution have significantly lower UCB values than the maximum (as estimated by the forward GP). The invTGP can therefore be said to exploit the source data to achieve more focused target exploration. The effectiveness of the resulting search behavior of the invTrEMO is demonstrated in the experimental studies that follow.

\section{Experimental Studies}
In this section, we study the performance of the invTrEMO on multi- and many-objective benchmark problems and on real-world vehicle crashworthiness design problems.

\subsection{Multiobjective Benchmark Problems}
We first create eight base problems, each with three objective functions, by modifying the well-known DTLZ and DTLZ$^{-1}$ benchmark suites~\cite{deb2005scalable,ishibuchi2016performance}. These eight base problems are denoted as mDTLZ1-($\delta_1$,$\delta_2$), mDTLZ2-($\delta_1$,$\delta_2$), mDTLZ3-($\delta_1$,$\delta_2$), mDTLZ4-($\delta_1$,$\delta_2$), mDTLZ1$^{-1}$-($\delta_1$,$\delta_2$), mDTLZ2$^{-1}$-($\delta_1$,$\delta_2$), mDTLZ3$^{-1}$-($\delta_1$,$\delta_2$), and mDTLZ4$^{-1}$-($\delta_1$,$\delta_2$), where $\delta_1$ and $\delta_2$ serve as two control parameters to vary the problem. The formulations of these eight benchmarks can be found in the supplementary file. By setting $\delta_1$ and $\delta_2$ to 1 and 0, respectively, we generate four target tasks with $d_T = 8$, namely mDTLZ1-($1$,$0$), mDTLZ2-($1$,$0$), mDTLZ3-($1$,$0$), mDTLZ4-($1$,$0$), mDTLZ1$^{-1}$-($1$,$0$), mDTLZ2$^{-1}$-($1$,$0$), mDTLZ3$^{-1}$-($1$,$0$), and mDTLZ4$^{-1}$-($1$,$0$). Additionally, a series of source tasks are created by adjusting the values of $\delta_1$ and $\delta_2$.
\begin{figure}[!t]
	\begin{center}
		\subfigure[]{\label{fig:similarity_1}\includegraphics[width=0.4\columnwidth]{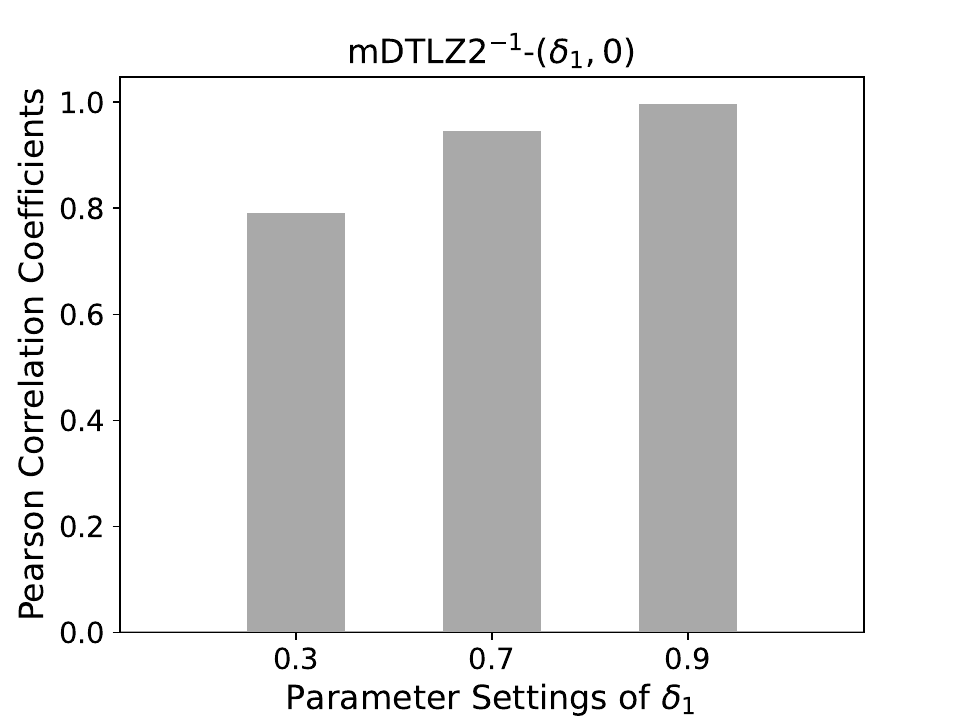}}
		\subfigure[]{\label{fig:similarity_2}\includegraphics[width=0.4\columnwidth]{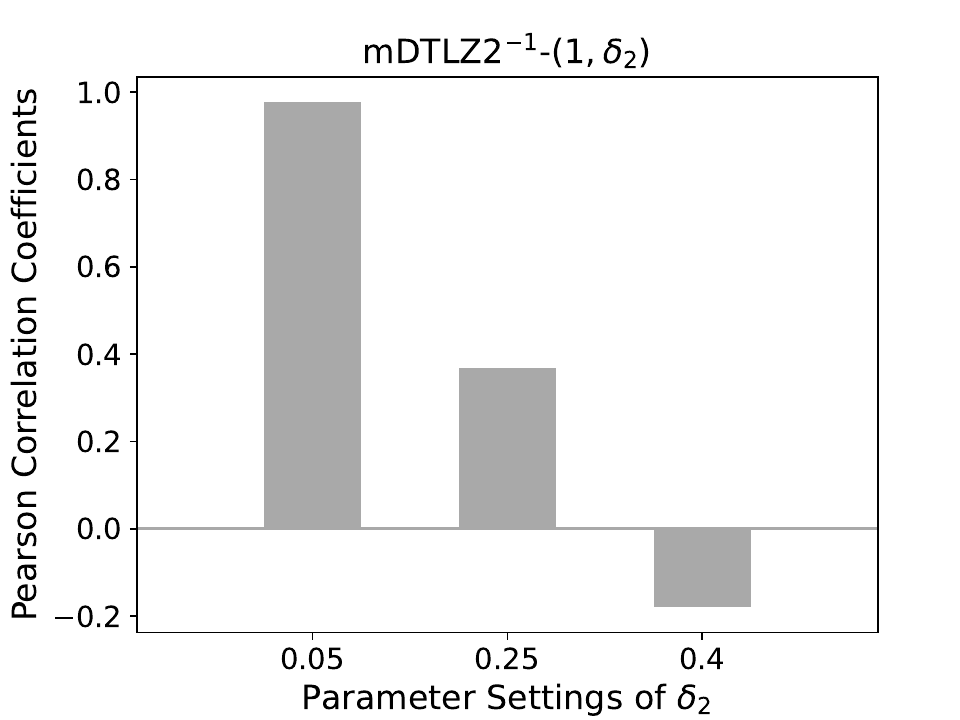}}
		\caption{Influence of $\delta_1$ and $\delta_2$ on source-target correlations. (a) Pearson correlation coefficients between mDTLZ2$^{-1}$-$(1,0)$ and mDTLZ2$^{-1}$-$(\delta_1,0)$ when $\delta_1$ is set to 0.3, 0.7, or 0.9. (b) Pearson correlation coefficients between mDTLZ2$^{-1}$-$(1,0)$ and mDTLZ2$^{-1}$-$(1,\delta_2)$ when $\delta_2$ is set to 0.05, 0.25, or 0.4.}\label{Fig:Control_Par}
	\end{center}
\end{figure}

\begin{table*}[t]
	\centering
	\caption{The Pearson correlation coefficients between the augmented Tchebycheff scalarized objective functions of the target task and various categories of source tasks.}\label{Tab:Pearson}
	\resizebox{12cm}{!}{\begin{tabular}{c|c|c|c|cc|c}      
			\hline
			\diagbox{Source Tasks}{Target Tasks}   & {mDTLZ1-($1,0$) }  & mDTLZ2-($1,0$)  & mDTLZ3-($1,0$)  & mDTLZ4-($1,0$)  \\
			\hline
			Highly-Correlation Source (HS)     & 0.9787  & 0.9831  & 0.9693  & 0.9804   \\
			Medium-Correlation Source (MS)   & 0.7762  & 0.7839  & 0.7761  & 0.7832   \\
			Low-Correlation Source (LS)      & 0.4899 & 0.4922 & 0.4928 & 0.4546  \\
			\hline
			\hline
			\diagbox{Source Tasks}{Target Tasks}   & {mDTLZ1$^{-1}$-($1,0$) }  & mDTLZ2$^{-1}$-($1,0$)  & mDTLZ3$^{-1}$-($1,0$)  & mDTLZ4$^{-1}$-($1,0$)  \\
			\hline
			Highly-Correlation Source (HS)     & 0.8934  & 0.9731  & 0.9729  & 0.9775   \\
			Medium-Correlation Source (MS)   & 0.3435  & 0.3642  & 0.3641  & 0.3459   \\
			Low-Correlation Source (LS)      & 0.1107 & -0.0905 & -0.1146 & -0.1169  \\
			\hline
	\end{tabular}}
\end{table*}

The choice of $\delta_1$ and $\delta_2$ values can influence the correlation between the source and target tasks. To approximately quantify this correlation, we compute the Pearson correlation coefficient between the \textit{scalarized} objective functions of the source and the target task. Fig.~\ref{Fig:Control_Par} illustrates the influence of these two control parameters on mDTLZ2$^{-1}$-($\delta_1$,$\delta_2$). Notably, when $\delta_1$ or $\delta_2$ deviate significantly from 1 or 0, the source-target correlation is low; conversely, when they are close to 1 or 0, a high source-target correlation is observed. Based on these observations, for each target task, we create source tasks with three levels of source-target correlation by setting $\delta_1$ and $\delta_2$ to different values. These levels are as follows: highly-correlation source (HS: $\delta_1=0.9$ and $\delta_2=0.05$), medium-correlation source (MS: $\delta_1=0.7$ and $\delta_2=0.25$), and low-correlation source (LS: $\delta_1=0.3$ and $\delta_2=0.4$). The Pearson correlation coefficients for all source-target pairs are summarized in Table~\ref{Tab:Pearson}. To assess invTrEMO's performance in coping with heterogeneous source-target pairs, we set $d_S = 6$; thus $d_S < d_T$. We assume that the first six decision variables of the target task overlap with the variables of the source task. The two remaining variables in the target are non-overlapping.

To construct the source dataset $\mathcal{D}^{inv}_{S}$ based on the generated source tasks, the following steps were followed: 1) the source MOP was optimized to obtain a set of nondominated solutions, and 2) preference vectors corresponding to the nondominated solutions were generated as per \eqref{Eq:transform_w}, thereby creating $\mathcal{D}^{inv}_{S}$. In our experiments, we employed NSGA-III~\cite{deb2013evolutionary} from Pymoo~\cite{pymoo} to solve the source tasks. The population size and the number of reference vectors in NSGA-III were both set to 100, and the population was evolved for 500 generations. This provided us source datasets containing 100 entries of optimized solutions. 

\subsection{Evaluation Metrics}
We utilize two metrics, one to evaluate the performance of a multiobjective optimization algorithm and the other the accuracy of the inverse models created. The former is measured by the inverted generational distance (IGD), while accuracy is quantified by the root mean square error (RMSE).

The IGD measures the Euclidean distance between elements in the approximated PF and the true PF. It is calculated as follows:
\begin{equation}\label{eqn:IGD}
\begin{aligned}
IGD = \frac{1}{N^r} \sum_{r=1}^{N^r} \min \{ || \textbf{f}_{T}(\textbf{x}_{opt}^{(r)}) - \textbf{f}_T(\textbf{x}_T^{(1)}) ||_2, \ldots, || \textbf{f}_{T}(\textbf{x}_{opt}^{(r)}) - \textbf{f}_T(\textbf{x}_T^{(N_T^{nd})}) ||_2 \}
\end{aligned}
\end{equation}
where $\{ \textbf{x}_{opt}^{(1)},\ldots,\textbf{x}_{opt}^{(N^r)} \}$ is a set of true Pareto optimal solutions that map to well-distributed points along the PF, and  $\textbf{x}_T^{(1)},\ldots,\textbf{x}_T^{(N_T^{nd})}$ are the nondominated solutions in $\mathcal{D}_{T}^{M}$. $N^r$ was set to 10000.

RMSE measures the predictive accuracy of the inverse models on a test set $\mathcal{D}_{opt} = \{(\textbf{w}_{opt}^{(r)},\textbf{x}_{opt}^{(r)})\}_{r=1}^{N^r}$, where $\textbf{w}_{opt}^{(r)}$ is the preference vector corresponding to $\textbf{x}_{opt}^{(r)}$. Since the goal of the inverse model is to help satisfy DM preferences articulated in the objective space, we calculate the RMSE also in the objective space as:
\begin{equation}\label{eqn:RMSE}
\begin{aligned}
RMSE = \sqrt{ \frac{\sum_{r=1}^{N^r} || \textbf{f}_T(\textbf{x}_{opt}^{(r)}) - \textbf{f}_T( \textbf{x}_{pred}^{(r)} ) ||_2^2 }{N^r}}
\end{aligned}
\end{equation}
where $\textbf{x}_{pred}^{(r)}$ is the prediction of the inverse models corresponding to $\textbf{w}_{opt}^{(r)}$. The predicted means of the inverse GP or invTGP are taken as the prediction $\textbf{x}_{pred}^{(r)}$. 

\subsection{Parameter Settings and Experimental Details}
In all of our experiments we set $N^{init}$ to 20 and considered a total evaluation budget of 100 evaluations. The Riesz s-energy method~\cite{ref_dirs_energy} was employed to create a set $\mathcal{W}$ of 50 uniformly distributed preference vectors. Both the forward and inverse GP models are implemented using GPyTorch~\cite{gardner2018gpytorch}. The squared exponential function is selected as the GP kernel, and the L-BFGS~\cite{zhu1997algorithm} algorithm is used to train the model (hyper-)parameters.

During the optimization process, the minimum additive
noise level  $\sigma_0$ was set to 0.01 for all inverse models. $N^{u}=10000$ offspring are sampled in each iteration of the invTrEMO. In accordance with the recommendation in~\cite{lin2022pareto}, we set $\beta$ to 0.5 in the UCB acquisition function. Note that these parameters were not extensively fine-tuned as they generally demonstrated good performance.

\subsection{Comparison with State-of-the-Art Evolutionary and Bayesian Optimizers}

\begin{figure*}[]
	\begin{center}
		\subfigure[]{\label{fig:convergence_1}\includegraphics[width=0.24\columnwidth]{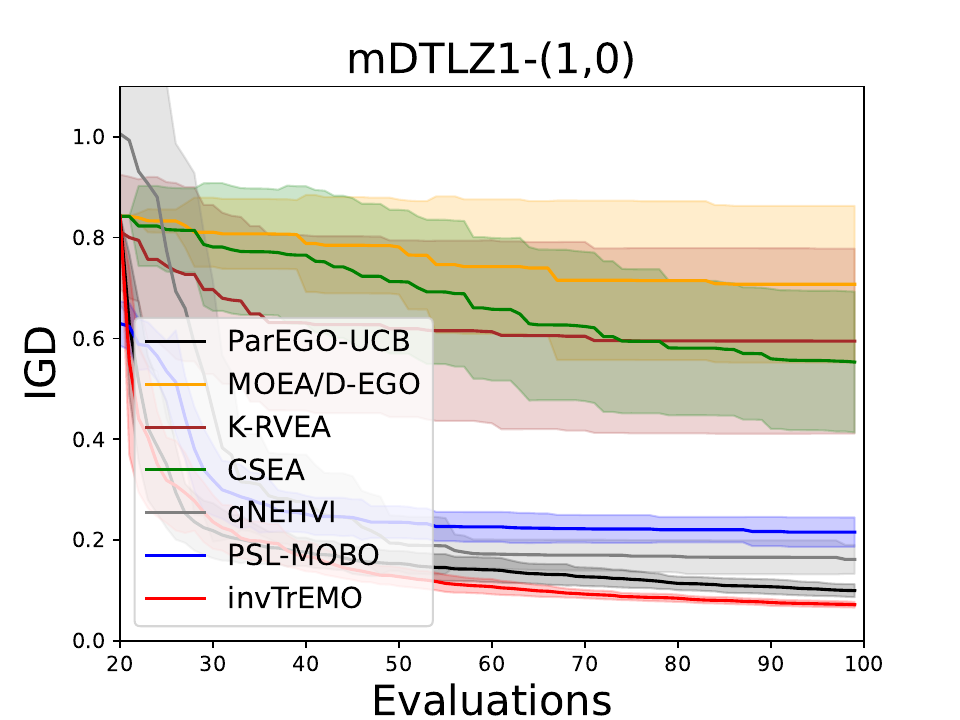}}
		\subfigure[]{\label{fig:convergence_2}\includegraphics[width=0.24\columnwidth]{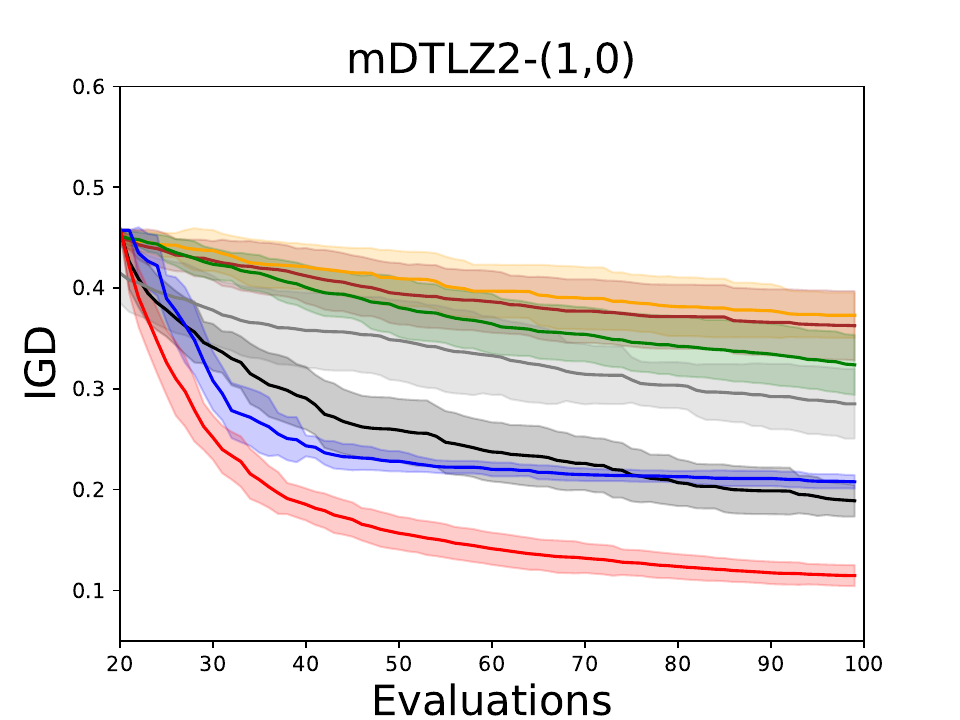}}
		\subfigure[]{\label{fig:convergence_3}\includegraphics[width=0.24\columnwidth]{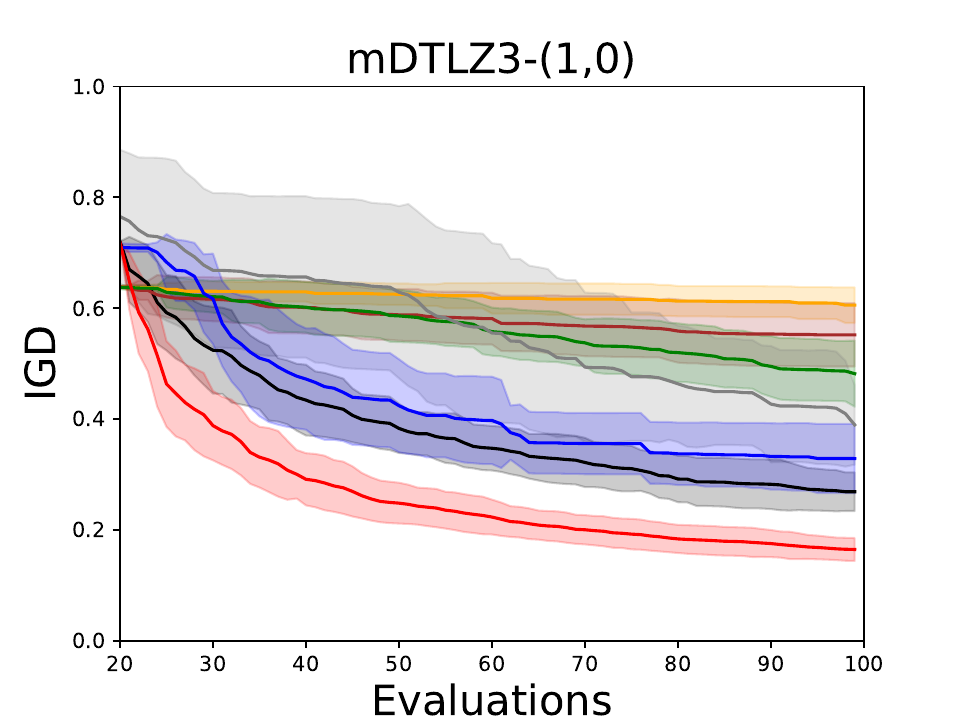}}
		\subfigure[]{\label{fig:convergence_4}\includegraphics[width=0.24\columnwidth]{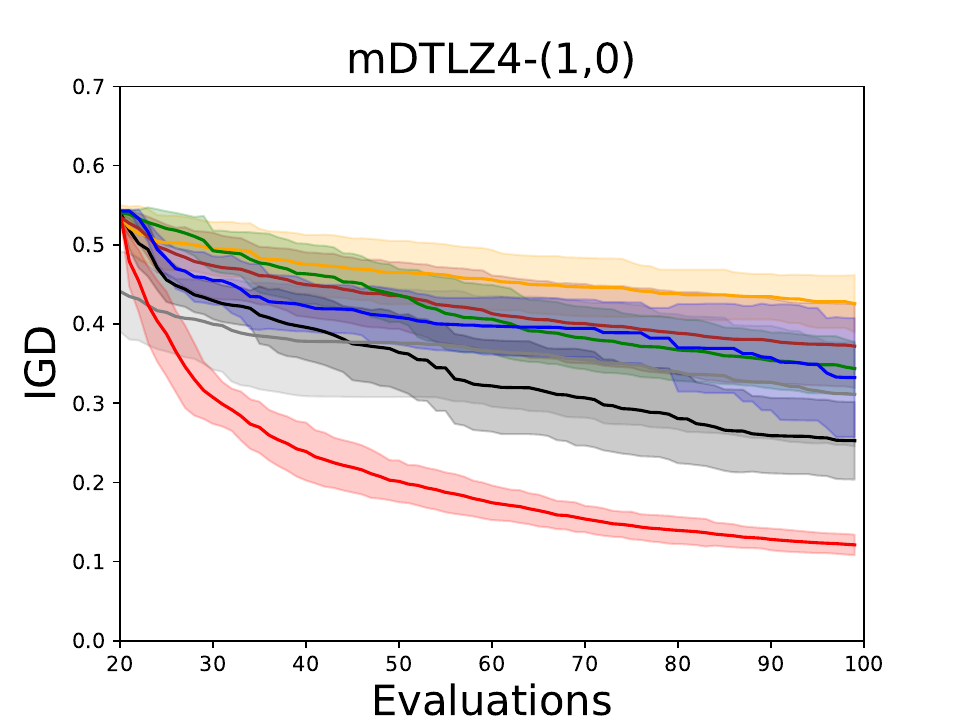}}
		\subfigure[]{\label{fig:convergence_1}\includegraphics[width=0.24\columnwidth]{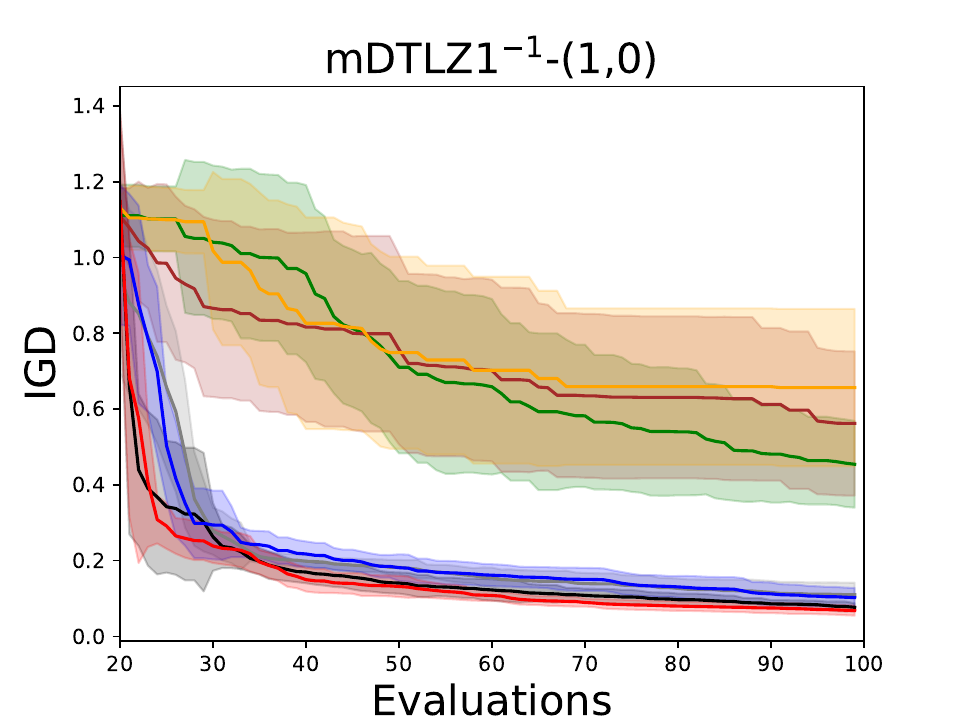}}
		\subfigure[]{\label{fig:convergence_2}\includegraphics[width=0.24\columnwidth]{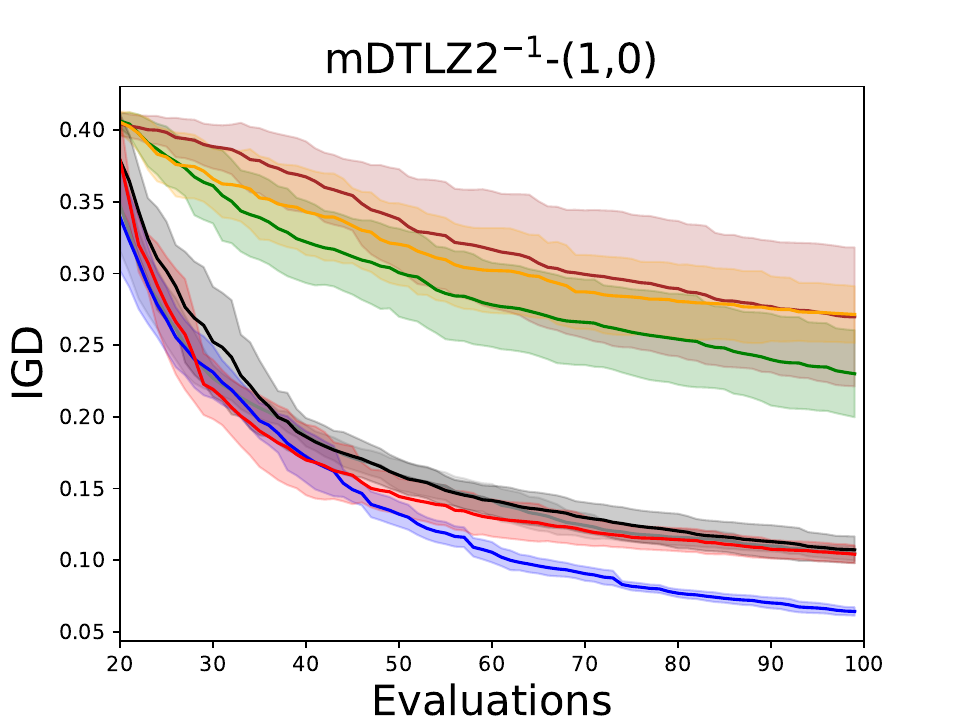}}
		\subfigure[]{\label{fig:convergence_3}\includegraphics[width=0.24\columnwidth]{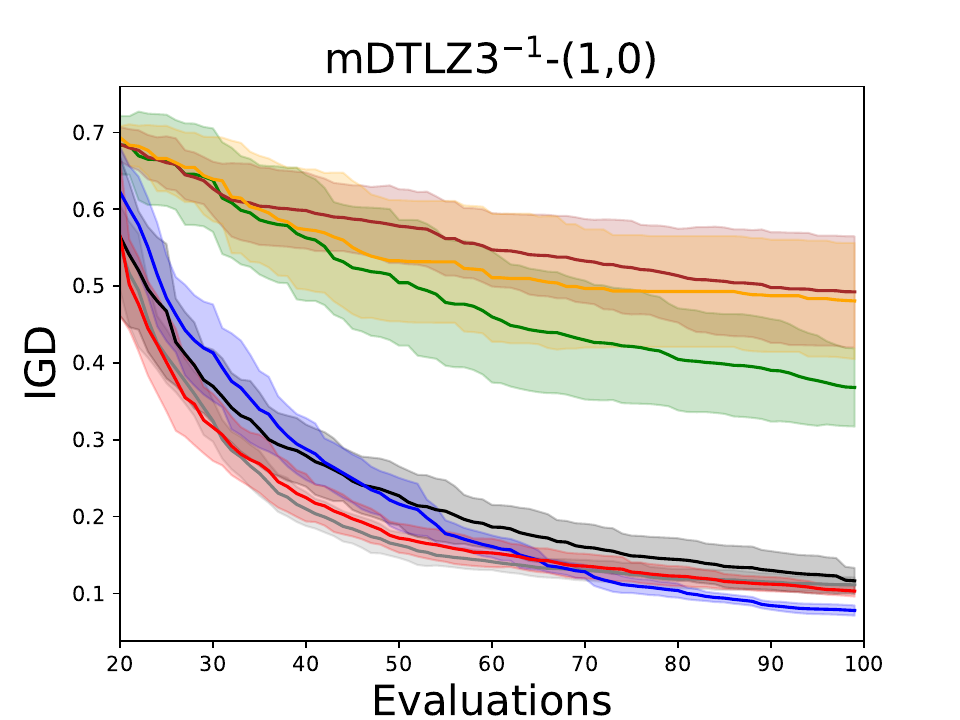}}
		\subfigure[]{\label{fig:convergence_4}\includegraphics[width=0.24\columnwidth]{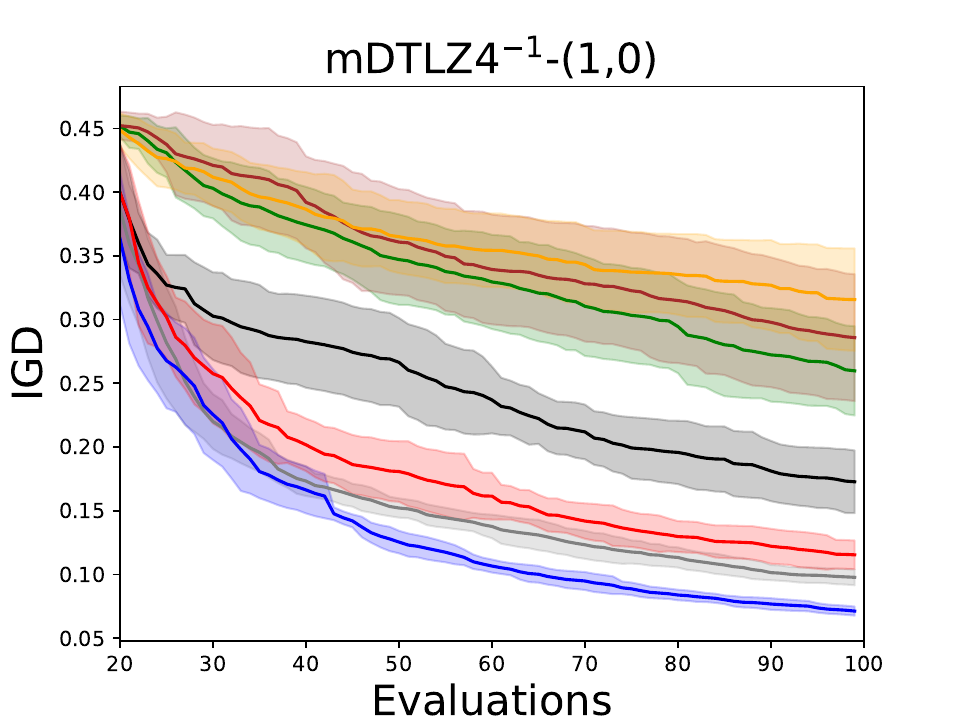}}
		\caption{Comparison of IGD convergence trends averaged over 20 independent runs of ParEGO-UCB, MOEA/D-EGO, K-RVEA, CSEA, qNEHVI, PSL-MOBO, and the invTrEMO. Shaded areas represent one standard deviation on either side of the mean. (a) mDTLZ1-($1,0$). (b) mDTLZ2-($1,0$). (c) mDTLZ3-($1,0$). (d) mDTLZ4-($1,0$). (e) mDTLZ1$^{-1}$-($1,0$). (f) mDTLZ2$^{-1}$-($1,0$). (g) mDTLZ3$^{-1}$-($1,0$). (h) mDTLZ4$^{-1}$-($1,0$).}\label{Fig:Baseline}
	\end{center}
\end{figure*}

In this subsection, we utilize an MS source dataset to inform the invTrEMO in solving the target task. A more comprehensive study of the effect of source-target correlations is given in Section \ref{subsec:source-target corr}. We compare the performance of the invTrEMO with the following algorithms:
\begin{itemize}
	\item ParEGO-UCB~\cite{1583627}: This is a variant of the classical ParEGO algorithm in which the UCB is used as the acquisition function, and the GP model is implemented using GPyTorch.
	
	\item MOEA/D-EGO~\cite{5353656}: A GP-assisted MOEA/D.
	
	\item K-RVEA~\cite{7723883}: A GP-assisted reference vector guided evolutionary algorithm for many-objective optimization.
	
	\item CSEA~\cite{8281523}: A classification surrogate model assisted evolutionary algorithm.
	
	\item PSL-MOBO~\cite{lin2022pareto}: A state-of-the-art MOBO algorithm with Pareto set learning (i.e., inverse modeling). For fairness of comparison, we set the batch size of PSL-MOBO to 1 which gives better performance on our test problems than the default of 5.
	
	\item qNEHVI~\cite{daulton2021parallel}: A Bayesian optimization algorithm of multiple noisy objectives with expected hypervolume improvement. For fair comparison, we set the batch size of qNEHVI to 1.
\end{itemize}
The above forms a representative set of proficient algorithms in evolutionary and Bayesian optimization, for problems with objectives that are costly to evaluate. Source codes of the invTrEMO and ParEGO-UCB can be found at https://github.com/LiuJ-2023/invTrEMO, with the performance of the ParEGO-UCB favorably verified against the original ParEGO. The source code of PSL-MOBO can be found at https://github.com/Xi-L/PSL-MOBO. The qNEHVI was implemented using BoTorch~\cite{balandat2020botorch}. To ensure fairness in our comparative analysis, the hyperparameters of the forward GPs employed in qNEHVI, ParEGO-UCB, and invTrEMO are configured identically. The implementations of MOEA/D-EGO, K-RVEA, and CSEA were obtained from PlatEMO~\cite{8065138}. We have summarized the averaged IGD values achieved by the six algorithms after 25, 50, 75, and 100 evaluations (which include the initial $N^{init}$ samples) in Table~\ref{Tab:Baseline}, and the IGD convergence trends are depicted in Fig.~\ref{Fig:Baseline}.

\begin{table*}[]
	\centering
	\caption{IGD results of ParEGO-UCB, MOEA/D-EGO, K-RVEA, CSEA, qNEHVI, PSL-MOBO, and the invTrEMO (informed by an MS source dataset) at 25, 50, 75, and 100 evaluations, averaged over 20 independent runs of every optimizer. The Kruskal-Wallis Test at a 0.05 significance level was performed to compare invTrEMO with the other algorithms.}\label{Tab:Baseline}
	\resizebox{14cm}{!}{\begin{tabular}{c|c|c|c|c|c|c|c|c|c|c|c|c|c|c}
			\hline
			\multirow{2}{*}{Target Tasks} & \multirow{2}{*}{Evaluations} & \multicolumn{2}{c|}{ParEGO-UCB} & \multicolumn{2}{c|}{MOEA/D-EGO} & \multicolumn{2}{c|}{K-RVEA} & \multicolumn{2}{c|}{CSEA} & \multicolumn{2}{c|}{qNEHVI} & \multicolumn{2}{c|}{PSL-MOBO} & \multicolumn{1}{c}{invTrEMO}      \\
			\cline{3-15}
			&  & \multicolumn{2}{c|}{Average IGD $\pm$ Std}  & \multicolumn{2}{c|}{Average IGD $\pm$ Std} & \multicolumn{2}{c|}{Average IGD $\pm$ Std} & \multicolumn{2}{c|}{Average IGD $\pm$ Std} & \multicolumn{2}{c|}{Average IGD $\pm$ Std} & \multicolumn{2}{c|}{Average IGD $\pm$ Std} & \multicolumn{1}{c}{Average IGD$\pm$ Std}     \\
			\hline
			\multirow{4}{*}{mDTLZ1-($1,0$)}  & 25  & 0.3858$\pm$0.1333 & \multirow{4}{*}{$+$} & 0.5089$\pm$0.0243 & \multirow{4}{*}{$+$} & 0.4699$\pm$0.0568 & \multirow{4}{*}{$+$} & 0.5179$\pm$0.0189 & \multirow{4}{*}{$+$} & 0.8803$\pm$0.3462          & \multirow{4}{*}{$+$} & 0.5649$\pm$0.0484          & \multirow{4}{*}{$+$} & \textbf{0.3509}$\pm$\textbf{0.1047} \\
			& 50  & 0.1535$\pm$0.0250 &     & 0.4613$\pm$0.0497 &     & 0.3789$\pm$0.0709 &     & 0.4450$\pm$0.0628 &     & 0.1940$\pm$0.0548          &     & 0.2353$\pm$0.0318          &     & \textbf{0.1299}$\pm$\textbf{0.0182} \\
			& 75  & 0.1234$\pm$0.0176 &     & 0.4297$\pm$0.0447 &     & 0.3499$\pm$0.0710 &     & 0.3743$\pm$0.0632 &     & 0.1698$\pm$0.0304          &     & 0.2218$\pm$0.0274          &     & \textbf{0.0877}$\pm$\textbf{0.0069} \\
			& 100 & 0.0996$\pm$0.0130 &     & 0.4127$\pm$0.0467 &     & 0.3399$\pm$0.0737 &     & 0.3267$\pm$0.0550 &     & 0.1615$\pm$0.0287          &     & 0.2155$\pm$0.0292          &     & \textbf{0.0716}$\pm$\textbf{0.0055} \\
			\hline
			\multirow{4}{*}{mDTLZ2-($1,0$)}  & 25  & 0.3853$\pm$0.0272 & \multirow{4}{*}{$+$} & 0.4488$\pm$0.0126 & \multirow{4}{*}{$+$} & 0.4436$\pm$0.0198 & \multirow{4}{*}{$+$} & 0.4490$\pm$0.0105 & \multirow{4}{*}{$+$} & 0.3964$\pm$0.0327          & \multirow{4}{*}{$+$} & 0.4222$\pm$0.0300          & \multirow{4}{*}{$+$} & \textbf{0.3469}$\pm$\textbf{0.0319} \\
			& 50  & 0.2601$\pm$0.0313 &     & 0.4122$\pm$0.0287 &     & 0.3953$\pm$0.0322 &     & 0.3847$\pm$0.0297 &     & 0.3487$\pm$0.0398          &     & 0.2282$\pm$0.0100          &     & \textbf{0.1587}$\pm$\textbf{0.0171} \\
			& 75  & 0.2192$\pm$0.0272 &     & 0.3863$\pm$0.0297 &     & 0.3735$\pm$0.0337 &     & 0.3472$\pm$0.0281 &     & 0.3134$\pm$0.0286          &     & 0.2140$\pm$0.0056          &     & \textbf{0.1280}$\pm$\textbf{0.0127} \\
			& 100 & 0.1889$\pm$0.0156 &     & 0.3717$\pm$0.0230 &     & 0.3612$\pm$0.0357 &     & 0.3215$\pm$0.0306 &     & 0.2851$\pm$0.0346          &     & 0.2077$\pm$0.0067          &     & \textbf{0.1146}$\pm$\textbf{0.0103} \\
			\hline
			\multirow{4}{*}{mDTLZ3-($1,0$)}  & 25  & 0.6104$\pm$0.0750 & \multirow{4}{*}{$+$} & 0.6363$\pm$0.0066 & \multirow{4}{*}{$+$} & 0.6240$\pm$0.0356 & \multirow{4}{*}{$+$} & 0.6356$\pm$0.0050 & \multirow{4}{*}{$+$} & 0.7290$\pm$0.1423          & \multirow{4}{*}{$+$} & 0.7006$\pm$0.0181          & \multirow{4}{*}{$+$} & \textbf{0.5150}$\pm$\textbf{0.0928} \\
			& 50  & 0.3923$\pm$0.0482 &     & 0.6253$\pm$0.0195 &     & 0.5887$\pm$0.0458 &     & 0.5868$\pm$0.0474 &     & 0.6379$\pm$0.1501          &     & 0.4342$\pm$0.0852          &     & \textbf{0.2496}$\pm$\textbf{0.0373} \\
			& 75  & 0.3113$\pm$0.0418 &     & 0.6158$\pm$0.0281 &     & 0.5666$\pm$0.0546 &     & 0.5310$\pm$0.0497 &     & 0.4867$\pm$0.1276          &     & 0.3556$\pm$0.0554          &     & \textbf{0.1939}$\pm$\textbf{0.0265} \\
			& 100 & 0.2689$\pm$0.0348 &     & 0.6054$\pm$0.0321 &     & 0.5519$\pm$0.0571 &     & 0.4818$\pm$0.0595 &     & 0.3894$\pm$0.0717          &     & 0.3287$\pm$0.0625          &     & \textbf{0.1647}$\pm$\textbf{0.0206} \\
			\hline
			\multirow{4}{*}{mDTLZ4-($1,0$)}  & 25  & 0.4711$\pm$0.0281 & \multirow{4}{*}{$+$} & 0.5043$\pm$0.0332 & \multirow{4}{*}{$+$} & 0.4980$\pm$0.0326 & \multirow{4}{*}{$+$} & 0.5247$\pm$0.0203 & \multirow{4}{*}{$+$} & 0.4168$\pm$0.0485          & \multirow{4}{*}{$+$} & 0.4943$\pm$0.0201          & \multirow{4}{*}{$+$} & \textbf{0.4039}$\pm$\textbf{0.0514} \\
			& 50  & 0.3689$\pm$0.0480 &     & 0.4649$\pm$0.0400 &     & 0.4360$\pm$0.0421 &     & 0.4388$\pm$0.0367 &     & 0.3768$\pm$0.0692          &     & 0.4099$\pm$0.0346          &     & \textbf{0.2028}$\pm$\textbf{0.0252} \\
			& 75  & 0.2931$\pm$0.0568 &     & 0.4457$\pm$0.0367 &     & 0.3968$\pm$0.0512 &     & 0.3755$\pm$0.0383 &     & 0.3499$\pm$0.0765          &     & 0.3889$\pm$0.0385          &     & \textbf{0.1469}$\pm$\textbf{0.0162} \\
			& 100 & 0.2527$\pm$0.0490 &     & 0.4257$\pm$0.0359 &     & 0.3720$\pm$0.0528 &     & 0.3437$\pm$0.0336 &     & 0.3112$\pm$0.0656          &     & 0.3323$\pm$0.0750          &     & \textbf{0.1211}$\pm$\textbf{0.0129} \\
			\hline
			\multirow{4}{*}{mDTLZ1$^{-1}$-($1,0$)} & 25  & 0.3678$\pm$0.2043 & \multirow{4}{*}{$+$} & 1.1042$\pm$0.0846 & \multirow{4}{*}{$+$} & 0.9881$\pm$0.2087 & \multirow{4}{*}{$+$} & 1.1046$\pm$0.0863 & \multirow{4}{*}{$+$} & 0.7316$\pm$0.2295          & \multirow{4}{*}{$+$} & 0.6991$\pm$0.2237          & \multirow{4}{*}{$+$} & \textbf{0.3083}$\pm$\textbf{0.0630} \\
			& 50  & 0.1416$\pm$0.0280 &     & 0.7501$\pm$0.2540 &     & 0.8001$\pm$0.2586 &     & 0.7413$\pm$0.2284 &     & 0.1753$\pm$0.0194          &     & 0.1848$\pm$0.0299          &     & \textbf{0.1332}$\pm$\textbf{0.0274} \\
			& 75  & 0.1040$\pm$0.0187 &     & 0.6604$\pm$0.2067 &     & 0.6328$\pm$0.2150 &     & 0.5623$\pm$0.1763 &     & 0.1282$\pm$0.0265          &     & 0.1400$\pm$0.0295          &     & \textbf{0.0841}$\pm$\textbf{0.0129} \\
			& 100 & 0.0767$\pm$0.0107 &     & 0.6578$\pm$0.2079 &     & 0.5633$\pm$0.1905 &     & 0.4549$\pm$0.1150 &     & 0.1030$\pm$0.0316          &     & 0.1023$\pm$0.0254          &     & \textbf{0.0680}$\pm$\textbf{0.0132} \\
			\hline
			\multirow{4}{*}{mDTLZ2$^{-1}$-($1,0$)} & 25  & 0.3104$\pm$0.0356 & \multirow{4}{*}{$+$} & 0.3818$\pm$0.0193 & \multirow{4}{*}{$+$} & 0.3992$\pm$0.0094 & \multirow{4}{*}{$+$} & 0.3860$\pm$0.0176 & \multirow{4}{*}{$+$} & 0.2785$\pm$0.0206          & \multirow{4}{*}{$-$} & \textbf{0.2782}$\pm$\textbf{0.0226} & \multirow{4}{*}{$-$} & 0.2918$\pm$0.0293          \\
			& 50  & 0.1617$\pm$0.0118 &     & 0.3203$\pm$0.0278 &     & 0.3402$\pm$0.0331 &     & 0.3046$\pm$0.0284 &     & 0.1448$\pm$0.0103          &     & \textbf{0.1347}$\pm$\textbf{0.0093} &     & 0.1476$\pm$0.0142          \\
			& 75  & 0.1254$\pm$0.0115 &     & 0.2840$\pm$0.0267 &     & 0.2952$\pm$0.0475 &     & 0.2611$\pm$0.0299 &     & 0.1125$\pm$0.0099          &     & \textbf{0.0829}$\pm$\textbf{0.0024} &     & 0.1171$\pm$0.0082          \\
			& 100 & 0.1071$\pm$0.0093 &     & 0.2716$\pm$0.0195 &     & 0.2693$\pm$0.0489 &     & 0.2306$\pm$0.0305 &     & 0.0976$\pm$0.0053         &     & \textbf{0.0641}$\pm$\textbf{0.0029} &     & 0.1040$\pm$0.0062          \\
			\hline
			\multirow{4}{*}{mDTLZ3$^{-1}$-($1,0$)} & 25  & 0.6104$\pm$0.0750 & \multirow{4}{*}{$+$} & 0.6634$\pm$0.0431 & \multirow{4}{*}{$+$} & 0.6617$\pm$0.0294 & \multirow{4}{*}{$+$} & 0.6625$\pm$0.0596 & \multirow{4}{*}{$+$} & \textbf{0.4191$\pm$0.0841}          & \multirow{4}{*}{$-$} & {0.5150}$\pm${0.0674}          & \multirow{4}{*}{$-$} & {0.4637}$\pm${0.0636} \\
			& 50  & 0.3923$\pm$0.0482 &     & 0.5307$\pm$0.0784 &     & 0.5773$\pm$0.0515 &     & 0.5143$\pm$0.0852 &     & \textbf{0.1885}$\pm$\textbf{0.0323} &     & 0.2116$\pm$0.0338          &     & 0.1958$\pm$0.0178          \\
			& 75  & 0.3113$\pm$0.0418 &     & 0.4909$\pm$0.0718 &     & 0.5233$\pm$0.0620 &     & 0.4221$\pm$0.0741 &     & 0.1120$\pm$0.0108          &     & \textbf{0.1116}$\pm$\textbf{0.0109} &     & 0.1317$\pm$0.0160          \\
			& 100 & 0.2689$\pm$0.0348 &     & 0.4779$\pm$0.0748 &     & 0.4916$\pm$0.0729 &     & 0.3688$\pm$0.0511 &     & 0.0941$\pm$0.0066          &     & \textbf{0.0779}$\pm$\textbf{0.0066} &     & 0.1031$\pm$0.0072         \\
			\hline
			\multirow{4}{*}{mDTLZ4$^{-1}$-($1,0$)} & 25  & 0.3358$\pm$0.0269 & \multirow{4}{*}{$+$} & 0.4261$\pm$0.0214 & \multirow{4}{*}{$+$} & 0.4399$\pm$0.0158 & \multirow{4}{*}{$+$} & 0.4314$\pm$0.0168 & \multirow{4}{*}{$+$} & 0.3678$\pm$0.0358          & \multirow{4}{*}{$-$} & \textbf{0.2775}$\pm$\textbf{0.0352} & \multirow{4}{*}{$-$} & 0.3132$\pm$0.0348          \\
			& 50  & 0.2691$\pm$0.0362 &     & 0.3650$\pm$0.0275 &     & 0.3606$\pm$0.0427 &     & 0.3464$\pm$0.0349 &     & 0.1835$\pm$0.0184          &     & \textbf{0.1279}$\pm$\textbf{0.0096} &     & 0.1812$\pm$0.0236          \\
			& 75  & 0.2016$\pm$0.0225 &     & 0.3370$\pm$0.0317 &     & 0.3234$\pm$0.0454 &     & 0.3030$\pm$0.0371 &     & 0.1254$\pm$0.0071          &     & \textbf{0.0895}$\pm$\textbf{0.0050} &     & 0.1365$\pm$0.0145          \\
			& 100 & 0.1727$\pm$0.0246 &     & 0.3150$\pm$0.0405 &     & 0.2853$\pm$0.0498 &     & 0.2588$\pm$0.0348 &     & 0.1015$\pm$0.0040          &     & \textbf{0.0713}$\pm$\textbf{0.0035} &     & 0.1154$\pm$0.0112      \\   
			\hline
			\multicolumn{2}{c|}{$+/-/=$} & \multicolumn{2}{c|}{$8/0/0$}  & \multicolumn{2}{c|}{$8/0/0$} & \multicolumn{2}{c|}{$8/0/0$} & \multicolumn{2}{c|}{$8/0/0$} & \multicolumn{2}{c|}{$5/0/3$} & \multicolumn{2}{c|}{$5/0/3$} &  \multicolumn{1}{c}{} \\
			\hline
	\end{tabular}}
\end{table*}

Table~\ref{Tab:Baseline} indicates that the invTrEMO achieves faster convergence compared to its competitors on a majority of the benchmark problems. The table shows results of the Kruskal-Wallis test at a 0.05 significance level to confirm the statistical significance of the outcomes, where ``$+$", ``$-$", and ``$\approx$" denote that the invTrEMO performs better than, worse than, or similar to a competitor. It is noteworthy that the invTrEMO significantly outperforms ParEGO-UCB, MOEA/D-EGO, K-RVEA, and CSEA across all eight target MOPs, and beats qNEHVI and PSL-MOBO on five target problems. qNEHVI and the recent PSL-MOBO only outperform invTrEMO on the last three benchmark problems with inverted triangular PF shapes, i.e., mDTLZ2$^{-1}$-($1,0$), mDTLZ3$^{-1}$-($1,0$), and mDTLZ4$^{-1}$-($1,0$). The above results support the claim that our inverse transfer mechanism can enhance convergence behavior on a wide variety of problems. On problems with inverted PFs, the predefined static preference vectors may not be ideal, and hence the performance of invTrEMO could be further improved in the future with more adaptive objective function weighting schemes. This explains the better performance of qNEHVI and PSL-MOBO over the current version of the invTrEMO on a small subset of problems. The convergence trends in Fig.~\ref{Fig:Baseline} graphically reinforce the conclusions drawn from Table~\ref{Tab:Baseline}. Figs.~S-5 to S-8 of the supplementary file also show the points in objective space corresponding to the nondominated solutions returned by ParEGO-UCB, MOEA/D-EGO, K-RVEA, CSEA, qNEHVI, PSL-MOBO, and the invTrEMO on mDTLZ1-($1,0$) to mDTLZ4-($1,0$) after 100 evaluations. Compared to its competitors, the invTrEMO is found to return solutions that offer better convergence and PF coverage on these benchmark problems.

\subsection{Creation of High-precision Inverse Models}
A major byproduct of the invTrEMO is its ability to support DMs in generating nondominated solutions with desired trade-off preferences on-demand, through the use of inverse models. In this subsection, we demonstrate that by utilizing the source data, our invTGP models can provide more accurate predictions, thus facilitating better multiobjective decision-making. We compare the RMSE results obtained by models from the invTrEMO with the ones below.
\begin{itemize}
	\item PSL-MOBO: Inverse models are directly provided by the PSL-MOBO algorithm.
	\item Post-hoc:  In this method, we first obtain a set of nondominated solutions using ParEGO-UCB, forming a target dataset. Then, inverse GP models are built \emph{offline} based on this target dataset.
	\item Post-hoc-Transfer: Similar to the post-hoc method, we use ParEGO-UCB to find nondominated solutions and create a target dataset. However, in this approach, we build invTGP models \emph{offline} for the overlapping decision variables based on both the target dataset and the MS source dataset \cite{tan2023pareto}.
\end{itemize}
The RMSE results are listed in Table~\ref{Tab:Results}. The Wilcoxon's rank-sum test at a 0.05 significance level was applied to confirm the statistical significance of the results, where ``$+$", ``$-$", and ``$\approx$" denote that the invTrEMO performs better than, worse than, or similar to a competitor.

\begin{table*}[]
	\centering
	\caption{Inverse model RMSE results of PSL-MOBO, Post-hoc, Post-hoc-Transfer (with MS source dataset), and the invTrEMO (with MS source dataset) over 20 independent runs. The Wilcoxon's
		rank-sum test at a 0.05 significance level was applied to determine whether the difference between
		the performances of the algorithms was significant.}\label{Tab:Results}
	\resizebox{12cm}{!}{\begin{tabular}{c|c|c|c|c|c|c|c}
			\hline
			\multirow{2}{*}{Target Tasks}   & \multicolumn{2}{c|}{PSL-MOBO} & \multicolumn{2}{c|}{Post-hoc} & \multicolumn{2}{c|}{Post-hoc-Transfer} & invTrEMO \\
			\cline{2-8}
			& \multicolumn{2}{c|}{Average RMSE $\pm$ Std} & \multicolumn{2}{c|}{Average RMSE $\pm$ Std} & \multicolumn{2}{c|}{Average RMSE $\pm$ Std} & \multicolumn{1}{c}{Average RMSE $\pm$ Std} \\
			\hline
			mDTLZ1-($1,0$) & 0.5655$\pm$0.1131 & $+$ & 0.1223$\pm$0.0578 & $+$ & 0.1171$\pm$0.0684 & $+$ & \textbf{0.0424$\pm$0.0144} \\
			mDTLZ2-($1,0$) & 0.3347$\pm$0.0611 & $+$ & 0.0559$\pm$0.0103 & $+$ & 0.0492$\pm$0.0101 & $+$ & \textbf{0.0335$\pm$0.0109} \\
			mDTLZ3-($1,0$) & 0.5308$\pm$0.1143 & $+$ & 0.1443$\pm$0.0782 & $+$ & 0.1274$\pm$0.0673 & $+$ & \textbf{0.0686$\pm$0.0211} \\
			mDTLZ4-($1,0$) & 0.4942$\pm$0.0587 & $+$ & 0.1181$\pm$0.0389 & $+$ & 0.0929$\pm$0.0302 & $+$ & \textbf{0.0364$\pm$0.0126} \\  
			mDTLZ1$^{-1}$-($1,0$) & 0.2137$\pm$0.0232 & $+$ & 0.0826$\pm$0.0209 & $+$ & 0.0808$\pm$0.0173 & $\approx$ & \textbf{0.0798}$\pm$\textbf{0.0267} \\
			mDTLZ2$^{-1}$-($1,0$) & 0.6078$\pm$0.0389 & $+$ & 0.0508$\pm$0.0062 & $+$ & 0.0573$\pm$0.0068 & $+$       & \textbf{0.0472}$\pm$\textbf{0.0121} \\
			mDTLZ3$^{-1}$-($1,0$) & 0.6105$\pm$0.0108 & $+$ & 0.0799$\pm$0.0234 & $+$ & 0.0741$\pm$0.0218 & $+$       & \textbf{0.0451}$\pm$\textbf{0.0103} \\
			mDTLZ4$^{-1}$-($1,0$) & 0.5914$\pm$0.0179 & $+$ & 0.0903$\pm$0.0144 & $+$ & 0.1057$\pm$0.0189 & $+$       & \textbf{0.0603}$\pm$\textbf{0.0112} \\
			\hline
			$+/-/=$ & \multicolumn{2}{c|}{$8/0/0$} & \multicolumn{2}{c|}{$8/0/0$} & \multicolumn{2}{c|}{$7/0/1$} &  \\
			\hline
	\end{tabular}}
\end{table*}

Table~\ref{Tab:Results} clearly indicates that, with the assistance of the MS source dataset, the inverse models from the invTrEMO consistently achieve the best RMSE results. While the Post-hoc-Transfer method also utilizes an external MS dataset for building the inverse models, the absence of knowledge transfer during the online optimization process often leads to a lower-quality, suboptimal target dataset compared to the invTrEMO. This difference in dataset quality explains why the invTrEMO may lead to superior RMSE results. According to the Wilcoxon's rank-sum test, the invTrEMO outperforms PSL-MOBO, Post-hoc, and Post-hoc-Transfer on eight, eight, and seven target tasks, respectively, strengthening claims of its effectiveness.

\subsection{Influence of Source-Target Correlations}\label{subsec:source-target corr}
\begin{figure}[]
	\begin{center}
		\subfigure[]{\label{fig:similarity_1}\includegraphics[width=0.35\columnwidth]{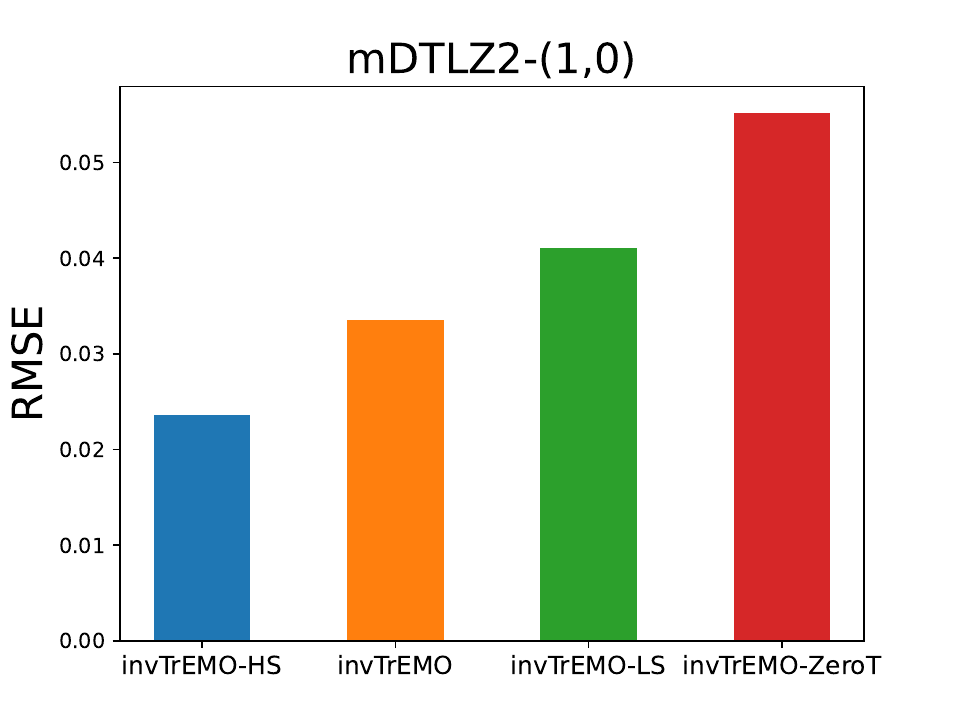}}
		\subfigure[]{\label{fig:similarity_2}\includegraphics[width=0.35\columnwidth]{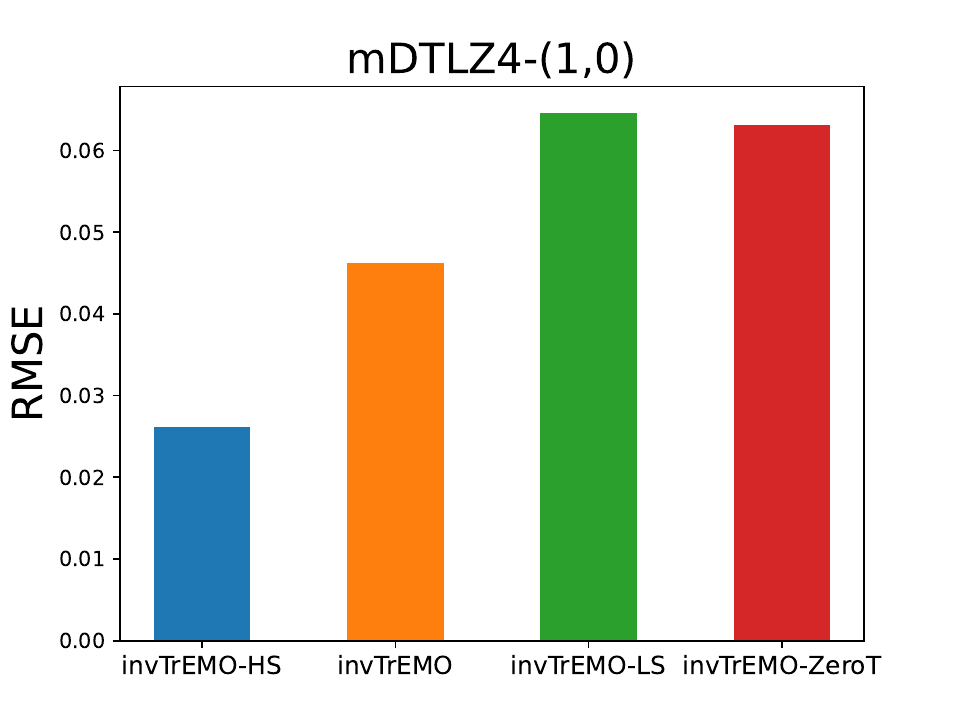}}
		\caption{Comparisons of RMSE results provided by the inverse models of invTrEMO-HS, invTrEMO, invTrEMO-LS, and invTrEMO-ZeroT on mDTLZ2-($1,0$) and mDTLZ4-($1,0$). invTrEMO-HS, invTrEMO, and invTrEMO-LS indicate that the HS, MS, or LS source datasets are used, respectively. Additionally, no source MOP data is employed in invTrEMO-ZeroT. (a) mDTLZ2-($1,0$). (b) mDTLZ4-($1,0$).}\label{Fig:Similarity}
	\end{center}
\end{figure}

Here, we explore the impact of source-target correlation on the invTrEMO's performance using mDTLZ2-($1,0$) and mDTLZ4-($1,0$) as examples. We employ three source datasets with varying levels of source-target correlation (i.e., HS, MS, and LS) to assist the algorithm, resulting in invTrEMO-HS, invTrEMO, and invTrEMO-LS, respectively. Additionally, we provide the results of a no-transfer version of invTrEMO, denoted as invTrEMO-ZeroT, for comparison. In invTrEMO-ZeroT, we replace the invTGP models of the original invTrEMO with standard inverse GP models (without transfer learning). Table~\ref{Tab:SimilarityIGD} presents the averaged IGD values achieved by the invTrEMO-HS, invTrEMO, invTrEMO-LS, and invTrEMO-ZeroT after 25, 50, 75, and 100 evaluations. Furthermore, the RMSE results of the inverse models derived are depicted in Fig.~\ref{Fig:Similarity}.

\begin{table*}[]
	\centering
	\caption{IGD results of invTrEMO-HS, invTrEMO, invTrEMO-LS, and invTrEMO-ZeroT after 25, 50, 75, and 100 evaluations on mDTLZ2-($1,0$) and mDTLZ4-($1,0$), averaged over 20 independent runs.}\label{Tab:SimilarityIGD}
	\resizebox{12cm}{!}{\begin{tabular}{c|c|c|c|c|ccc}
			\hline
			\multirow{2}{*}{Target Tasks} & \multirow{2}{*}{Evaluations} & \multicolumn{1}{c|}{invTrEMO-HS} & \multicolumn{1}{c|}{invTrEMO} & \multicolumn{1}{c|}{invTrEMO-LS} & invTrEMO-ZeroT \\
			\cline{3-6}
			& & \multicolumn{1}{c|}{Average IGD $\pm$ Std} & \multicolumn{1}{c|}{Average IGD $\pm$ Std} & \multicolumn{1}{c|}{Average RMSE $\pm$ Std} & \multicolumn{1}{c}{Average RMSE $\pm$ Std} \\
			\hline
			\multirow{4}{*}{mDTLZ2-($1,0$)} & 25  &  \textbf{0.3359}$\pm$\textbf{0.0360} & 0.3526$\pm$0.0332 & 0.3648$\pm$0.0250 & 0.3597$\pm$0.0256 \\
			& 50  &  \textbf{0.1562}$\pm$\textbf{0.0185} & 0.1569$\pm$0.0129 & 0.1666$\pm$0.0142 & 0.1649$\pm$0.0123 \\
			& 75  &  \textbf{0.1212}$\pm$\textbf{0.0089} & 0.1261$\pm$0.0100 & 0.1301$\pm$0.0109 & 0.1376$\pm$0.0114 \\
			& 100 &  \textbf{0.1069}$\pm$\textbf{0.0076} & 0.1139$\pm$0.0095 & 0.1175$\pm$0.0110 & 0.1219$\pm$0.0108 \\
			\hline
			\multirow{4}{*}{mDTLZ4-($1,0$)} & 25  &  \textbf{0.4008}$\pm$\textbf{0.0438} & 0.4039$\pm$0.0514 & 0.4383$\pm$0.0409 & 0.4425$\pm$0.0339 \\
			& 50  &  \textbf{0.1974}$\pm$\textbf{0.0393} & 0.2028$\pm$0.0252 & 0.2102$\pm$0.0357 & 0.2209$\pm$0.0428 \\
			& 75  &  \textbf{0.1423}$\pm$\textbf{0.0241} & 0.1469$\pm$0.0162 & 0.1566$\pm$0.0235 & 0.1568$\pm$0.0211 \\
			& 100 &  \textbf{0.1203}$\pm$\textbf{0.0205} & 0.1211$\pm$0.0129 & 0.1347$\pm$0.0207 & 0.1326$\pm$0.0214 \\
			\hline
	\end{tabular}}
\end{table*}

Perhaps unsurprisingly, Table~\ref{Tab:SimilarityIGD} demonstrates that invTrEMO-HS converges more rapidly than invTrEMO, which in turn exhibits better convergence than invTrEMO-LS. This suggests that higher levels of source-target correlation tend to lead to faster convergence. However, it is worth noting that invTrEMO-LS does not achieve faster convergence than invTrEMO-ZeroT on mDTLZ4-($1,0$). This points to the possibility that low source-target correlations could in some cases result in negative transfers, combating which is an important direction for continued research to prevent deleterious effects on the optimization process.

With regard to the accuracy of the inverse models, Fig.~\ref{Fig:Similarity} reveals that invTrEMO-HS attains better RMSE results compared to invTrEMO, which in turn outperforms invTrEMO-LS. These findings indicate that higher source-target similarities also contribute to improved inverse modeling performance. Moreover, once again, negative transfers could potentially lead to poor accuracy of the inverse models, as  suggested by the fact that the RMSE of the the inverse model drawn from invTrEMO-LS is inferior to that of invTrEMO-ZeroT on mDTLZ4-($1,0$).

\subsection{Influence of the Number of Overlapping Decision Variables}
\begin{figure}[]
	\begin{center}
		\subfigure[]{\label{fig:dimension_RMSE_1}\includegraphics[width=0.24\columnwidth]{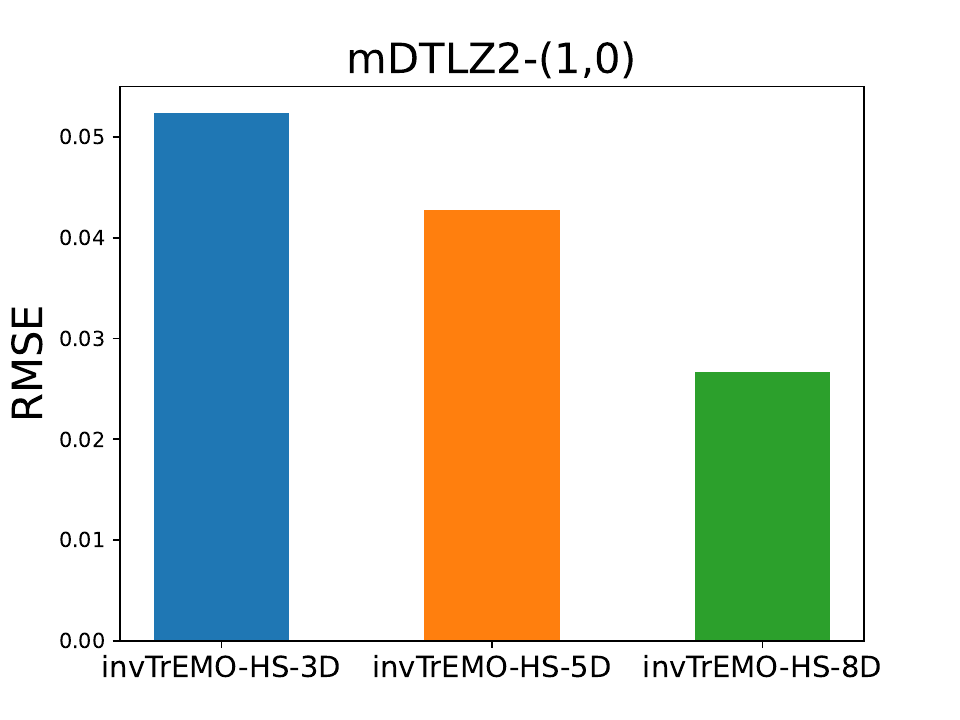}}
		\subfigure[]{\label{fig:dimension_RMSE_2}\includegraphics[width=0.24\columnwidth]{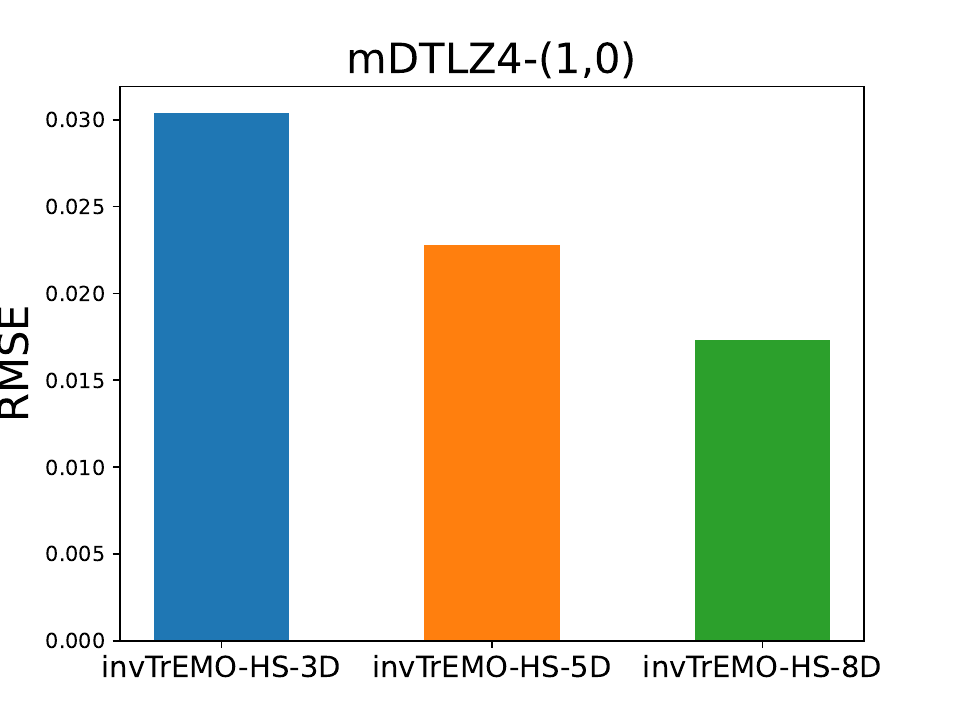}}
		\subfigure[]{\label{fig:dimension_1}\includegraphics[width=0.24\columnwidth]{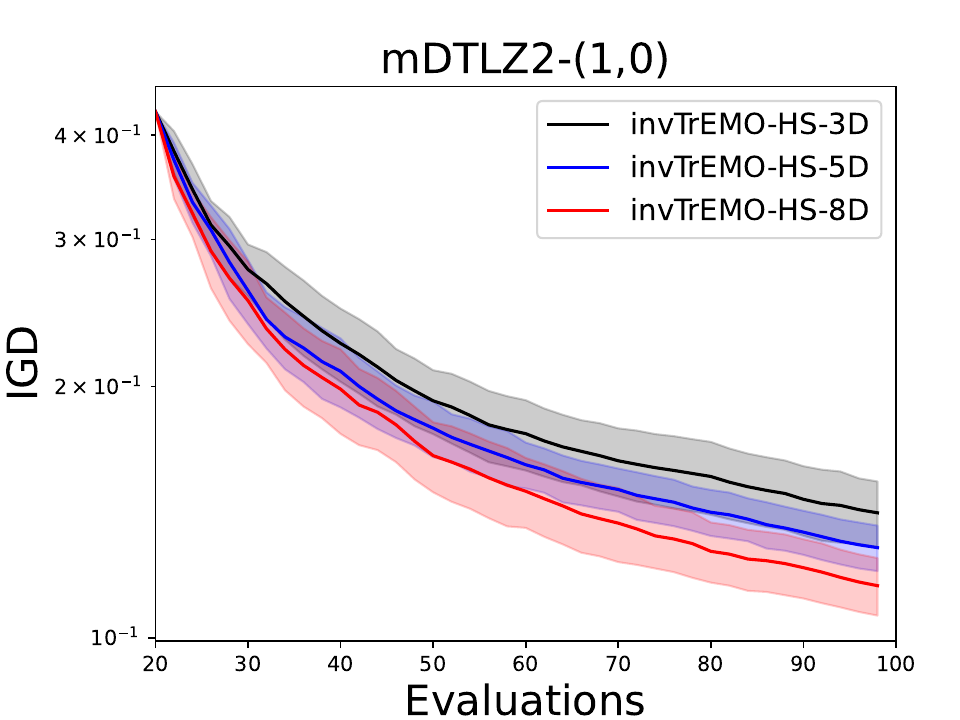}}
		\subfigure[]{\label{fig:dimension_2}\includegraphics[width=0.24\columnwidth]{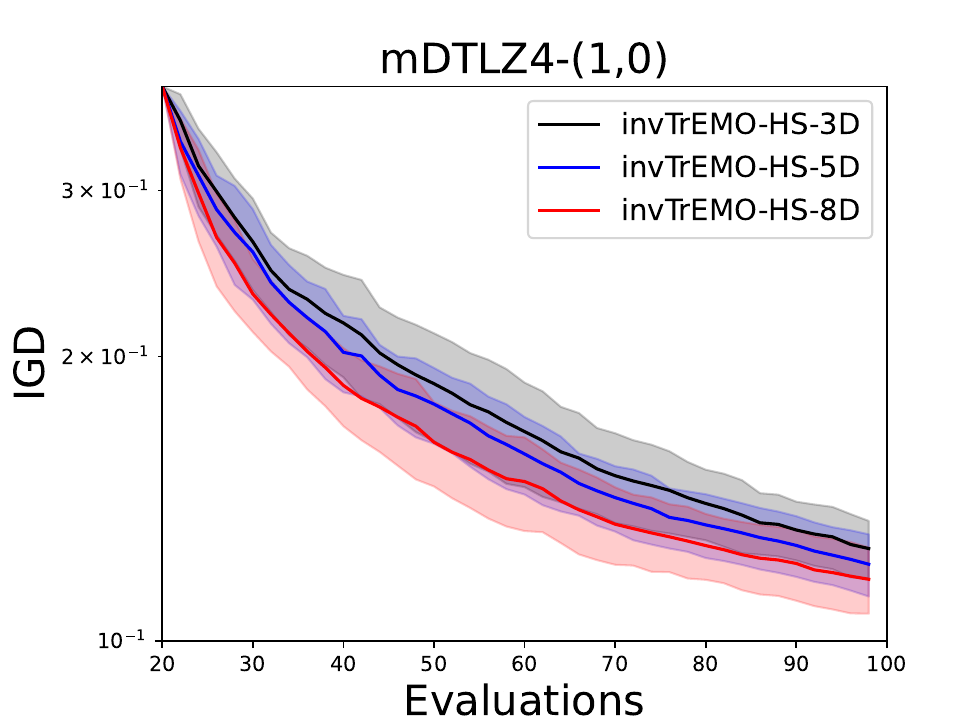}}
		\caption{RMSE results and IGD convergence trends of invTrEMO with different numbers of overlapping decision variables. invTrEMO-HS-3D, invTrEMO-HS-5D, and invTrEMO-HS-8D denote that the decision variables of the source tasks overlap with the first 3, 5, or 8 decision variables of the target tasks, respectively. The shaded areas represent one standard deviation in performance on either side of the mean. (a) RMSE results on mDTLZ2-($1,0$). (b) RMSE results on mDTLZ4-($1,0$). (c) IGD convergence trends on mDTLZ2-($1,0$). (d) IGD convergence trends on mDTLZ4-($1,0$).}\label{Fig:dimension}
	\end{center}
\end{figure}

As mentioned in Section~\ref{Section3}, the inverse transfer mechanism only operates on the  overlapping source and target decision variables. Hence, it is of interest to investigate the influence of the number of overlapping decision variables on the target optimization performance. To explore this, we generate source datasets for mDTLZ2-$(1,0)$ and mDTLZ4-$(1,0)$ based on the corresponding HS tasks, and with 3, 5, and 8 decision variables. These decision variables overlap with the first 3, 5, and 8 decision variables of the target, respectively. The results are recorded as invTrEMO-HS-3D, invTrEMO-HS-5D, and invTrEMO-HS-8D. The RMSE results and IGD convergence trends are presented in Fig.\ref{Fig:dimension}. From Fig.\ref{fig:dimension_RMSE_1} and Fig.\ref{fig:dimension_RMSE_2}, it can be observed that having more overlapping decision variables leads to better RMSE results. This is intuitively pleasing. Similarly, Fig.\ref{fig:dimension_1} and Fig.~\ref{fig:dimension_2} show that more overlapping decision variables also result in relatively faster convergence. These findings can be understood by the fact that when there are more overlapping decision variables, more information from the source is utilized to enhance performance in the target tasks.

\begin{table*}[]
	\centering
	\caption{IGD results of ParEGO-UCB, MOEA/D-EGO, K-RVEA, CSEA, PSL-MOBO, and invTrEMO-HS at 25, 50, 75, and 100 evaluations over 20 independent runs on target tasks with 5 Objectives. The Kruskal-Wallis test at a 0.05 significance level was performed to compare invTrEMO-HS with the other algorithms.}\label{Tab:igd_10obj}
	\resizebox{14cm}{!}{\begin{tabular}{c|c|c|c|c|c|c|c|c|c|c|c|c}
			\hline
			\multirow{2}{*}{Target Tasks} & \multirow{2}{*}{Evaluations} & \multicolumn{2}{c|}{ParEGO-UCB} & \multicolumn{2}{c|}{MOEA/D-EGO} & \multicolumn{2}{c|}{K-RVEA} & \multicolumn{2}{c|}{CSEA} & \multicolumn{2}{c|}{PSL-MOBO} & \multicolumn{1}{c}{invTrEMO-HS}      \\
			\cline{3-13}
			&  & \multicolumn{2}{c|}{Average IGD $\pm$ Std}  & \multicolumn{2}{c|}{Average IGD $\pm$ Std} & \multicolumn{2}{c|}{Average IGD $\pm$ Std} & \multicolumn{2}{c|}{Average IGD $\pm$ Std} & \multicolumn{2}{c|}{Average IGD $\pm$ Std} & \multicolumn{1}{c}{Average IGD$\pm$ Std}     \\
			\hline
			\multirow{4}{*}{mDTLZ1-($1,0$)} & 25  & 0.3858$\pm$0.1333 & \multirow{4}{*}{$+$} & 0.8321$\pm$0.0293 & \multirow{4}{*}{$+$} & 0.7772$\pm$0.1314 & \multirow{4}{*}{$+$} & 0.8166$\pm$0.0530 & \multirow{4}{*}{$+$} & 0.5738$\pm$0.0494 & \multirow{4}{*}{$+$} & \textbf{0.3323}$\pm$\textbf{0.1240} \\
			& 50  & 0.1535$\pm$0.0250 &     & 0.7817$\pm$0.1034 &     & 0.5964$\pm$0.1962 &     & 0.6905$\pm$0.1369 &     & 0.2300$\pm$0.0304 &     & \textbf{0.1228}$\pm$\textbf{0.0211} \\
			& 75  & 0.1234$\pm$0.0176 &     & 0.7631$\pm$0.1092 &     & 0.5706$\pm$0.1879 &     & 0.5711$\pm$0.1632 &     & 0.2174$\pm$0.0261 &     & \textbf{0.0821}$\pm$\textbf{0.0078} \\
			& 100 & 0.0996$\pm$0.0130 &     & 0.7569$\pm$0.1193 &     & 0.5679$\pm$0.1853 &     & 0.4963$\pm$0.1679 &     & 0.2112$\pm$0.0275 &     & \textbf{0.0669}$\pm$\textbf{0.0038} \\
			\hline
			\multirow{4}{*}{mDTLZ2-($1,0$)} & 25  & 0.5568$\pm$0.0507 & \multirow{4}{*}{$+$} & 0.5941$\pm$0.0443 & \multirow{4}{*}{$+$} & 0.5748$\pm$0.0234 & \multirow{4}{*}{$+$} & 0.5737$\pm$0.0315 & \multirow{4}{*}{$+$} & 0.5798$\pm$0.0250 & \multirow{4}{*}{$+$} & \textbf{0.5064}$\pm$\textbf{0.0471} \\
			& 50  & 0.4307$\pm$0.0430 &     & 0.5799$\pm$0.0480 &     & 0.5616$\pm$0.0247 &     & 0.5223$\pm$0.0392 &     & 0.4428$\pm$0.0318 &     & \textbf{0.3795}$\pm$\textbf{0.0299} \\
			& 75  & 0.3900$\pm$0.0420 &     & 0.5606$\pm$0.0400 &     & 0.5597$\pm$0.0247 &     & 0.4803$\pm$0.0469 &     & 0.4104$\pm$0.0223 &     & \textbf{0.3542}$\pm$\textbf{0.0262} \\
			& 100 & 0.3669$\pm$0.0382 &     & 0.5481$\pm$0.0482 &     & 0.5566$\pm$0.0270 &     & 0.4573$\pm$0.0522 &     & 0.3951$\pm$0.0259 &     & \textbf{0.3416}$\pm$\textbf{0.0242} \\
			\hline
			\multirow{4}{*}{mDTLZ3-($1,0$)} & 25  & 0.9325$\pm$0.1660 & \multirow{4}{*}{$+$} & 0.9855$\pm$0.0869 & \multirow{4}{*}{$+$} & 1.0189$\pm$0.0865 & \multirow{4}{*}{$+$} & 1.0563$\pm$0.0928 & \multirow{4}{*}{$+$} & 1.0215$\pm$0.1454 & \multirow{4}{*}{$+$} & \textbf{0.6793}$\pm$\textbf{0.0924} \\
			& 50  & 0.6331$\pm$0.0895 &     & 0.9366$\pm$0.0762 &     & 0.9797$\pm$0.1184 &     & 0.9045$\pm$0.1097 &     & 0.6780$\pm$0.1296 &     & \textbf{0.4843}$\pm$\textbf{0.0640} \\
			& 75  & 0.5784$\pm$0.0827 &     & 0.9066$\pm$0.0631 &     & 0.9659$\pm$0.1349 &     & 0.8218$\pm$0.1044 &     & 0.5704$\pm$0.0599 &     & \textbf{0.4196}$\pm$\textbf{0.0554} \\
			& 100 & 0.5465$\pm$0.0738 &     & 0.8960$\pm$0.0651 &     & 0.9524$\pm$0.1289 &     & 0.7409$\pm$0.0717 &     & 0.5217$\pm$0.0390 &     & \textbf{0.3751}$\pm$\textbf{0.0514} \\
			\hline
			\multirow{4}{*}{mDTLZ4-($1,0$)} & 25  & 0.6275$\pm$0.0792 & \multirow{4}{*}{$+$} & 0.6303$\pm$0.0484 & \multirow{4}{*}{$+$} & 0.6299$\pm$0.0286 & \multirow{4}{*}{$+$} & 0.6216$\pm$0.0523 & \multirow{4}{*}{$+$} & 0.6160$\pm$0.0312 & \multirow{4}{*}{$+$} & \textbf{0.5574}$\pm$\textbf{0.0496} \\
			& 50  & 0.4974$\pm$0.0550 &     & 0.6023$\pm$0.0476 &     & 0.5996$\pm$0.0386 &     & 0.5781$\pm$0.0474 &     & 0.4873$\pm$0.0452 &     & \textbf{0.4001}$\pm$\textbf{0.0492} \\
			& 75  & 0.4540$\pm$0.0457 &     & 0.5956$\pm$0.0460 &     & 0.5963$\pm$0.0392 &     & 0.5473$\pm$0.0473 &     & 0.4503$\pm$0.0410 &     & \textbf{0.3539}$\pm$\textbf{0.0342} \\
			& 100 & 0.4237$\pm$0.0395 &     & 0.5810$\pm$0.0418 &     & 0.5857$\pm$0.0480 &     & 0.5105$\pm$0.0408 &     & 0.4365$\pm$0.0386 &     & \textbf{0.3306}$\pm$\textbf{0.0299} \\
			\hline
			\multicolumn{2}{c|}{$+/-/=$} & \multicolumn{2}{c|}{$4/0/0$}  & \multicolumn{2}{c|}{$4/0/0$} & \multicolumn{2}{c|}{$4/0/0$} & \multicolumn{2}{c|}{$4/0/0$} & \multicolumn{2}{c|}{$4/0/0$} &  \multicolumn{1}{c}{} \\
			\hline
	\end{tabular}}
\end{table*}

\begin{table*}[]
	\centering
	\caption{IGD results of ParEGO-UCB, MOEA/D-EGO, K-RVEA, CSEA, PSL-MOBO, and invTrEMO-HS at 25, 50, 75, and 100 evaluations over 20 independent runs on target tasks with 8 Objectives. The Kruskal-Wallis test at a 0.05 significance level was performed to compare invTrEMO-HS with the other algorithms.}\label{Tab:igd_5obj}
	\resizebox{14cm}{!}{\begin{tabular}{c|c|c|c|c|c|c|c|c|c|c|c|c}
			\hline
			\multirow{2}{*}{Target Tasks} & \multirow{2}{*}{Evaluations} & \multicolumn{2}{c|}{ParEGO-UCB} & \multicolumn{2}{c|}{MOEA/D-EGO} & \multicolumn{2}{c|}{K-RVEA} & \multicolumn{2}{c|}{CSEA} & \multicolumn{2}{c|}{PSL-MOBO} & \multicolumn{1}{c}{invTrEMO-HS}      \\
			\cline{3-13}
			&  & \multicolumn{2}{c|}{Average IGD $\pm$ Std}  & \multicolumn{2}{c|}{Average IGD $\pm$ Std} & \multicolumn{2}{c|}{Average IGD $\pm$ Std} & \multicolumn{2}{c|}{Average IGD $\pm$ Std} & \multicolumn{2}{c|}{Average IGD $\pm$ Std} & \multicolumn{1}{c}{Average IGD$\pm$ Std}     \\
			\hline
			\multirow{4}{*}{mDTLZ1-($1,0$)} & 25  & 0.5974$\pm$0.2008 & \multirow{4}{*}{$+$} & 0.6492$\pm$0.2032 & \multirow{4}{*}{$+$} & 0.6725$\pm$0.1993 & \multirow{4}{*}{$+$} & 0.6734$\pm$0.1373 & \multirow{4}{*}{$+$} & 0.6310$\pm$0.1404 & \multirow{4}{*}{$+$} & \textbf{0.4485}$\pm$\textbf{0.1935} \\
			& 50  & 0.1918$\pm$0.0434 &     & 0.5927$\pm$0.2205 &     & 0.6424$\pm$0.2260 &     & 0.5969$\pm$0.1422 &     & 0.3287$\pm$0.0698 &     & \textbf{0.1609}$\pm$\textbf{0.0432} \\
			& 75  & 0.1585$\pm$0.0366 &     & 0.5805$\pm$0.2044 &     & 0.5660$\pm$0.1849 &     & 0.5087$\pm$0.1647 &     & 0.2484$\pm$0.0763 &     & \textbf{0.1165}$\pm$\textbf{0.0178} \\
			& 100 & 0.1345$\pm$0.0302 &     & 0.5135$\pm$0.1432 &     & 0.5384$\pm$0.1816 &     & 0.4657$\pm$0.1666 &     & 0.2210$\pm$0.0772 &     & \textbf{0.1059}$\pm$\textbf{0.0159} \\
			\hline
			\multirow{4}{*}{mDTLZ2-($1,0$)} & 25  & 0.4597$\pm$0.0226 & \multirow{4}{*}{$+$} & 0.4486$\pm$0.0067 & \multirow{4}{*}{$+$} & 0.4453$\pm$0.0186 & \multirow{4}{*}{$+$} & 0.4554$\pm$0.0264 & \multirow{4}{*}{$+$} & 0.4211$\pm$0.0256 & \multirow{4}{*}{$+$} & \textbf{0.4358}$\pm$\textbf{0.0282} \\
			& 50  & 0.4009$\pm$0.0228 &     & 0.4271$\pm$0.0140 &     & 0.4160$\pm$0.0228 &     & 0.4071$\pm$0.0324 &     & 0.3892$\pm$0.0256 &     & \textbf{0.3395}$\pm$\textbf{0.0158} \\
			& 75  & 0.3797$\pm$0.0193 &     & 0.4207$\pm$0.0165 &     & 0.3962$\pm$0.0240 &     & 0.3795$\pm$0.0312 &     & 0.3730$\pm$0.0233 &     & \textbf{0.3021}$\pm$\textbf{0.0178} \\
			& 100 & 0.3654$\pm$0.0186 &     & 0.4097$\pm$0.0168 &     & 0.3796$\pm$0.0270 &     & 0.3609$\pm$0.0302 &     & 0.3639$\pm$0.0209 &     & \textbf{0.2866}$\pm$\textbf{0.0179} \\
			\hline
			\multirow{4}{*}{mDTLZ3-($1,0$)} & 25  & 0.7173$\pm$0.0847 & \multirow{4}{*}{$+$} & 0.6877$\pm$0.0726 & \multirow{4}{*}{$+$} & 0.6975$\pm$0.0756 & \multirow{4}{*}{$+$} & 0.7058$\pm$0.0660 & \multirow{4}{*}{$+$} & 0.6778$\pm$0.0427 & \multirow{4}{*}{$+$} & \textbf{0.6440}$\pm$\textbf{0.0743} \\
			& 50  & 0.6060$\pm$0.0812 &     & 0.6629$\pm$0.0713 &     & 0.5978$\pm$0.0593 &     & 0.6309$\pm$0.0631 &     & 0.5485$\pm$0.0409 &     & \textbf{0.4569}$\pm$\textbf{0.0434} \\
			& 75  & 0.5575$\pm$0.0684 &     & 0.6483$\pm$0.0818 &     & 0.5652$\pm$0.0672 &     & 0.5742$\pm$0.0608 &     & 0.4976$\pm$0.0420 &     & \textbf{0.3996}$\pm$\textbf{0.0331} \\
			& 100 & 0.5322$\pm$0.0593 &     & 0.6411$\pm$0.0770 &     & 0.5102$\pm$0.0421 &     & 0.5163$\pm$0.0574 &     & 0.4738$\pm$0.0550 &     & \textbf{0.3766}$\pm$\textbf{0.0277} \\
			\hline
			\multirow{4}{*}{mDTLZ4-($1,0$)} & 25  & 0.4927$\pm$0.0360 & \multirow{4}{*}{$+$} & 0.5142$\pm$0.0341 & \multirow{4}{*}{$+$} & 0.5181$\pm$0.0441 & \multirow{4}{*}{$+$} & 0.4960$\pm$0.0439 & \multirow{4}{*}{$+$} & 0.4639$\pm$0.0670 & \multirow{4}{*}{$+$} & \textbf{0.4585}$\pm$\textbf{0.0263} \\
			& 50  & 0.4329$\pm$0.0163 &     & 0.4992$\pm$0.0376 &     & 0.4885$\pm$0.0472 &     & 0.4360$\pm$0.0455 &     & 0.3947$\pm$0.0552 &     & \textbf{0.3827}$\pm$\textbf{0.0201} \\
			& 75  & 0.4179$\pm$0.0160 &     & 0.4874$\pm$0.0326 &     & 0.4656$\pm$0.0406 &     & 0.4068$\pm$0.0422 &     & 0.3789$\pm$0.0435 &     & \textbf{0.3498}$\pm$\textbf{0.0165} \\
			& 100 & 0.4085$\pm$0.0208 &     & 0.4807$\pm$0.0291 &     & 0.4573$\pm$0.0452 &     & 0.3898$\pm$0.0403 &     & 0.3766$\pm$0.0417 &     & \textbf{0.3319}$\pm$\textbf{0.0150} \\
			\hline
			\multicolumn{2}{c|}{$+/-/=$} & \multicolumn{2}{c|}{$4/0/0$}  & \multicolumn{2}{c|}{$4/0/0$} & \multicolumn{2}{c|}{$4/0/0$} & \multicolumn{2}{c|}{$4/0/0$} & \multicolumn{2}{c|}{$4/0/0$} &  \multicolumn{1}{c}{} \\
			\hline
	\end{tabular}}
\end{table*}

\begin{table*}[]
	\centering
	\caption{RMSE Results of PSL-MOBO, Post-hoc, Post-hoc-Transfer (with HS source dataset), and invTrEMO-HS on target problems with 5 and 8 Objectives over 20 independent runs. The Wilcoxon's
		rank-sum test at a 0.05 significance level was implemented between invTrEMO-HS and PSL-MOBO, Post-hoc, and Post-hoc-Transfer.}\label{Tab:RMSE_high_dim}
	\resizebox{12cm}{!}{\begin{tabular}{c|c|c|c|c|c|c|c|c}
			\hline
			\multirow{2}{*}{Target Tasks} & \multirow{2}{*}{$m$} & \multicolumn{2}{c|}{PSL-MOBO} & \multicolumn{2}{c|}{Post-hoc} & \multicolumn{2}{c|}{Post-hoc-Transfer} & invTrEMO-HS \\
			\cline{3-9}
			&    & \multicolumn{2}{c|}{Average RMSE $\pm$ Std} & \multicolumn{2}{c|}{Average RMSE $\pm$ Std} & \multicolumn{2}{c|}{Average RMSE $\pm$ Std} & \multicolumn{1}{c}{Average RMSE $\pm$ Std} \\
			\hline
			\multirow{2}{*}{mDTLZ1-($1,0$)} & 5-Obj  & 0.5019$\pm$0.1151 & $+$ & 0.1719$\pm$0.0823 & $+$ & 0.1567$\pm$0.0079 & $+$ & \textbf{0.0791}$\pm$\textbf{0.0307} \\
			& 8-Obj & 0.4801$\pm$0.1861 & $+$ & 0.0817$\pm$0.0363 & $+$ & 0.0724$\pm$0.0339 & $+$ & \textbf{0.0662}$\pm$\textbf{0.0342} \\
			\hline
			\multirow{2}{*}{mDTLZ2-($1,0$)} & 5-Obj  & 0.4466$\pm$0.0433 & $+$ & 0.1131$\pm$0.0112 & $+$ & 0.0805$\pm$0.1058 & $+$ & \textbf{0.0318}$\pm$\textbf{0.0123} \\
			& 8-Obj & 0.3664$\pm$0.0087 & $+$ & 0.0744$\pm$0.0101 & $+$ & 0.0653$\pm$0.0127 & $+$ & \textbf{0.0323}$\pm$\textbf{0.0099} \\
			\hline
			\multirow{2}{*}{mDTLZ3-($1,0$)} & 5-Obj  & 0.5714$\pm$0.0529 & $+$ & 0.2355$\pm$0.0702 & $+$ & 0.1919$\pm$0.0771 & $+$ & \textbf{0.0538}$\pm$\textbf{0.0117} \\
			& 8-Obj & 0.4106$\pm$0.0391 & $+$ & 0.1772$\pm$0.0359 & $+$ & 0.1282$\pm$0.0361 & $+$ & \textbf{0.0673}$\pm$\textbf{0.0152} \\
			\hline
			\multirow{2}{*}{mDTLZ4-($1,0$)} & 5-Obj  & 0.4405$\pm$0.0317 & $+$ & 0.1492$\pm$0.0498 & $+$ & 0.0882$\pm$0.0176 & $+$ & \textbf{0.0436}$\pm$\textbf{0.0054} \\
			& 8-Obj & 0.4239$\pm$0.0407 & $+$ & 0.1232$\pm$0.0248 & $+$ & 0.0805$\pm$0.0213 & $+$ & \textbf{0.0533}$\pm$\textbf{0.0041} \\
			\hline
			\multicolumn{2}{c|}{$+/-/=$} & \multicolumn{2}{c|}{$8/0/0$} & \multicolumn{2}{c|}{$8/0/0$} & \multicolumn{2}{c|}{$8/0/0$} &
			\\
			\hline
	\end{tabular}}
\end{table*}

\subsection{Scaling to High-Dimensional Objective Spaces}
In previous experiments, the invTrEMO's performance was evaluated in problems with 3 objective functions. In this subsection, we extend our investigation to assess the effectiveness of the invTrEMO in higher-dimensional objective spaces (i.e., $m > 3$). To this end, we employ mDTLZ1-($1,0$) to mDTLZ4-($1,0$) as the target tasks, but with $d_T = 12$ and $m = 5$ or $8$ objective functions. Additionally, we utilize data from HS source tasks with $d_S = 11$ as transferrable information in the optimization process. We present the averaged IGD results corresponding to 5- and 8-objective target tasks provided by ParEGO-UCB, MOEA/D-EGO, K-RVEA, CSEA, PSL-MOBO, and invTrEMO-HS after 25, 50, 75, and 100 evaluations in Table~\ref{Tab:igd_5obj} and Table~\ref{Tab:igd_10obj}. We do not provide the result of qNEHVI since it becomes extremely time-consuming to calculate the expected hypervolume improvement in high-dimensional objective space. Moreover, to underscore the performance of the invTrEMO-HS in improving the accuracy of the inverse model in high-dimensional objective spaces, we also show the RMSE results achieved by PSL-MOBO, Post-hoc, Post-hoc-Transfer (with HS source dataset), and the invTrEMO-HS in Table~\ref{Tab:RMSE_high_dim}.

It is evident that, even in the case of 5- and 8-objective optimization problems, the invTrEMO-HS consistently exhibits significantly better convergence trends and inverse modeling accuracy compared to the baselines. Statistical tests, including the Kruskal-Wallis test and Wilcoxon's rank-sum test, confirm that invTrEMO-HS outperforms all competitors in terms of both IGD and RMSE results. These findings collectively emphasize the effectiveness of invTrEMO-HS in addressing problems of higher-dimensionality.

\subsection{Application to Vehicle Crashworthiness Design}\label{sec:rwp}
In the field of vehicle engineering, crashworthiness design is receiving increasing attention from researchers~\cite{sun2022lightweight,liu2013lightweight}. The goal is to simultaneously reduce fuel consumption and enhance safety by developing vehicle structures that are lightweight yet exhibit high crashworthiness. The problem is a prime example of an expensive engineering MOP, as crash simulations, while possibly being more cost effective than actual crash tests of a car, are still extremely computationally demanding. For example, the Ford
Motor Company reports that one crash simulation on a full
passenger car takes about 36-160 hours \cite{simpson2004approximation}. In practice, different vehicles, especially those of similar type, often share similar structural components. This presents an opportunity to transfer and reuse past design experiences to expedite the design process for a new target vehicle structure. 

Here, we use the test crashworthiness design problem of a National Highway Transportation and Safety Association Vehicle, as introduced in~\cite{liao2008multiobjective}, as an illustrative example to investigate the performance of the invTrEMO. We assess the vehicle structure using the following three indicators.
\begin{itemize}
	\item \textit{Mass of the Vehicle Structure} (${f}_{mass}$): This indicator assesses the level of lightweight design of the vehicle structure.
	
	\item \textit{Jerk Intrusion in the Full Frontal Crash} (${f}_{jerk}$): This indicator evaluates the crashworthiness of the vehicle structure in the event of a full frontal crash.
	
	\item \textit{Toeboard Intrusion in the Offset-Frontal Crash} (${f}_{toeboard}$): This indicator assesses the crashworthiness of the vehicle structure in the case of an offset-frontal crash.
\end{itemize}
Based on these indicators, the following 3-objective MOP is defined:
\begin{equation}\label{Eq:CrashworthinessDesign}
\begin{aligned}
\min: \ & \textbf{f}(\textbf{x}) = \{f_{mass}(\textbf{x}),f_{jerk}(\textbf{x}),f_{toeboard}(\textbf{x})\} \\
\text{s.t.}\ & \textbf{x} = (x_1,\ldots,x_d), \\
\ & 0.5 \leq x_j \leq 2, \\ 
\ & j \in \{1,\ldots, 5\}, \\
\end{aligned}
\end{equation}
where $x_j, j \in \{1,\ldots,5 \}$ are the thicknesses of the five thin-walled parts shown in~\cite{liao2008multiobjective}.

\begin{figure*}[!t]
	\begin{center}
		\subfigure[]{\label{fig:case_1}\includegraphics[width=0.3\columnwidth]{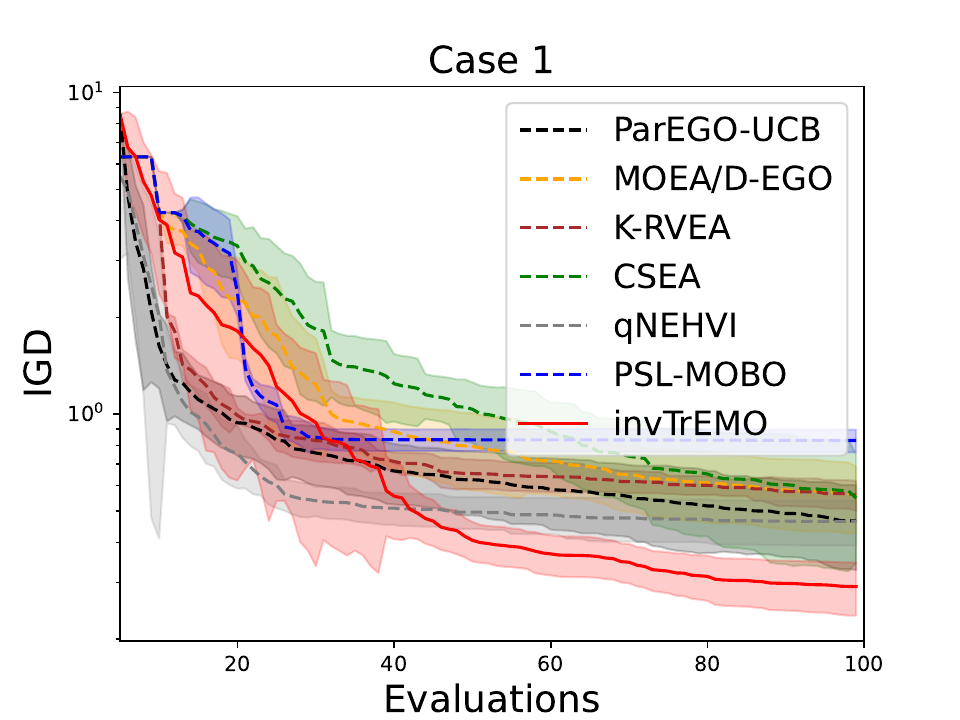}}
		\subfigure[]{\label{fig:case_2}\includegraphics[width=0.3\columnwidth]{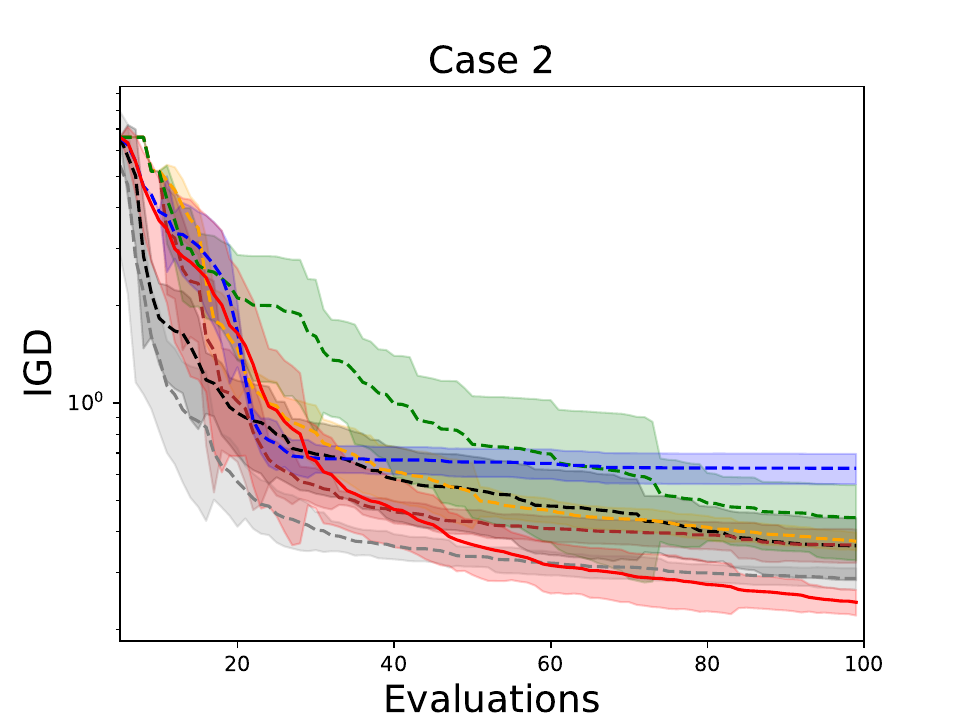}}
		\subfigure[]{\label{fig:case_3}\includegraphics[width=0.3\columnwidth]{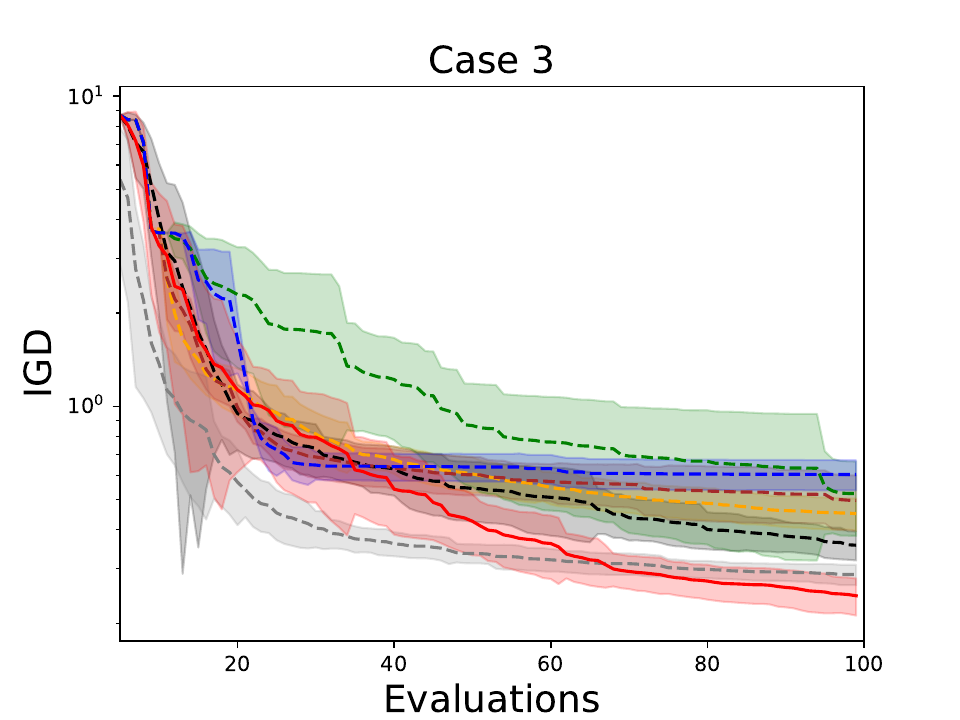}}
		\caption{IGD convergence trends of ParEGO-UCB, MOEA/D-EGO, K-RVEA, CSEA, qNEHVI, PSL-MOBO, and the proposed invTrEMO on the vehicle crashworthiness design cases 1, 2, and 3. The shaded areas represent one standard deviation in performance on either side of the mean. (a) Case 1. (b) Case 2. (c) Case 3.}\label{Fig:Crashoworthiness}
	\end{center}
\end{figure*}

\begin{table*}[]
	\centering
	\caption{RMSE results of PSL-MOBO, Post-hoc, Post-hoc-Transfer, and the invTrEMO on vehicle crashworthiness design cases 1, 2, and 3 over 20 independent runs. The Wilcoxon's
		rank-sum test at a 0.05 significance level was carried out between invTrEMO and PSL-MOBO, Post-hoc, and Post-hoc-Transfer.}\label{Tab:RMSE_Crash}
	\resizebox{12cm}{!}{\begin{tabular}{c|c|c|c|c|c|c|ccc}
			\hline
			\multirow{2}{*}{Target Tasks}   & \multicolumn{2}{c|}{PSL-MOBO} & \multicolumn{2}{c|}{Post-hoc} & \multicolumn{2}{c|}{Post-hoc-Transfer} & invTrEMO \\
			\cline{2-8}
			& \multicolumn{2}{c|}{Average RMSE $\pm$ Std} & \multicolumn{2}{c|}{Average RMSE $\pm$ Std} & \multicolumn{2}{c|}{Average RMSE $\pm$ Std} & \multicolumn{1}{c}{Average RMSE $\pm$ Std} \\
			\hline
			Case1 & 3.0947$\pm$0.0464 & $+$ & 0.7544$\pm$0.5442 & $+$ & 0.7221$\pm$0.5716 & $+$ & \textbf{0.3085$\pm$0.1435} \\
			Case2 & 2.9804$\pm$0.0139 & $+$ & 0.6538$\pm$0.6214 & $+$ & 0.4845$\pm$0.2010 & $+$ & \textbf{0.1934$\pm$0.0317} \\
			Case3 & 3.0868$\pm$0.0124 & $+$ & 0.5835$\pm$0.4022 & $+$ & 0.5255$\pm$0.3918 & $+$ & \textbf{0.1869$\pm$0.0358} \\
			\hline
			\multicolumn{1}{c|}{$+/-/=$} & \multicolumn{2}{c|}{$3/0/0$} & \multicolumn{2}{c|}{$3/0/0$} & \multicolumn{2}{c|}{$3/0/0$} &
			\\
			\hline
	\end{tabular}}
\end{table*}

The aforementioned indicators serve as general objective functions used for evaluating crashworthiness of vehicle structures of different types, thereby serving as a concrete example of the prevalence of common objective spaces within a domain. As a result, even if there are non-identical search spaces between source and target tasks, the common objectives coupled with a subset of overlapping decision variables, such as the thicknesses of analogous thin-walled components, motivates the transfer of data across them. Taking various scenarios into consideration, this subsection presents three transfer optimization use-cases.
\begin{itemize}
	\item \textit{Case 1}: The source task consists of four decision variables, while the target task includes five decision variables. Specifically, the target task is identical to \eqref{Eq:CrashworthinessDesign}, and the source MOP is formed by fixing the first variable of \eqref{Eq:CrashworthinessDesign} to 0.5 and optimizing only the last four variables. The source can therefore be thought of as a more constrained version of the target. We have $d_S < d_T$, with an overlap of four decision variables between the source and target tasks.
	\item \textit{Case 2}: Both the source and target tasks consist of four decision variables. Specifically, the target task is formed by setting the first variable of \eqref{Eq:CrashworthinessDesign} to 0.5 and optimizing only the last four variables. Conversely, the source task is formed by setting the last variable of \eqref{Eq:CrashworthinessDesign} to 0.5 and optimizing only the first four variables. In this case, the source and the target are differently constrained. Nevertheless, we have $d_S = d_T$, and there is an overlap of three decision variables between the source and the target tasks.
	\item \textit{Case 3}: The source task comprises five decision variables, while the target task consists of four decision variables. Here, the target task is formed by setting the first variable of \eqref{Eq:CrashworthinessDesign} to 0.5 and optimizing only the last four variables. On the other hand, the source task is identical to \eqref{Eq:CrashworthinessDesign}. This case is therefore the converse of \emph{Case 1}. We have $d_S > d_T$, and there is an overlap of four decision variables between the source and target tasks.
\end{itemize}

We employ the invTrEMO with $N^{init}$ set to 6 and an evolution budget of 100 to solve the three MOPs. For comparison, we utilize the ParEGO-UCB, MOEA/D-EGO, K-RVEA, CSEA, qNEHVI, and the PSL-MOBO algorithm. To calculate IGD and RMSE values, we utilize NSGA-III to solve all three test cases, taking the obtained nondominated solutions as a reasonable approximation of the true PFs. The population size of NSGA-III is set to 200, and the number of iterations is set to 500. The averaged IGD results are presented in Fig.\ref{Fig:Crashoworthiness}. The shaded regions in the figure represent one standard deviation in performance, illustrating the variation around the mean across 20 independent runs. Fig.\ref{Fig:Crashoworthiness} clearly demonstrates that in all three cases, the invTrEMO consistently outperforms the other five competitors in terms of IGD. This highlights invTrEMO's ability to enhance optimization convergence rates by leveraging prior experiences from heterogeneous sources. Additionally, we provide the RMSE results obtained by PSL-MOBO, Post-hoc, Post-hoc-Transfer, and the invTrEMO in Table~\ref{Tab:RMSE_Crash} to assess the accuracy of inverse modeling. The results indicate that the invTrEMO achieves the best performance, affirming the overall effectiveness of the inverse transfer optimization paradigm for expensive MOPs.

\section{Conclusion}
Diverging from previous transfer optimization algorithms that aggregate learnt models in decision space, this paper presents a new concept of \emph{inverse transfer multiobjective optimization} by objective space unification. The corresponding invTrEMO facilitates knowledge transfers by leveraging the prevalence of common objective spaces in problems within any given area of application. Centring on inverse models lends our method the following distinctive features.
\begin{itemize}
	\item By treating the common objective space as a bridge between source and target tasks, the invTrEMO is the first algorithm for MOPs to make possible the transfer of data even across heterogeneous tasks with non-identical decision spaces. This greatly expands the spectrum of problems that transfer optimization can be applied to effectively.
	\item The inverse transfer models created as a byproduct of the invTrEMO empower a DM to obtain nondominated solutions that better align with their desired trade-off preference. This hints at a future of seamless human-machine interaction in multiobjective decision-making.
\end{itemize}
The performance of the invTrEMO is evaluated through extensive experiments on a varied set of multi- and many-objective benchmark problems. The experimental results demonstrate that the invTrEMO, leveraging inverse transfer, not only achieves competitive or better convergence behavior compared to state-of-the-art evolutionary and Bayesian optimization algorithms, but also produces accurate inverse models for PF approximation. The results are in agreement with the intuition that stronger correlations and a greater number of overlapping decision variables between source and target tasks lead to improved optimization outcomes. We also show that the proposed invTrEMO scales well to 5- and 8-dimensional objective spaces. Finally, the real-world applicability of the method is presented through different use-cases in the crashworthiness design domain.

While the results of the invTrEMO are encouraging, they also shed light on important avenues for future exploration. First, in our implementation we build the invTGP models for each decision variable independently, without accounting for the potential relationships between variables in the PS of the target task. This is done due to computational considerations, but could, however, hinder inverse modeling in cases where nondominated solution data is scarce. Notably, the invTrEMO currently utilizes only nondominated solutions from the source data to guide the optimization of the target task, which also means that information contained in the dominated solutions is not fully utilized. Future research shall therefore focus on strategies to incorporate and make better use of such added information streams. Second, the predefined set of preference vectors used to decompose the MOP may not be the best suited to handle problems with complicated or non-standard PF shapes, pointing to yet another path for work on adaptive preference vectors. Addressing these limitations, and others, would further enhance the capabilities of the invTrEMO, providing us with a powerful and general-purpose tool for solving MOPs efficiently.


\appendix
\section{Appendix: Proof of Proposition 1}
\begin{proof}
	Let $\textbf{x}^{arb} \neq \textbf{x}^{ps}$ be an arbitrary solution in $\mathcal{X}$. Since $\textbf{x}^{ps}$ is a Pareto optimal solution to a multiobjective minimization problem, let there exist $s \in \{1,\ldots,m\}$ such that 
	\begin{equation}\label{Eq:1}
	\begin{aligned}
	f_{s}(\textbf{x}^{arb}) > f_{s}(\textbf{x}^{ps}).
	\end{aligned}
	\end{equation}
	
	Let $Z = \sum_{i=1}^{m} c_i^{(ps)} = \sum_{i=1}^{m}\frac{\sum_{v=1}^{m} f_{v}(\textbf{x}^{ps})}{f_{i}(\textbf{x}^{ps})} $, we have:
	
	\begin{equation}
	\begin{aligned}
	f^{tch}(\textbf{x}^{ps}) & = \max_{i \in \{1,\ldots,m\}} \{ w_{i} f_{i}(\textbf{x}^{ps}) \} + \eta \sum_{i=1}^{m} w_{i} f_{i}(\textbf{x}^{ps}) \\
	& {=  \max_{i \in \{1,\ldots,m\}} \{ \frac{c_i^{(ps)}}{\sum_{v=1}^{m} c_{v}^{(ps)}} \cdot f_{i}(\textbf{x}^{ps}) \}  } \\
	& {=  \max_{i \in \{1,\ldots,m\}} \{ \frac{1}{Z} \cdot \frac{\sum_{v=1}^{m} f_{v}(\textbf{x}^{ps})}{f_{i}(\textbf{x}^{ps})} \cdot f_{i}(\textbf{x}^{ps}) \}  } \\
	& {=  \begin{matrix} \max \{ \underbrace{ \frac{1}{Z} \cdot \sum_{v=1}^{m} f_{v}(\textbf{x}^{ps}), \ldots, \frac{1}{Z} \cdot \sum_{v=1}^{m} f_{v}(\textbf{x}^{ps}) \}  } \\    \text{totally } m \text{ equal terms} \end{matrix} } \\
	& = \frac{1}{Z} \cdot \sum_{i=1}^{m} f_{i}(\textbf{x}^{ps}) \\
	& < \frac{1}{Z} \cdot \frac{\sum_{i=1}^{m} f_{i}(\textbf{x}^{ps})}{f_{s}(\textbf{x}^{ps}) } \cdot f_{s}(\textbf{x}^{arb}) \\
	& = w_{s} f_{s}(\textbf{x}^{arb})\\
	& \leq \max_{i \in \{1,\ldots,m\}} \{ w_{i} f_{i}(\textbf{x}^{arb}) \} \\
	& = \max_{i \in \{1,\ldots,m\}} \{ w_{i} f_{i}(\textbf{x}^{arb}) \} + \eta \sum_{i=1}^{m} w_{i} f_{i}(\textbf{x}^{arb})\\
	& = f^{tch}(\textbf{x}^{arb}),
		\end{aligned}
	\end{equation}
	where the second equality holds by substituting (6) and $\eta=0$ into (2), and the first inequality holds according to \eqref{Eq:1}. Since $\textbf{x}^{arb}$ can be any arbitrary solution in $\mathcal{X}$, and we have shown that $\textbf{x}^{arb}$ cannot achieve a lower $f^{tch}$ value than $\textbf{x}^{ps}$, it follows that $f^{tch}(\textbf{x}^{ps}) = \min_{\textbf{x}} f^{tch}(\textbf{x})$.
\end{proof}

\bibliographystyle{ACM-Reference-Format}
\bibliography{myref}

\newpage
	\begin{center}
		\textbf{\LARGE Supplementary File of ``Bayesian Inverse Transfer in Evolutionary Multiobjective Optimization''}
	\end{center}

	
	\setcounter{section}{0}
	\renewcommand\thesection{S-\Roman{section}}
	
	\setcounter{table}{0}
	\renewcommand\thetable{S-\Roman{table}}
	
	\setcounter{figure}{0}
	\renewcommand\thefigure{S-\arabic{figure}}
	
	\setcounter{equation}{0}
	\renewcommand\theequation{S-\arabic{equation}}
	
	{
		\section{Benchmark Problems}
		
		\subsection{Fomulation of the Benchmark Problems}
		Based on the widely known DTLZ and DTLZ$^{-1}$ benchmark suite, eight base problems, i.e., mDTLZ1-($\delta_1,\delta_2$), mDTLZ2-($\delta_1,\delta_2$), mDTLZ3-($\delta_1,\delta_2$), mDTLZ4-($\delta_1,\delta_2$), mDTLZ1$^{-1}$-($\delta_1,\delta_2$), mDTLZ2$^{-1}$-($\delta_1,\delta_2$), mDTLZ3$^{-1}$-($\delta_1,\delta_2$), and mDTLZ4$^{-1}$-($\delta_1,\delta_2$) are constructed. These eight base problems are shown as follows:
		\begin{itemize}
			\item \textit{mDTLZ1-($\delta_1,\delta_2$)}: 
			\begin{equation}\label{eqn:DTLZ1}
			\begin{aligned}
			\min: \ & f_1(\textbf{x}) = \frac{1}{2} x_1^{\delta_1} x_2^{\delta_1} \cdots x_{m-1}^{\delta_1} (1 + g(\textbf{x})) \\
			& f_2(\textbf{x}) = \frac{1}{2} x_1^{\delta_1} x_2^{\delta_1} \cdots (1 - x_{m-1}^{\delta_1}) (1 + g(\textbf{x})) \\
			& \ \ \ \ \ \ \cdots \\
			& f_{m-1}(\textbf{x}) = \frac{1}{2} x_1^{\delta_1} (1 - x_2^{\delta_1}) (1 + g(\textbf{x})) \\ 
			& f_m(\textbf{x}) = \frac{1}{2} (1 - x_1^{\delta_1}) (1 + g(\textbf{x})) \\ 
			\text{s.t.} \ & 0 \leq x_j \leq 1, j = 1,\ldots,d
			\end{aligned}
			\end{equation}
			where
			\begin{equation}\label{eqn:DTLZ1_g}
			\begin{aligned}
			g(\textbf{x}) = (d - m + 1) + \sum_{j=m}^{d} ((x_j - 0.5 - \delta_2)^2 - \cos (2 \pi (x_j - 0.5 - \delta_2)))
			\end{aligned}
			\end{equation}
			
			\item \textit{mDTLZ2-($\delta_1,\delta_2$)}: 
			\begin{equation}\label{eqn:DTLZ2}
			\begin{aligned}
			\min: \ & f_1(\textbf{x}) = (1 + g(\textbf{x})) \cos (\frac{\pi}{2} {x}_1^{\delta_1}) \cos (\frac{\pi}{2} {x}_2^{\delta_1}) \cdots \cos (\frac{\pi}{2} {x}_{m-1}^{\delta_1}) \\
			& f_2(\textbf{x}) = (1 + g(\textbf{x})) \cos (\frac{\pi}{2} {x}_1^{\delta_1}) \cos (\frac{\pi}{2} {x}_2^{\delta_1}) \cdots \sin (\frac{\pi}{2} {x}_{m-1}^{\delta_1}) \\
			& \ \ \ \ \ \ \cdots \\
			& f_{m-1}(\textbf{x}) = (1 + g(\textbf{x})) \cos (\frac{\pi}{2} {x}_1^{\delta_1}) \sin (\frac{\pi}{2} {x}_2^{\delta_1}) \\
			& f_m(\textbf{x}) = (1 + g(\textbf{x})) \sin (\frac{\pi}{2} {x}_1^{\delta_1}) \\
			\text{s.t.} \ & 0 \leq x_j \leq 1, j = 1,\ldots,d
			\end{aligned}
			\end{equation}
			where
			\begin{equation}\label{eqn:DTLZ2_g}
			\begin{aligned}
			g(\textbf{x}) = \sum_{j=m}^{d} (x_j - 0.5 - \delta_2)^2
			\end{aligned}
			\end{equation}
			
			\item \textit{mDTLZ3-($\delta_1,\delta_2$)}: 
			\begin{equation}\label{eqn:DTLZ3}
			\begin{aligned}
			\min: \ & f_1(\textbf{x}) = (1 + g(\textbf{x})) \cos (\frac{\pi}{2} {x}_1^{\delta_1}) \cos (\frac{\pi}{2} {x}_2^{\delta_1}) \cdots \cos (\frac{\pi}{2} {x}_{m-1}^{\delta_1}) \\
			& f_2(\textbf{x}) = (1 + g(\textbf{x})) \cos (\frac{\pi}{2} {x}_1^{\delta_1}) \cos (\frac{\pi}{2} {x}_2^{\delta_1}) \cdots \sin (\frac{\pi}{2} {x}_{m-1}^{\delta_1}) \\
			& \ \ \ \ \ \ \cdots \\
			& f_{m-1}(\textbf{x}) = (1 + g(\textbf{x})) \cos (\frac{\pi}{2} {x}_1^{\delta_1}) \sin (\frac{\pi}{2} {x}_2^{\delta_1}) \\
			& f_m(\textbf{x}) = (1 + g(\textbf{x})) \sin (\frac{\pi}{2} {x}_1^{\delta_1}) \\
			\text{s.t.} \ & 0 \leq x_j \leq 1, j = 1,\ldots,d
			\end{aligned}
			\end{equation}
			where
			\begin{equation}\label{eqn:DTLZ3_g}
			\begin{aligned}
			g(\textbf{x}) = \frac{1}{10}(d-m+1) + \sum_{j=m}^{d} ((x_j - 0.5 - \delta_2)^2 - \frac{1}{10}\cos (2 \pi (x_j - 0.5 - \delta_2)))
			\end{aligned}
			\end{equation}
			
			\item \textit{mDTLZ4-($\delta_1,\delta_2$)}: 
			\begin{equation}\label{eqn:DTLZ4}
			\begin{aligned}
			\min: \ & f_1(\textbf{x}) = (1 + g(\textbf{x})) \cos (\frac{\pi}{2} {x}_1^{2 \cdot \delta_1}) \cos (\frac{\pi}{2} {x}_2^{2 \cdot \delta_1}) \cdots \cos (\frac{\pi}{2} {x}_{m-1}^{2 \cdot \delta_1}) \\
			& f_2(\textbf{x}) = (1 + g(\textbf{x})) \cos (\frac{\pi}{2} {x}_1^{2 \cdot \delta_1}) \cos (\frac{\pi}{2} {x}_2^{2 \cdot \delta_1}) \cdots \sin (\frac{\pi}{2} {x}_{m-1}^{2 \cdot \delta_1}) \\
			& \ \ \ \ \ \ \cdots \\
			& f_{m-1}(\textbf{x}) = (1 + g(\textbf{x})) \cos (\frac{\pi}{2} {x}_1^{2 \cdot \delta_1}) \sin (\frac{\pi}{2} {x}_2^{2 \cdot \delta_1}) \\
			& f_m(\textbf{x}) = (1 + g(\textbf{x})) \sin (\frac{\pi}{2} {x}_1^{2 \cdot \delta_1}) \\
			\text{s.t.} \ & 0 \leq x_j \leq 1, j = 1,\ldots,d
			\end{aligned}
			\end{equation}
			where
			\begin{equation}\label{eqn:DTLZ4_g}
			\begin{aligned}
			g(\textbf{x}) = \sum_{j=m}^{d} (x_j - 0.5 - \delta_2)^2
			\end{aligned}
			\end{equation}
			
			\item \textit{mDTLZ1$^{-1}$-($\delta_1,\delta_2$)}: 
			\begin{equation}\label{eqn:DTLZ1}
			\begin{aligned}
			\min: \ & f_1(\textbf{x}) = -\frac{1}{2} x_1^{\delta_1} x_2^{\delta_1} \cdots x_{m-1}^{\delta_1} (1 - g(\textbf{x})) \\
			& f_2(\textbf{x}) = -\frac{1}{2} x_1^{\delta_1} x_2^{\delta_1} \cdots (1 - x_{m-1}^{\delta_1}) (1 - g(\textbf{x})) \\
			& \ \ \ \ \ \ \cdots \\
			& f_{m-1}(\textbf{x}) = -\frac{1}{2} x_1^{\delta_1} (1 - x_2^{\delta_1}) (1 - g(\textbf{x})) \\ 
			& f_m(\textbf{x}) = -\frac{1}{2} (1 - x_1^{\delta_1}) (1 - g(\textbf{x})) \\ 
			\text{s.t.} \ & 0 \leq x_j \leq 1, j = 1,\ldots,d
			\end{aligned}
			\end{equation}
			where
			\begin{equation}\label{eqn:DTLZ1_g}
			\begin{aligned}
			g(\textbf{x}) = (d - m + 1) + \sum_{j=m}^{d} ((x_j - 0.5 - \delta_2)^2 - \cos (2 \pi (x_j - 0.5 - \delta_2)))
			\end{aligned}
			\end{equation}
			
			\item \textit{mDTLZ2$^{-1}$-($\delta_1,\delta_2$)}: 
			\begin{equation}\label{eqn:DTLZ2}
			\begin{aligned}
			\min: \ & f_1(\textbf{x}) = -(1 - g(\textbf{x})) \cos (\frac{\pi}{2} {x}_1^{\delta_1}) \cos (\frac{\pi}{2} {x}_2^{\delta_1}) \cdots \cos (\frac{\pi}{2} {x}_{m-1}^{\delta_1}) \\
			& f_2(\textbf{x}) = -(1 - g(\textbf{x})) \cos (\frac{\pi}{2} {x}_1^{\delta_1}) \cos (\frac{\pi}{2} {x}_2^{\delta_1}) \cdots \sin (\frac{\pi}{2} {x}_{m-1}^{\delta_1}) \\
			& \ \ \ \ \ \ \cdots \\
			& f_{m-1}(\textbf{x}) = -(1 - g(\textbf{x})) \cos (\frac{\pi}{2} {x}_1^{\delta_1}) \sin (\frac{\pi}{2} {x}_2^{\delta_1}) \\
			& f_m(\textbf{x}) = -(1 - g(\textbf{x})) \sin (\frac{\pi}{2} {x}_1^{\delta_1}) \\
			\text{s.t.} \ & 0 \leq x_j \leq 1, j = 1,\ldots,d
			\end{aligned}
			\end{equation}
			where
			\begin{equation}\label{eqn:DTLZ2_g}
			\begin{aligned}
			g(\textbf{x}) = \sum_{j=m}^{d} (x_j - 0.5 - \delta_2)^2
			\end{aligned}
			\end{equation}
			
			\item \textit{mDTLZ3$^{-1}$-($\delta_1,\delta_2$)}: 
			\begin{equation}\label{eqn:DTLZ3}
			\begin{aligned}
			\min: \ & f_1(\textbf{x}) = -(1 - g(\textbf{x})) \cos (\frac{\pi}{2} {x}_1^{\delta_1}) \cos (\frac{\pi}{2} {x}_2^{\delta_1}) \cdots \cos (\frac{\pi}{2} {x}_{m-1}^{\delta_1}) \\
			& f_2(\textbf{x}) = -(1 - g(\textbf{x})) \cos (\frac{\pi}{2} {x}_1^{\delta_1}) \cos (\frac{\pi}{2} {x}_2^{\delta_1}) \cdots \sin (\frac{\pi}{2} {x}_{m-1}^{\delta_1}) \\
			& \ \ \ \ \ \ \cdots \\
			& f_{m-1}(\textbf{x}) = -(1 - g(\textbf{x})) \cos (\frac{\pi}{2} {x}_1^{\delta_1}) \sin (\frac{\pi}{2} {x}_2^{\delta_1}) \\
			& f_m(\textbf{x}) = -(1 - g(\textbf{x})) \sin (\frac{\pi}{2} {x}_1^{\delta_1}) \\
			\text{s.t.} \ & 0 \leq x_j \leq 1, j = 1,\ldots,d
			\end{aligned}
			\end{equation}
			where
			\begin{equation}\label{eqn:DTLZ3_g}
			\begin{aligned}
			g(\textbf{x}) = \frac{1}{10}(d-m+1) + \sum_{j=m}^{d} ((x_j - 0.5 - \delta_2)^2 - \frac{1}{10}\cos (2 \pi (x_j - 0.5 - \delta_2)))
			\end{aligned}
			\end{equation}
			
			\item \textit{mDTLZ4$^{-1}$-($\delta_1,\delta_2$)}: 
			\begin{equation}\label{eqn:DTLZ4}
			\begin{aligned}
			\min: \ & f_1(\textbf{x}) = -(1 - g(\textbf{x})) \cos (\frac{\pi}{2} {x}_1^{2 \cdot \delta_1}) \cos (\frac{\pi}{2} {x}_2^{2 \cdot \delta_1}) \cdots \cos (\frac{\pi}{2} {x}_{m-1}^{2 \cdot \delta_1}) \\
			& f_2(\textbf{x}) = -(1 - g(\textbf{x})) \cos (\frac{\pi}{2} {x}_1^{2 \cdot \delta_1}) \cos (\frac{\pi}{2} {x}_2^{2 \cdot \delta_1}) \cdots \sin (\frac{\pi}{2} {x}_{m-1}^{2 \cdot \delta_1}) \\
			& \ \ \ \ \ \ \cdots \\
			& f_{m-1}(\textbf{x}) = -(1 - g(\textbf{x})) \cos (\frac{\pi}{2} {x}_1^{2 \cdot \delta_1}) \sin (\frac{\pi}{2} {x}_2^{2 \cdot \delta_1}) \\
			& f_m(\textbf{x}) = -(1 - g(\textbf{x})) \sin (\frac{\pi}{2} {x}_1^{2 \cdot \delta_1}) \\
			\text{s.t.} \ & 0 \leq x_j \leq 1, j = 1,\ldots,d
			\end{aligned}
			\end{equation}
			where
			\begin{equation}\label{eqn:DTLZ4_g}
			\begin{aligned}
			g(\textbf{x}) = \sum_{j=m}^{d} (x_j - 0.5 - \delta_2)^2
			\end{aligned}
			\end{equation}
		\end{itemize}
		
		\subsection{Further Discussions on the Benchmark Problems}
		{
			\subsubsection{\textbf{Influence of $\delta_1$ and $\delta_2$}}
			Let's begin by discussing the characteristics of the mDTLZ and mDTLZ$^{-1}$ benchmark problems. Across all these problems, a notable pattern emerges: the first $(m-1)$ decision variables (referred to as $\textbf{x}_I$) define the PF, while the remaining decision variables (denoted as $\textbf{x}_{II}$) determine the position of the PS within the decision space. All of the nondominated solutions in the PS share the same $\textbf{x}_{II}$ values. Then, according to the formulations, the first parameter, $\delta_1$, primarily influences $\textbf{x}_I$, thereby affecting the mapping relationship from a preference vector to a Pareto optimal solution. Specifically, for two mDTLZ problems with different $\delta_1$ values and the same $\delta_2$ values, the same preference vector $\textbf{w}$ will correspond to different $\textbf{x}_I$ values. Notably, however, this parameter does not alter the shape of the PF. Conversely, the second parameter, $\delta_2$, predominantly influences $\textbf{x}_{II}$. It distinguishes the PS corresponding to mDTLZ or mDTLZ$^{-1}$ problems with varying $\delta_2$ values by shifting the PS within the decision space.
			
			An important aspect to consider is that, although $\delta_1$ doesn't alter the shape of the PF, it remains pertinent to examine its influence. This is due to a distinctive requirement in our context compared to traditional multiobjective optimization endeavors. Here, our concern extends beyond merely obtaining the PS of the problem; \textit{we also aim to ensure that, upon receiving a preference vector from the user, the inverse model accurately retrieves the corresponding nondominated solution}. Thus, in addition to the commonly discussed convergence and diversity metrics, it becomes imperative to delve deeper into the mapping relationship from preference vectors to nondominated solutions. Fortunately, parameter $\delta_1$ fulfills this need.
			
			\subsubsection{\textbf{Limitations of mDTLZ and mDTLZ$^{-1}$ Benchmark Problems}}
			As previously mentioned, since no parameters are specifically designed to adjust the shape of the PF in the benchmark problems, it remains unexplored whether the inverse transfer technique can be effectively employed when the source and target exhibit different PF shapes. Furthermore, the benchmark problems, particularly mDTLZ benchmarks, often feature regular-shaped PFs (e.g., hyper-plane or hyper-sphere), which might not readily generalize to more realistic scenarios.
			
			To further explore the effectiveness of invTrEMO when the source and target problems feature different PF shapes, we apply it to solve WFG2 within the WFG benchmark suite. We employ mDTLZ2-(1,0) as the source task to generate the source dataset. For comparison, we also utilize invTrEMO-ZeroT to solve WFG2, aiming to assess the efficacy of inverse transfer. The convergence trend in terms of IGD and RMSE results are depicted in Fig.~\ref{Fig:wfg}. The results indicate that invTrEMO, leveraging the source data, exhibits faster convergence and achieves superior RMSE performance compared to invTrEMO-ZeroT. This observation suggests that even in scenarios where the source and target problems possess different PF shapes, the inverse transfer technique can still bolster convergence rates and enhance the accuracy of inverse models.
			
			\begin{figure*}[h]
				\begin{center}
					\subfigure[]{\label{wfg_igd}\includegraphics[width=0.35\columnwidth]{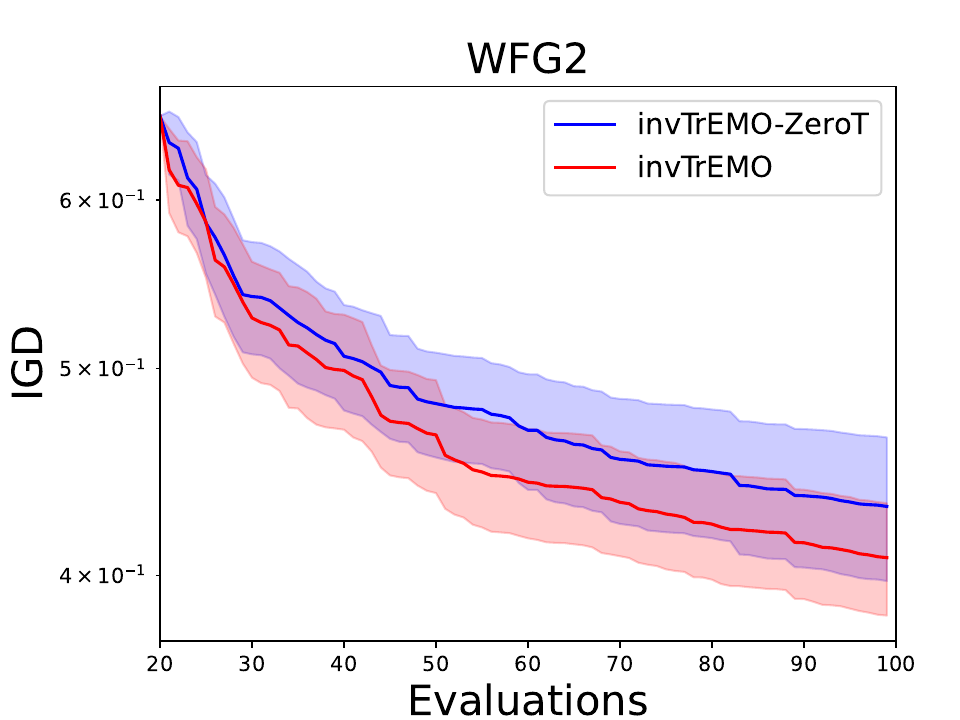}}
					\subfigure[]{\label{wfg_rmse}\includegraphics[width=0.35\columnwidth]{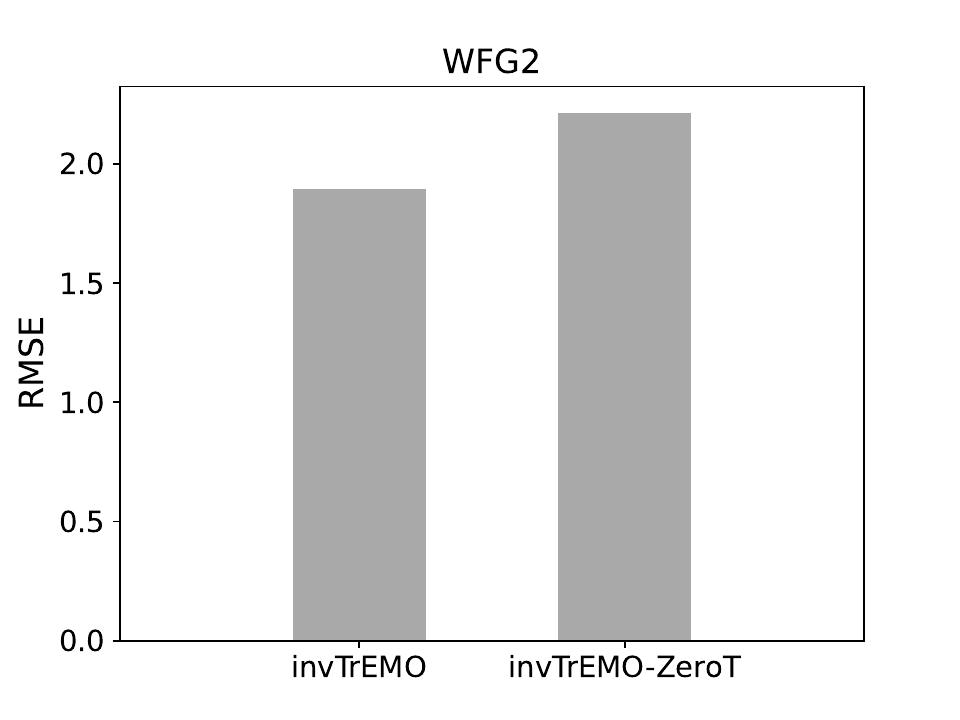}}
					\caption{The IGD convergence trends and the RMSE results provided by invTrEMO and invTrEMO-ZeroT on WFG2. The shaded areas represent half a standard deviation in performance on either side of the mean. (a) IGD convergence trends. (b) RMSE results.}\label{Fig:wfg}
				\end{center}
			\end{figure*}
			
			Indeed, from the perspective of the inverse modeling, there remains a necessity for further exploration of the characteristics and construction of benchmark problems. While current benchmarks predominantly emphasize the shapes of the PF, they often overlook the mapping from preference vectors to nondominated solutions—an essential aspect for inverse modeling. This area offers ample research opportunities, but the focus of this study does not lie here, so we will not delve into it in depth.
		}
		
		\newpage
		
		\section{Additional Results}
		\subsection{Further Investigations on the Transfer Mechanism}
		{
			In Section 3.4, we outlined how invTGPs harness information from source data to enhance optimization. This section delves deeper into the experimental investigation of this transfer mechanism. Specifically, we compare the UCB value of the selected candidate solution $\textbf{x}_T$ with the maximum UCB value in each iteration. Figure~\ref{Fig:ucb} displays the corresponding data from a single run utilizing invTrEMO to tackle mDTLZ-$(1,0)$. Notably, throughout each iteration, the UCB value of $\textbf{x}_T$ consistently remains lower than the maximum UCB value. This observation suggests that invTGPs confine the search within the region covered by their generated offspring solutions rather than pursuing the solution with the highest UCB across the entire search space. Such an approach yields optimization benefits, as discussed in Section 3.4. By concentrating the search primarily on regions learned by the invTGPs, the algorithm effectively avoids over-exploration and redundant evaluations in unpromising regions, thereby reducing the algorithm's evaluation budget.
			\begin{figure*}[h]
				\begin{center}
					{\label{wfg_igd}\includegraphics[width=0.7\columnwidth]{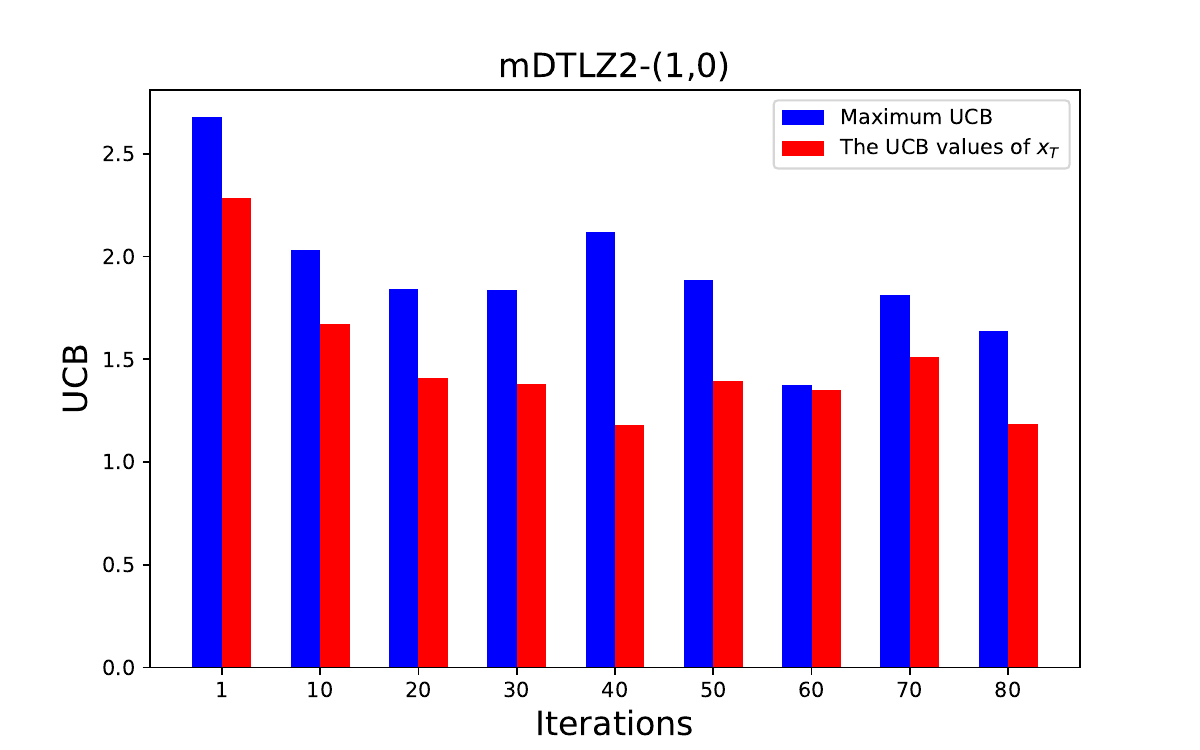}}
					\caption{The UCB values of $\textbf{x}_T$ and the maximum UCB values provided by the invTrEMO on solving mDTLZ-$(1,0)$ in the 1\textit{st}-80\textit{th} iterations. The results are shown every 10 iterations.}\label{Fig:ucb}
				\end{center}
			\end{figure*}

		}
		
		\subsection{Effectiveness of the Two Step Training Process}
		{
			\begin{figure*}[h]
				\begin{center}
					\subfigure[]{\label{two_stage_igd}\includegraphics[width=0.35\columnwidth]{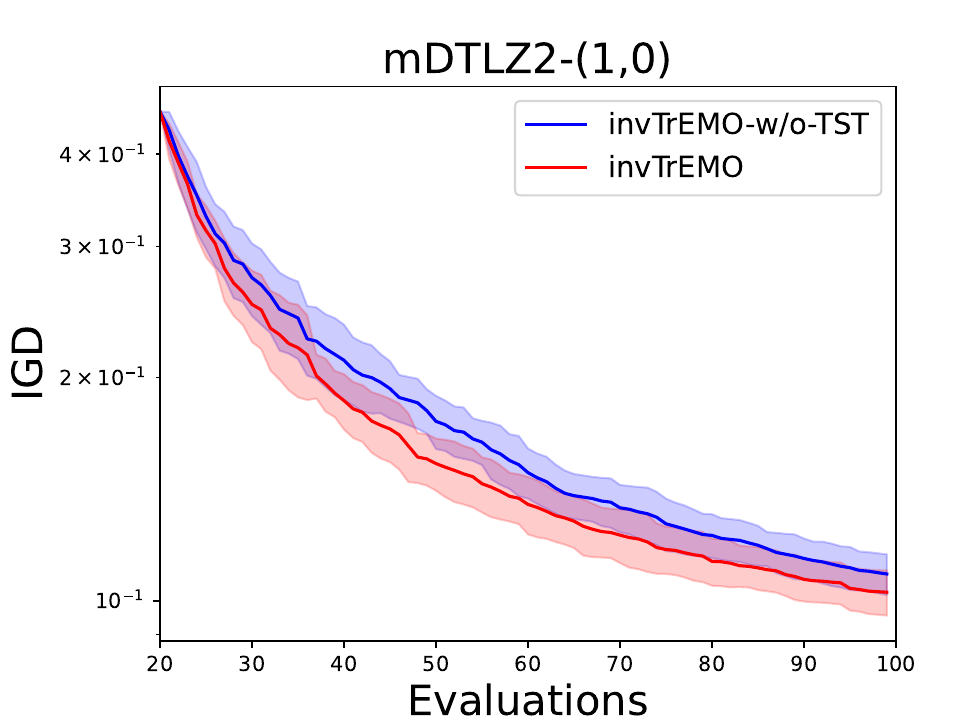}}
					\subfigure[]{\label{two_stage_rmse}\includegraphics[width=0.35\columnwidth]{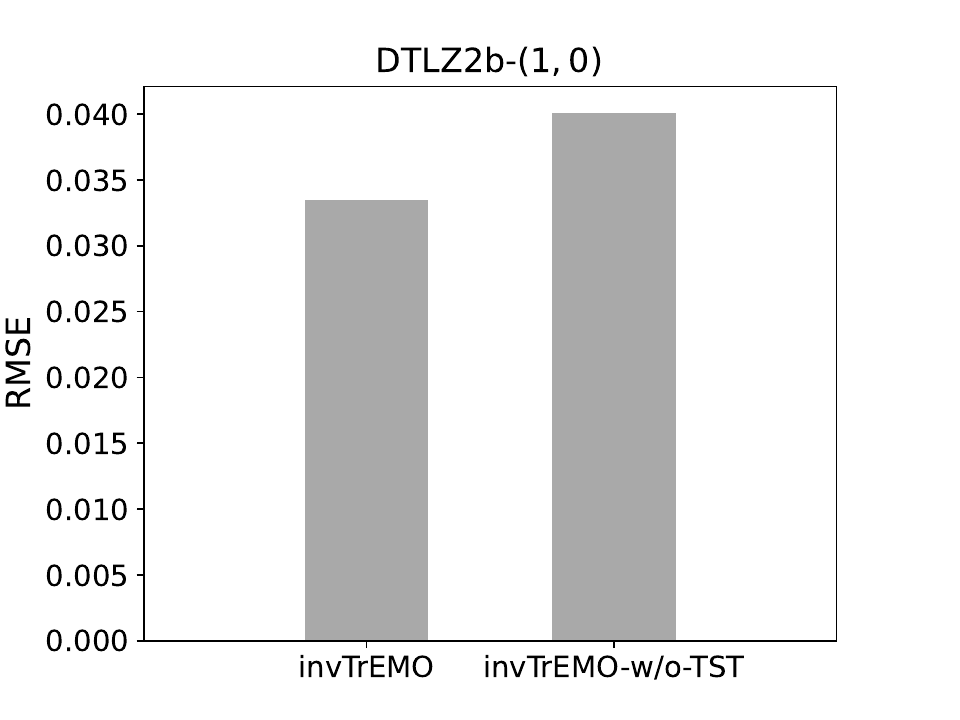}}
					\caption{The IGD convergence trends and the RMSE results provided by invTrEMO and invTrEMO-w/o-TST on mDTLZ-$(1,0)$. The shaded areas represent one standard deviation in performance on either side of the mean. (a) IGD convergence trends. (b) RMSE results.}\label{Fig:TwoStage}
				\end{center}
			\end{figure*}
			
			In Section 3.2, we employ a two step training process for optimizing the hyperparameters of the invTGPs. This strategy is designed to mitigate the risk of the inverse models exhibiting bias towards the source task rather than the target task, especially when the number of samples in the target dataset is significantly smaller than that in the source dataset. By doing so, the negative transfer can be alleviated to some extent. To assess the efficacy of this two step training process, we compare the performance of invTrEMO with a variant referred to as invTrEMO-w/o-TST on solving mDTLZ2-$(1,0)$, leveraging a MS source dataset for assistance. In invTrEMO-w/o-TST, we train the hyperparameters of the invTGPs jointly in one pass. The convergence trend of IGD results and the RMSE results are illustrated in Fig.~\ref{Fig:TwoStage}. Notably, invTrEMO demonstrates superior IGD convergence trends and RMSE results than invTrEMO-w/o-TST. These outcomes can be attributed to the removal of the two step training process, which potentially leads to biased predictions of the invTGPs-w/o-TST towards source tasks. Consequently, negative transfer occurs, resulting in misguided optimization. 
		}
		
		\subsection{Comparison with Forward Transfer}
		{
			\begin{figure*}[!h]
				\begin{center}
					\subfigure[]{\label{extremo_dtlz1}\includegraphics[width=0.24\columnwidth]{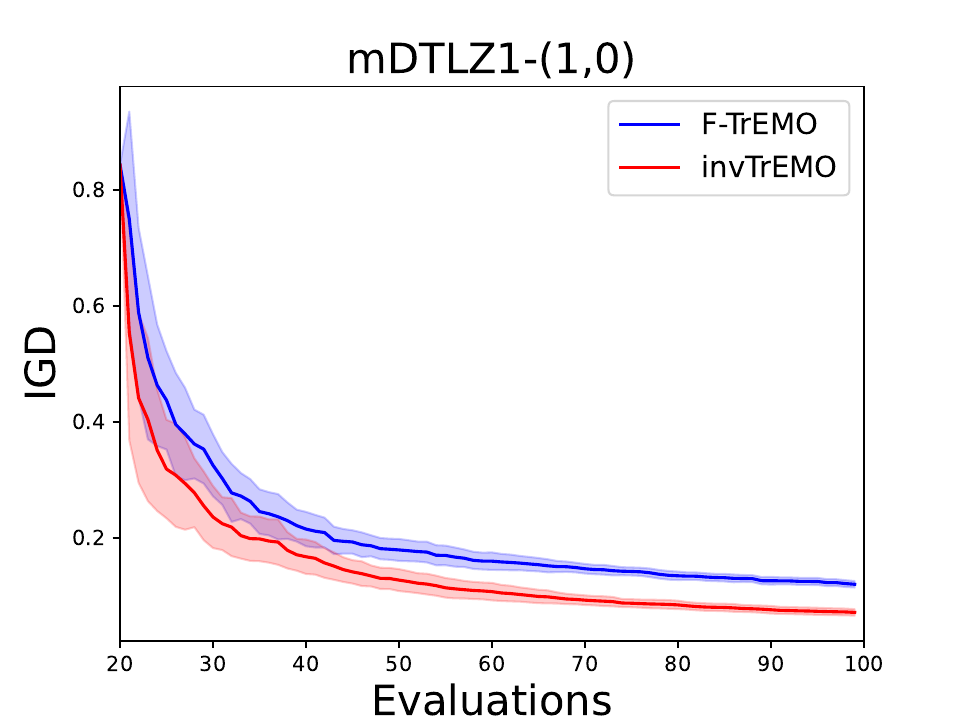}}
					\subfigure[]{\label{extremo_dtlz2}\includegraphics[width=0.24\columnwidth]{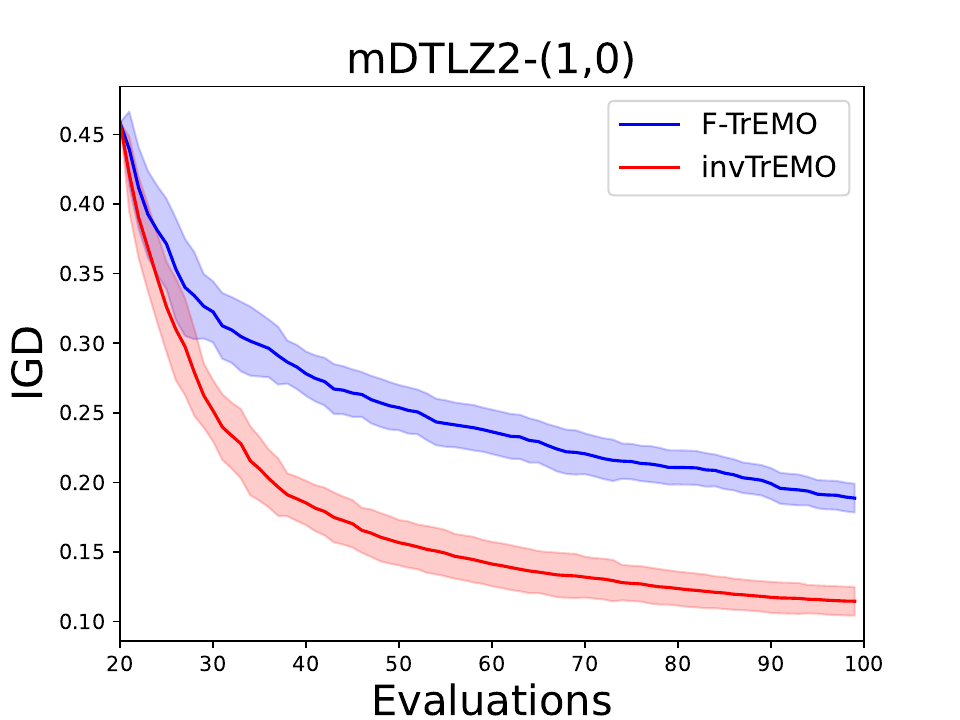}}
					\subfigure[]{\label{extremo_dtlz3}\includegraphics[width=0.24\columnwidth]{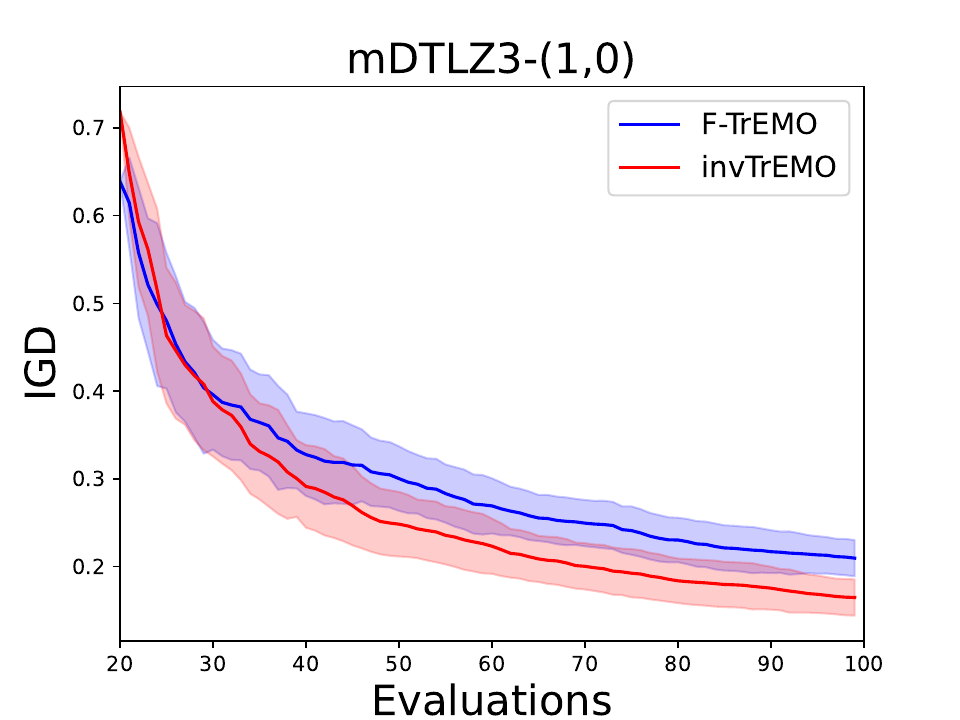}}
					\subfigure[]{\label{extremo_dtlz4}\includegraphics[width=0.24\columnwidth]{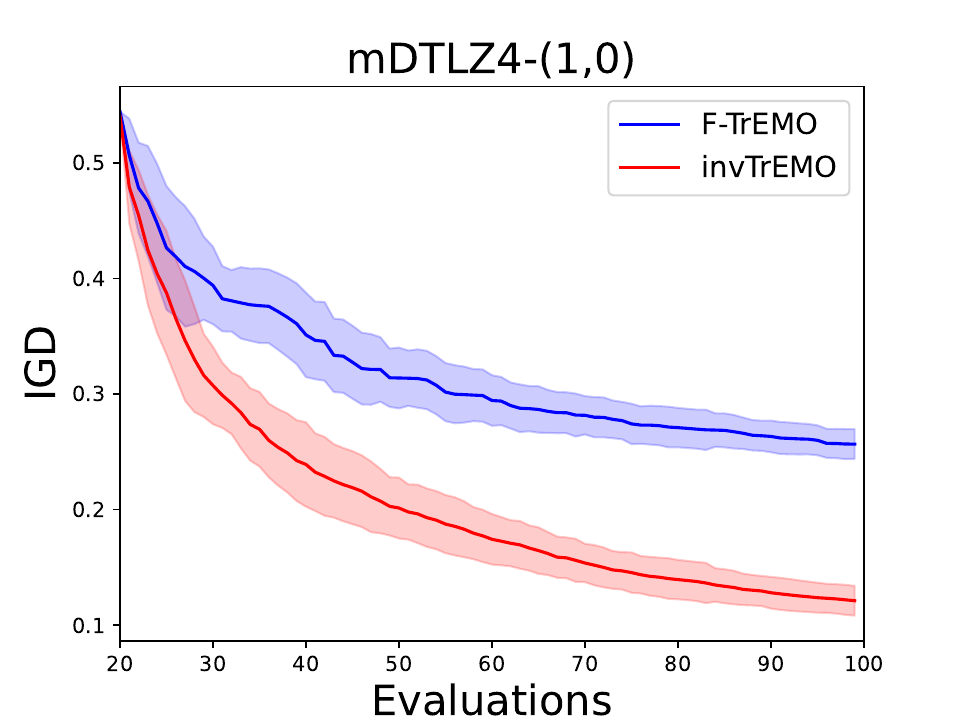}}
					\caption{Comparison of IGD convergence trends averaged over 20 independent runs of F-TrEMO and invTrEMO. Shaded areas represent one standard deviation on either side of the mean. (a) mDTLZ1-($1,0$). (b) mDTLZ2-($1,0$). (c) mDTLZ3-($1,0$). (d) mDTLZ4-($1,0$).}\label{Fig:compare_extremo}
				\end{center}
			\end{figure*}
			
			As discussed in Section 1, a prominent characteristic of invTrEMO is its ability to effectively transfer knowledge across heterogeneous source-target pairs. To further validate the effectiveness of our approach in handling such pairs, we conduct a comparative study between invTrEMO and a forward transfer-based algorithm, denoted as F-TrEMO, for solving mDTLZ problems. Unlike invTGPs used in invTrEMO, F-TrEMO utilizes a forward TGP for knowledge transfer. Specifically, the forward TGP approximates the scalarized function by utilizing both the source and target data, guiding optimization by identifying the solution with the maximum UCB value based on its predictions in each iteration. To adapt the forward TGP for heterogeneous decision spaces, we align the dimensions of the decision vectors of the source data with those of the target data by concatenating a set of random numbers within the range of [0,1]. The convergence trends of IGD for invTrEMO and F-TrEMO are illustrated in Fig.~\ref{Fig:compare_extremo}. Notably, invTrEMO exhibits significantly faster convergence compared to F-TrEMO. These results underscore the effectiveness of invTrEMO in facilitating efficient knowledge transfer across heterogeneous scenarios.
		}
		
		\newpage
		
		\subsection{Pareto Front Approximations}
		Figs.~S-1 to ~S-4  shows the Pareto front approximation provided by ParEGO-UCB, MOEA/D-EGO, K-RVEA, CSEA,  PSL-MOBO, qNEHVI, and invTrEMO on mDTLZ-$(1,0)$ benchmark problems after 100 evaluations. It is observed that compared with ParEGO-UCB, MOEA/D-EGO, K-RVEA, CSEA,  PSL-MOBO, qNEHVI, invTrEMO obtains solutions that offer better convergence and spread.
		
		\begin{figure*}[!h]
			\begin{center}
				\subfigure[]{\label{pareto17}\includegraphics[width=0.3\columnwidth]{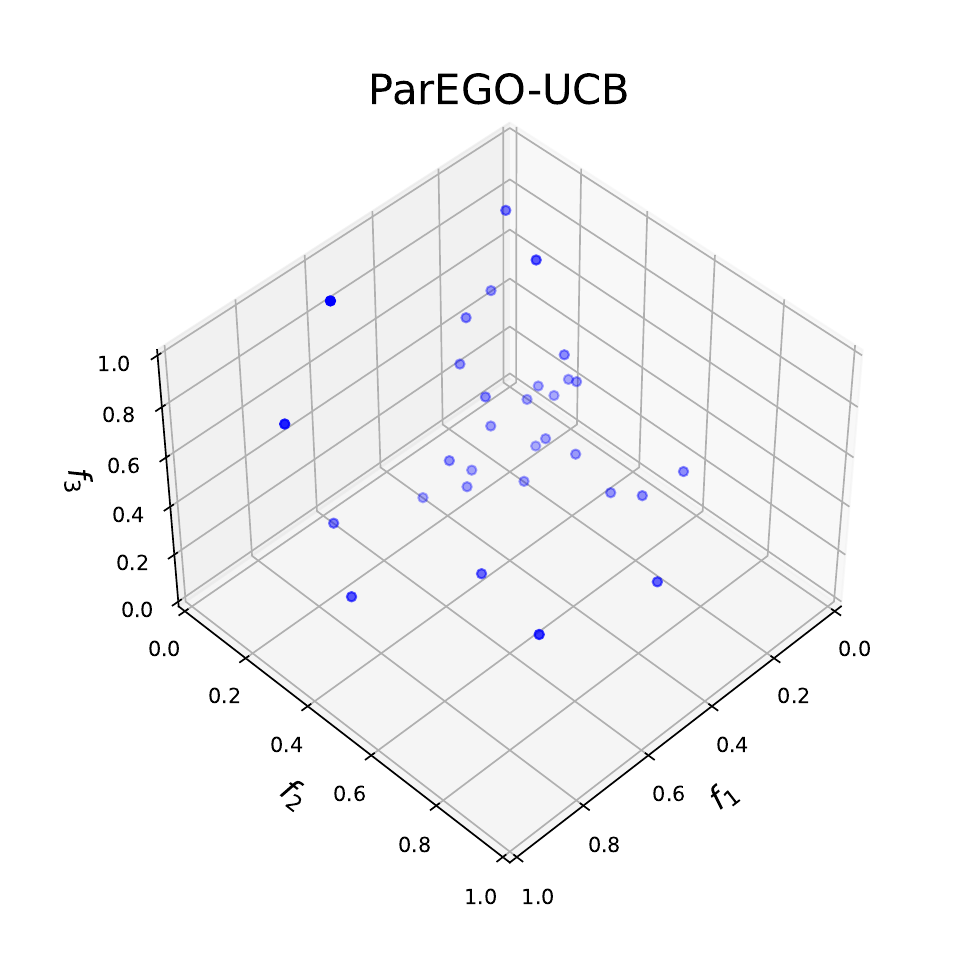}}
				\subfigure[]{\label{pareto16}\includegraphics[width=0.3\columnwidth]{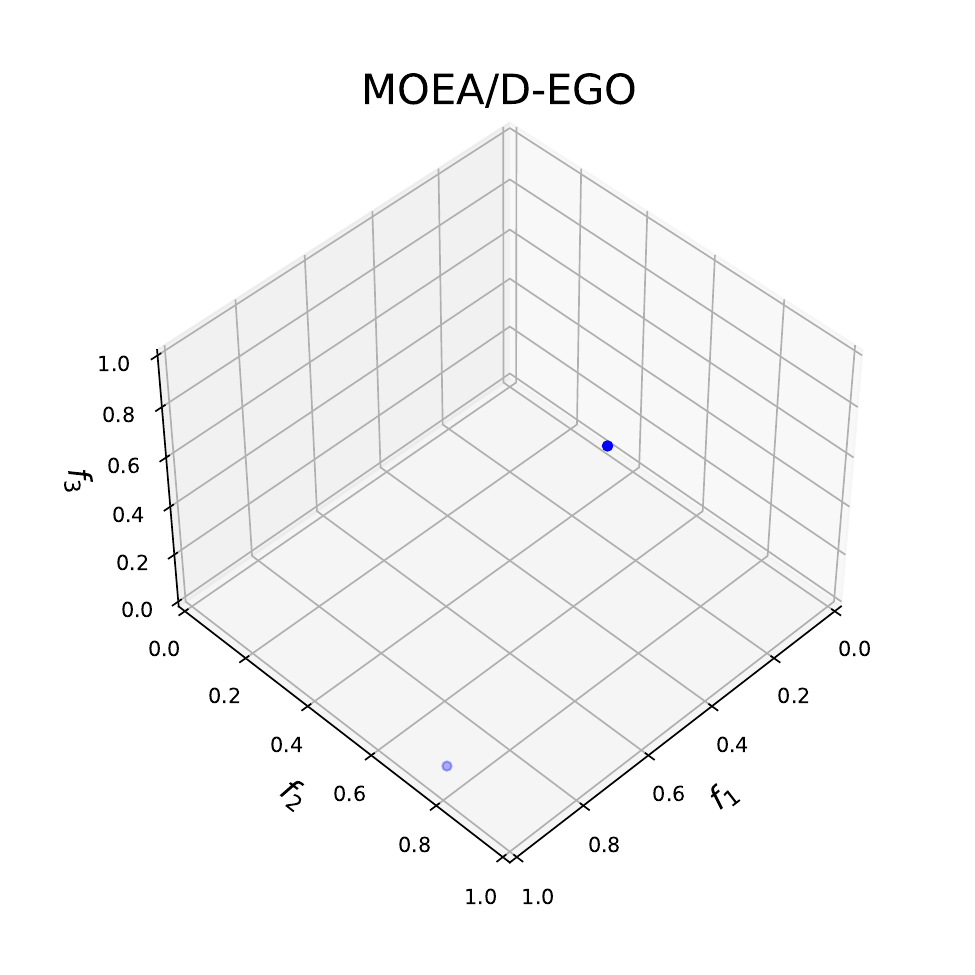}}
				\subfigure[]{\label{pareto16}\includegraphics[width=0.3\columnwidth]{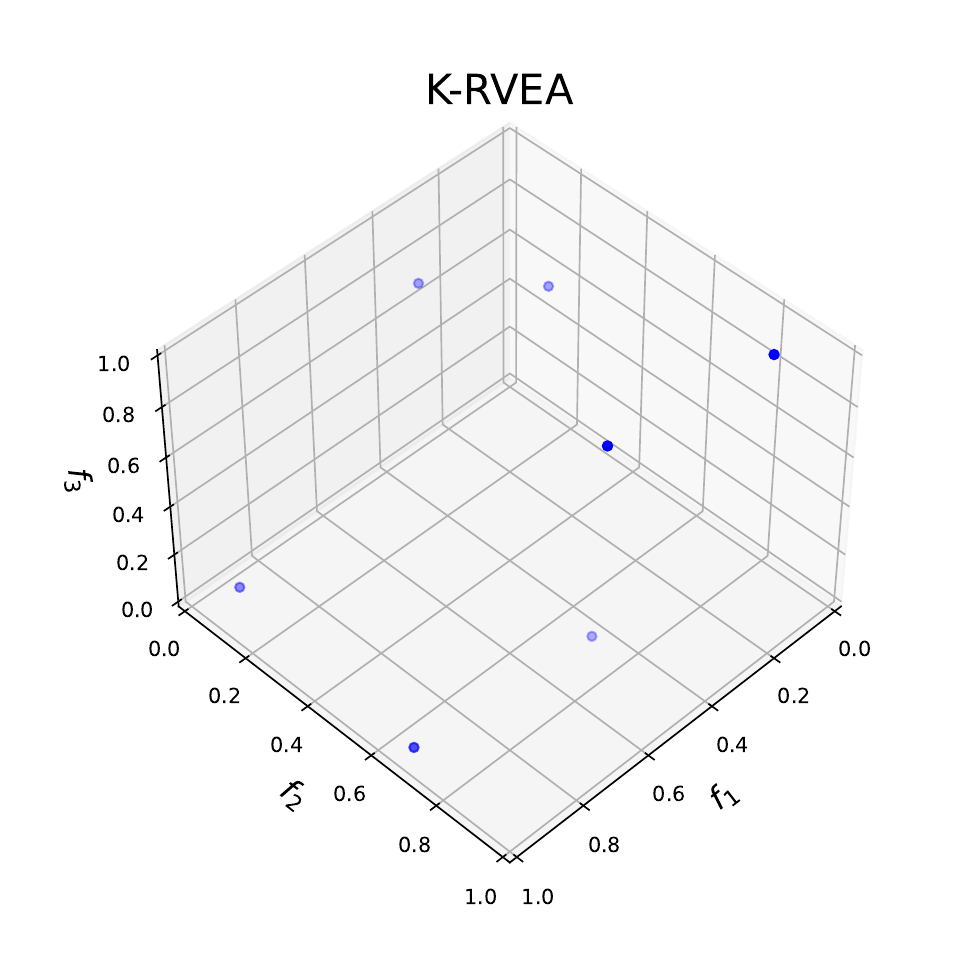}}
				\subfigure[]{\label{pareto15}\includegraphics[width=0.3\columnwidth]{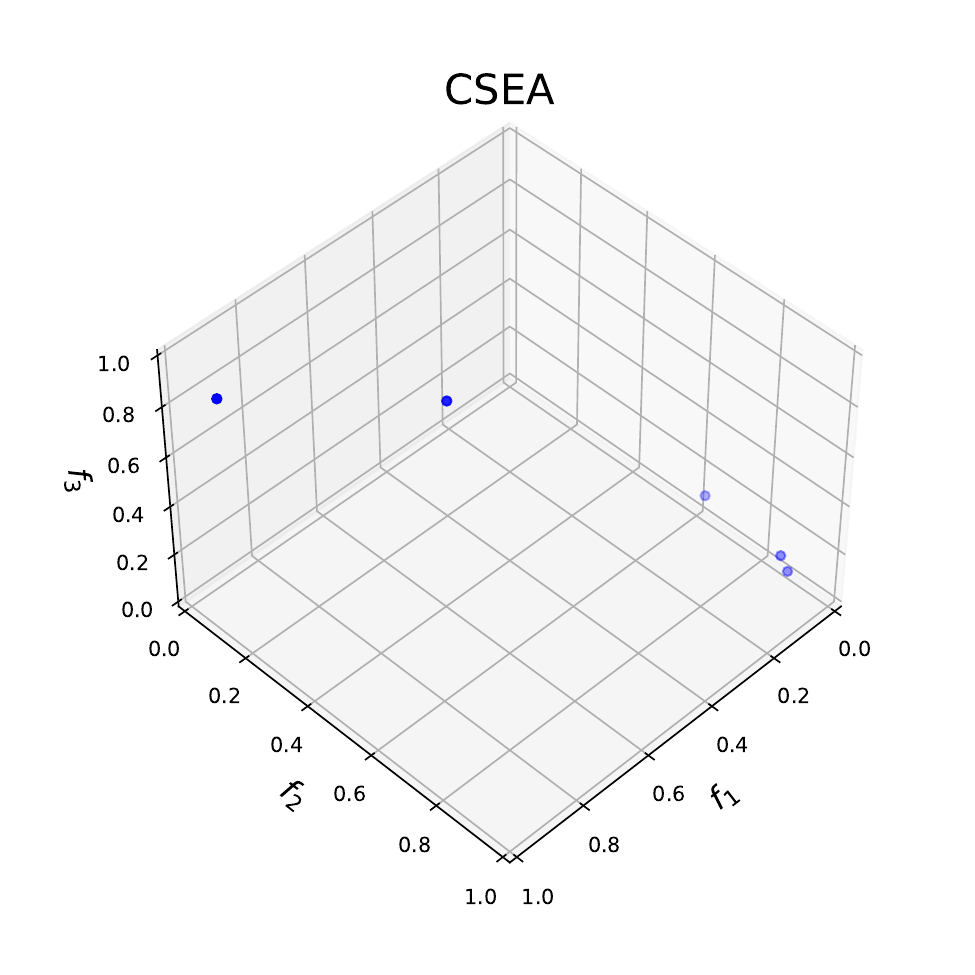}}
				\subfigure[]{\label{pareto17}\includegraphics[width=0.3\columnwidth]{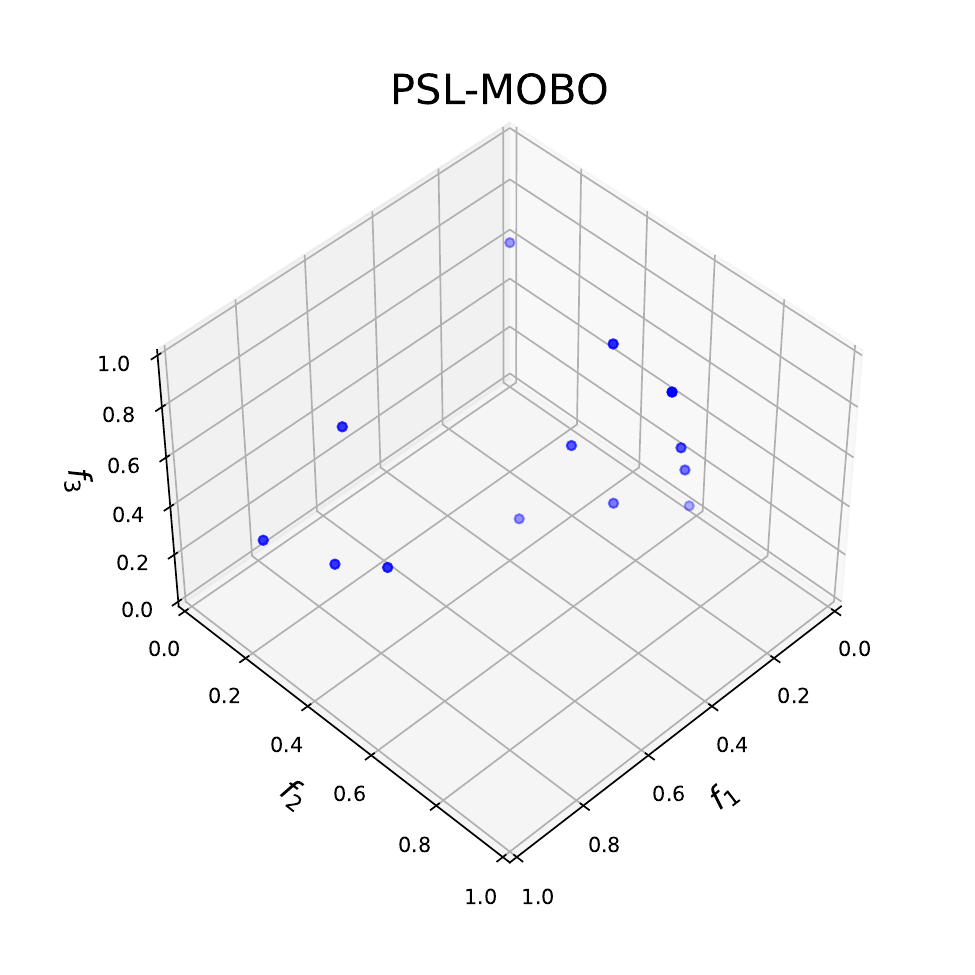}}
				\subfigure[]{\label{pareto17}\includegraphics[width=0.3\columnwidth]{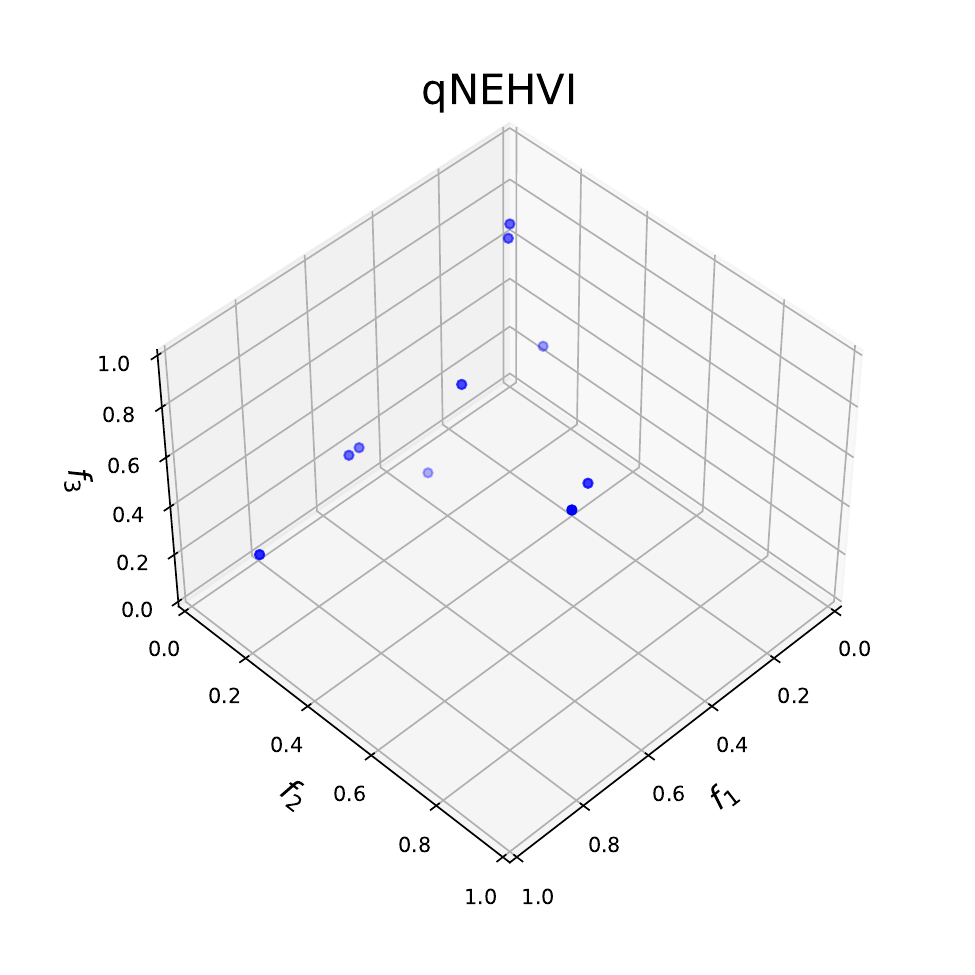}}
				\subfigure[]{\label{pareto11}\includegraphics[width=0.3\columnwidth]{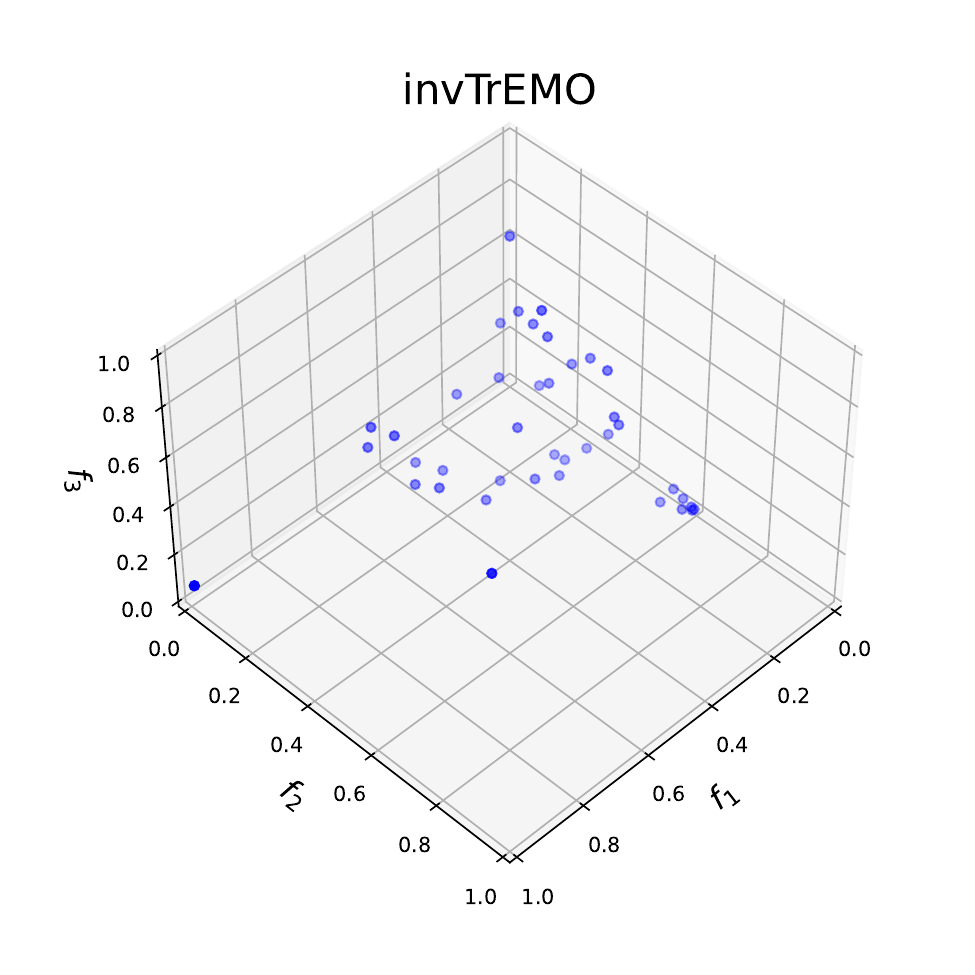}}
				\caption{Pareto front approximation by ParEGO-UCB, MOEA/D-EGO, K-RVEA, CSEA, qNEHVI, PSL-MOBO, and invTrEMO on mDTLZ1-($1,0$) over 100 evaluations. Note that, only the solutions with the objective function values in the region $[0,1] \times [0,1] \times [0,1]$ are shown in the figure. (a) ParEGO-UCB. (b) MOEA/D-EGO. (c) K-RVEA. (d) CSEA. (e) qNEHVI. (f) PSL-MOBO. (g) invTrEMO.}\label{Fig:PF}
			\end{center}
		\end{figure*}
		
		\begin{figure*}[!h]
			\begin{center}
				\subfigure[]{\label{pareto17}\includegraphics[width=0.3\columnwidth]{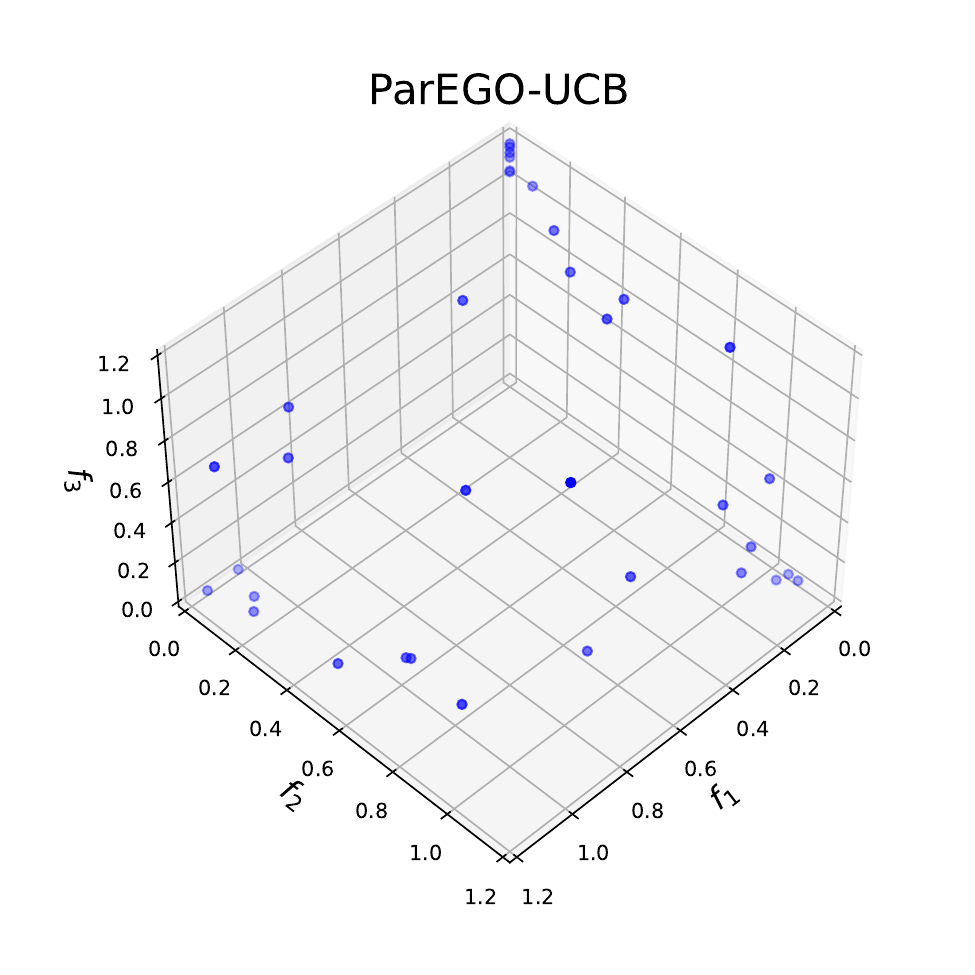}}
				\subfigure[]{\label{pareto16}\includegraphics[width=0.3\columnwidth]{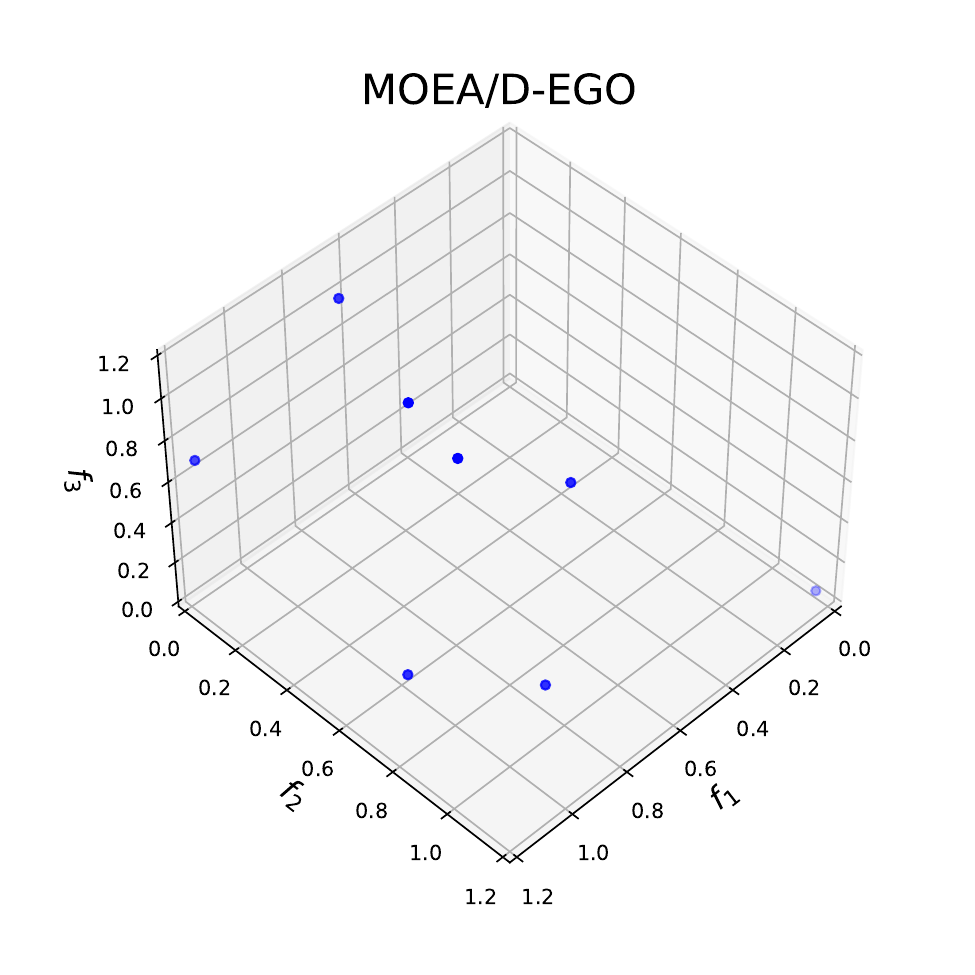}}
				\subfigure[]{\label{pareto16}\includegraphics[width=0.3\columnwidth]{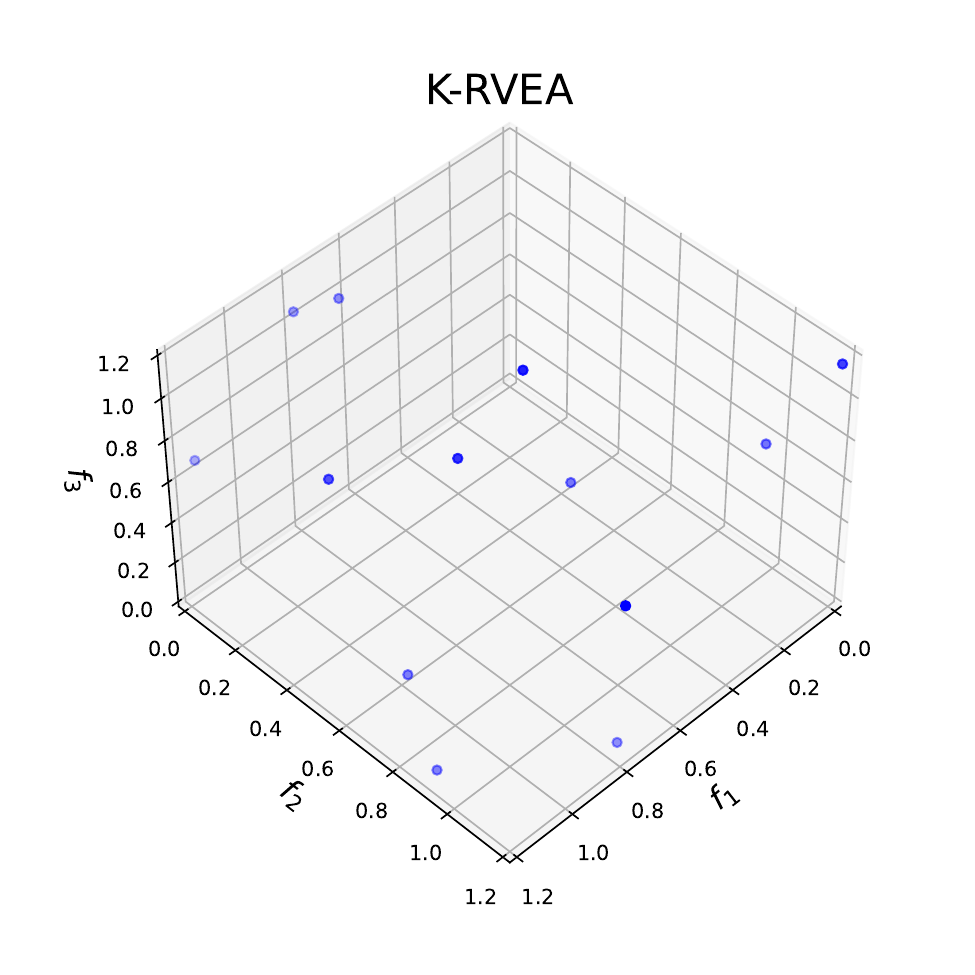}}
				\subfigure[]{\label{pareto15}\includegraphics[width=0.3\columnwidth]{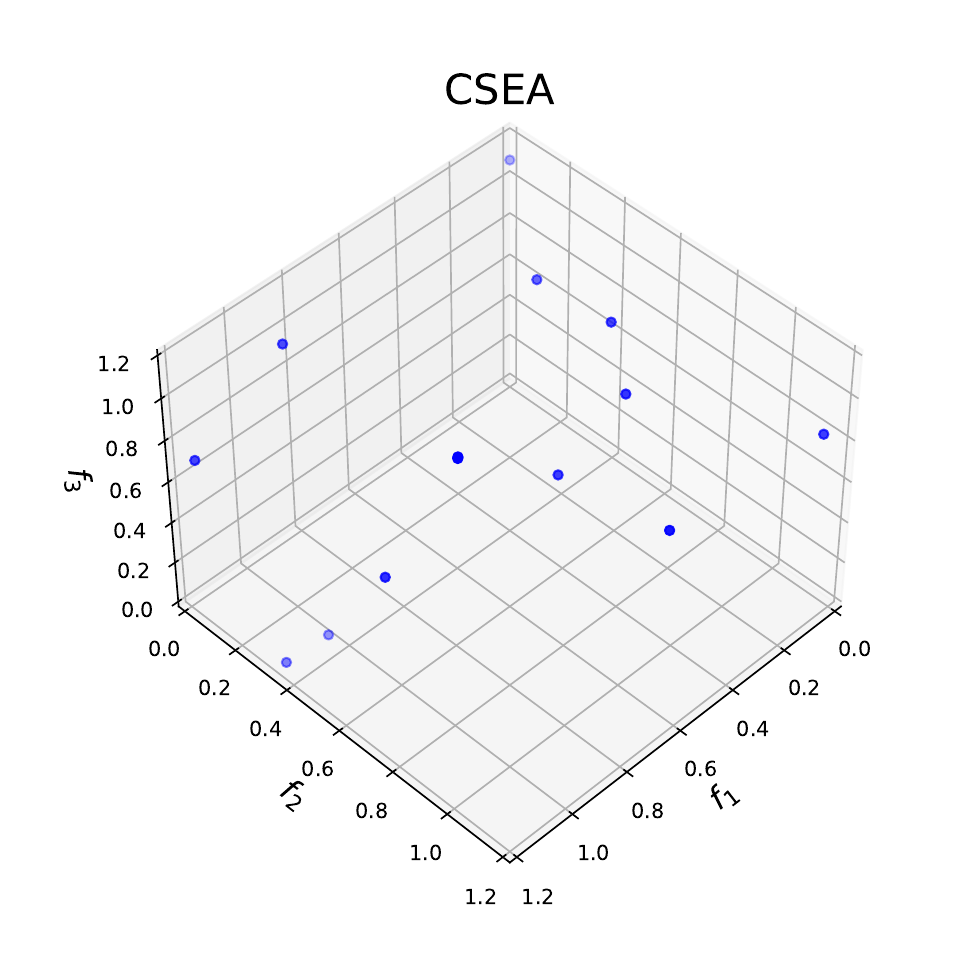}}    
				\subfigure[]{\label{pareto17}\includegraphics[width=0.3\columnwidth]{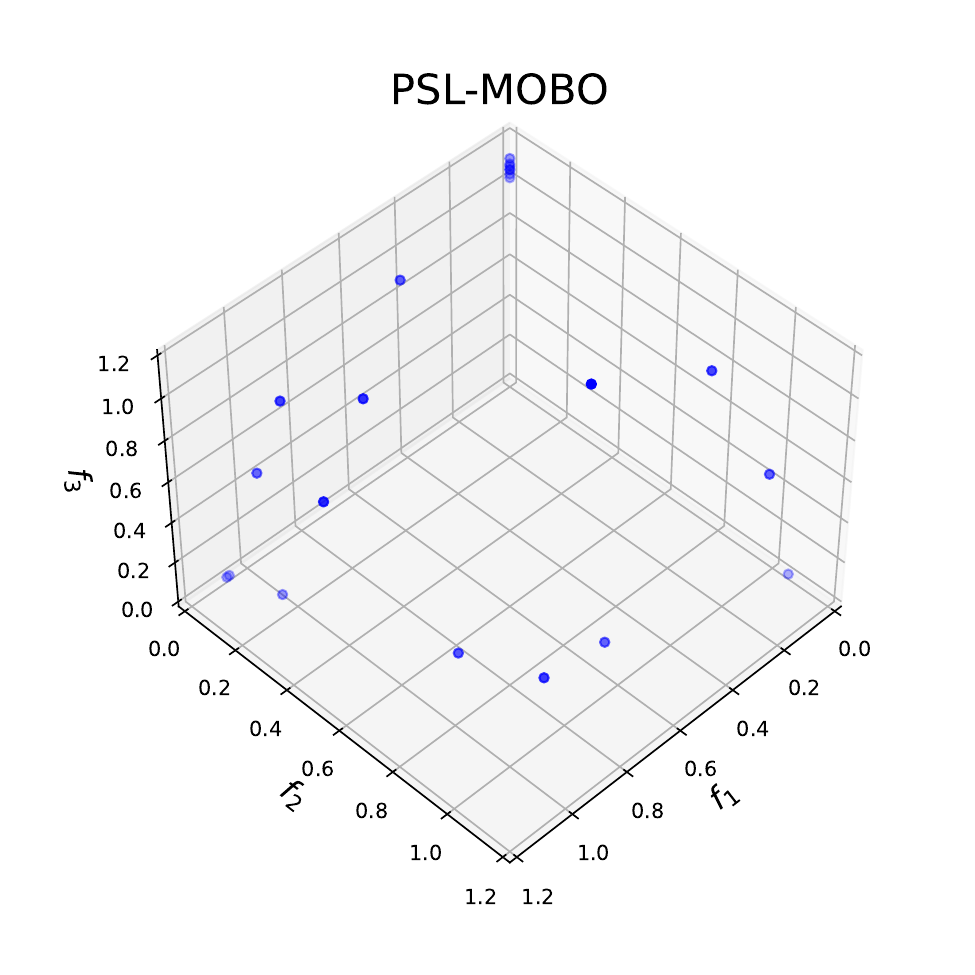}}
				\subfigure[]{\label{pareto17}\includegraphics[width=0.3\columnwidth]{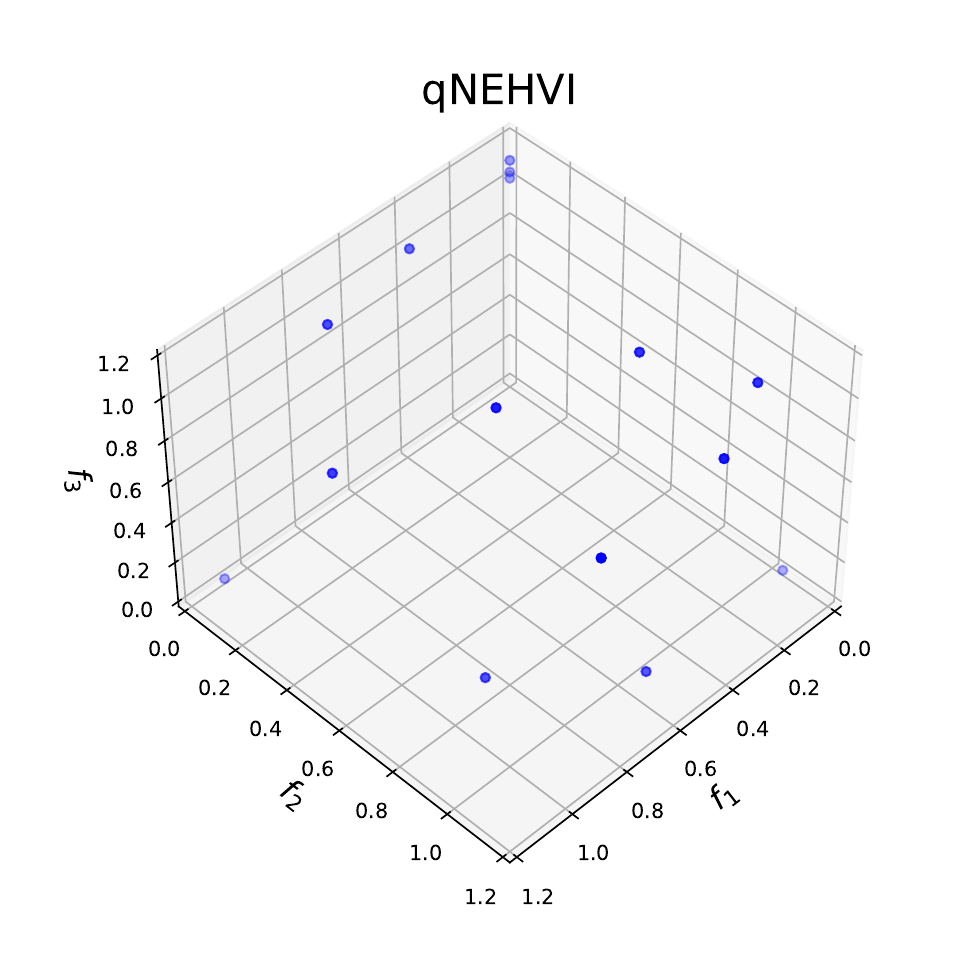}}
				\subfigure[]{\label{pareto11}\includegraphics[width=0.3\columnwidth]{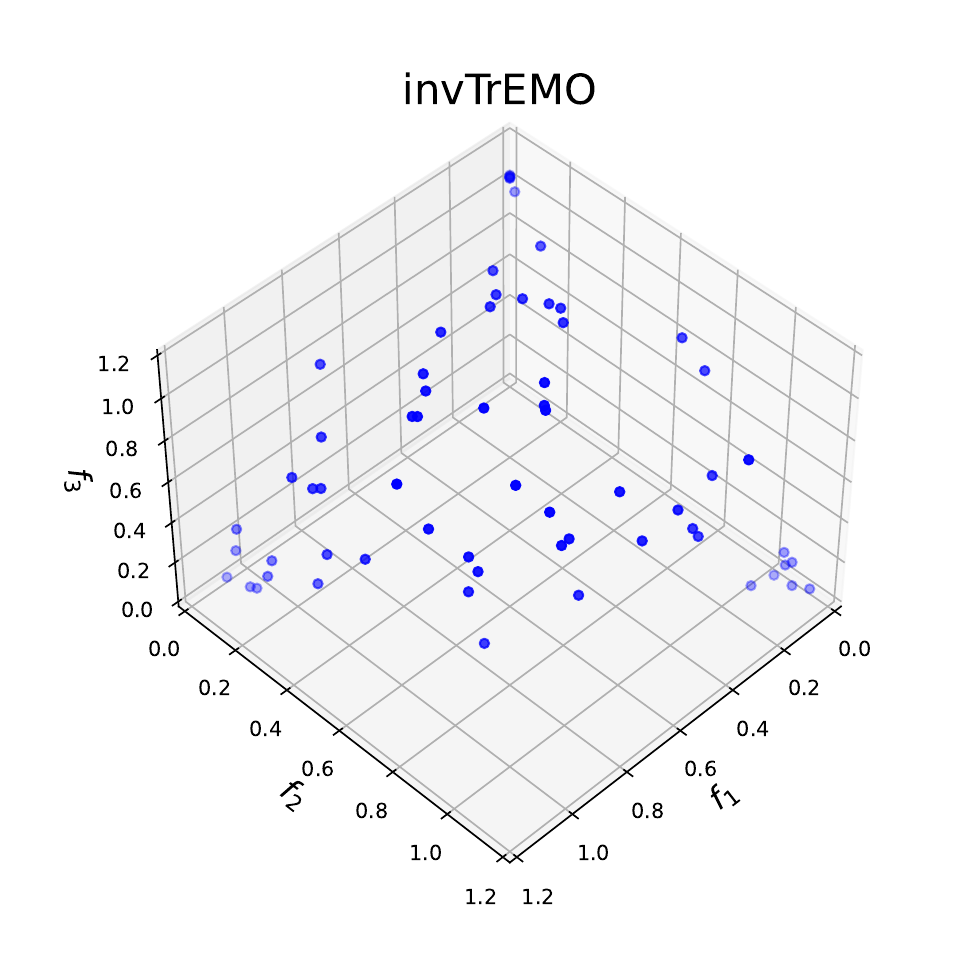}}
				\caption{Pareto front approximation by ParEGO-UCB, MOEA/D-EGO, K-RVEA, CSEA, qNEHVI, PSL-MOBO, and invTrEMO on mDTLZ2-($1,0$) over 100 evaluations. Note that, only the solutions with the objective function values in the region $[0,1.2] \times [0,1.2] \times [0,1.2]$ are shown in the figure. (a) ParEGO-UCB. (b) MOEA/D-EGO. (c) K-RVEA. (d) CSEA. (e) qNEHVI. (f) PSL-MOBO. (g) invTrEMO.}\label{Fig:PF}
			\end{center}
		\end{figure*}
		
		\begin{figure*}[!h]
			\begin{center}
				\subfigure[]{\label{pareto17}\includegraphics[width=0.3\columnwidth]{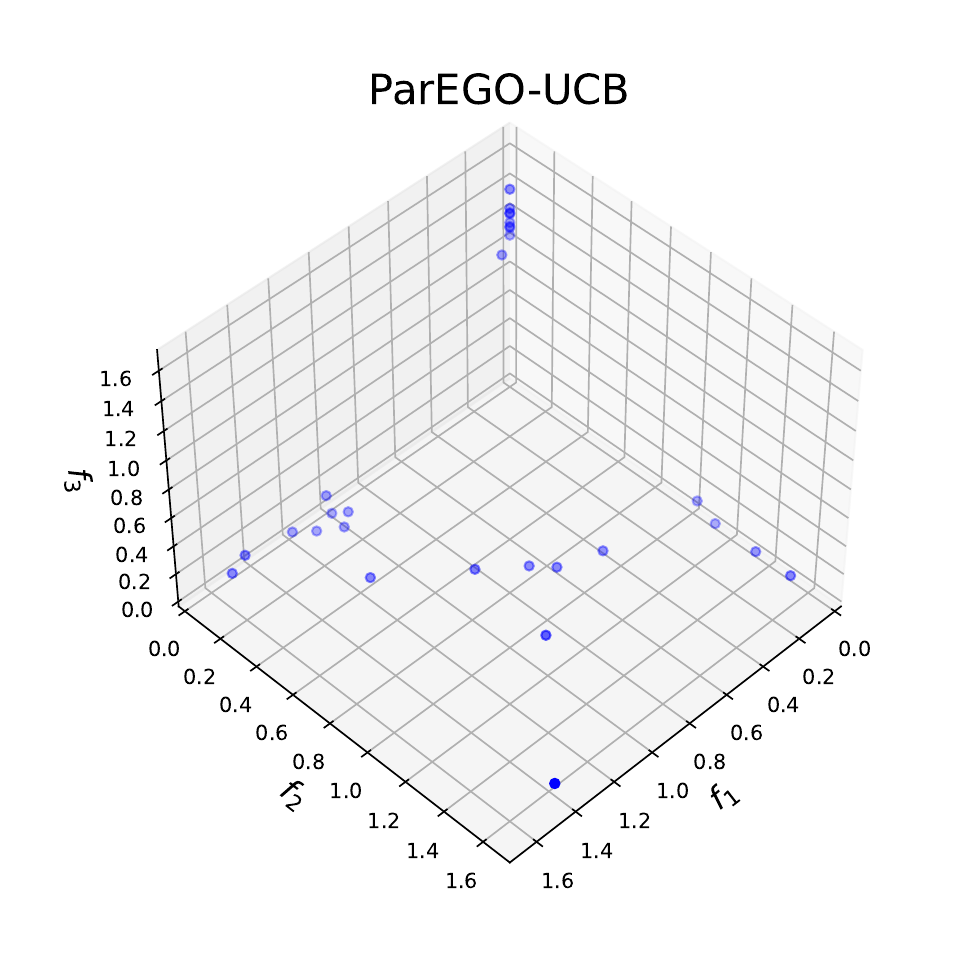}}
				\subfigure[]{\label{pareto16}\includegraphics[width=0.3\columnwidth]{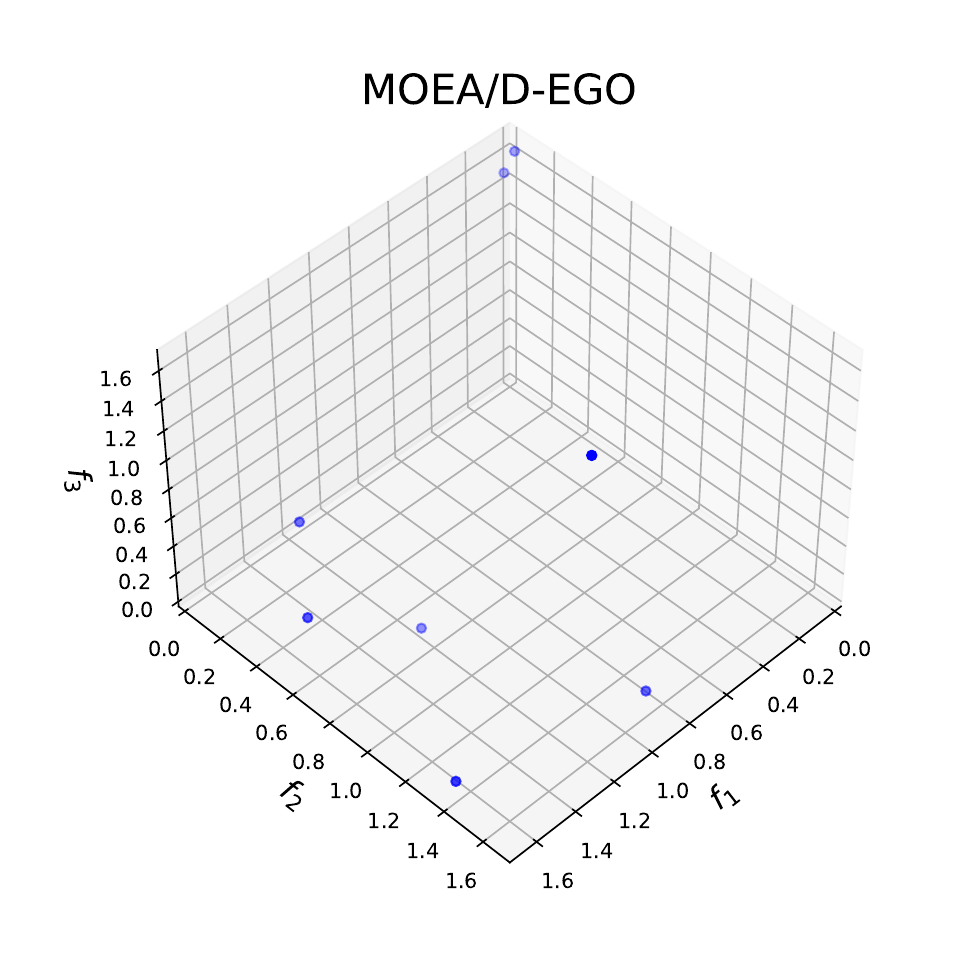}}
				\subfigure[]{\label{pareto16}\includegraphics[width=0.3\columnwidth]{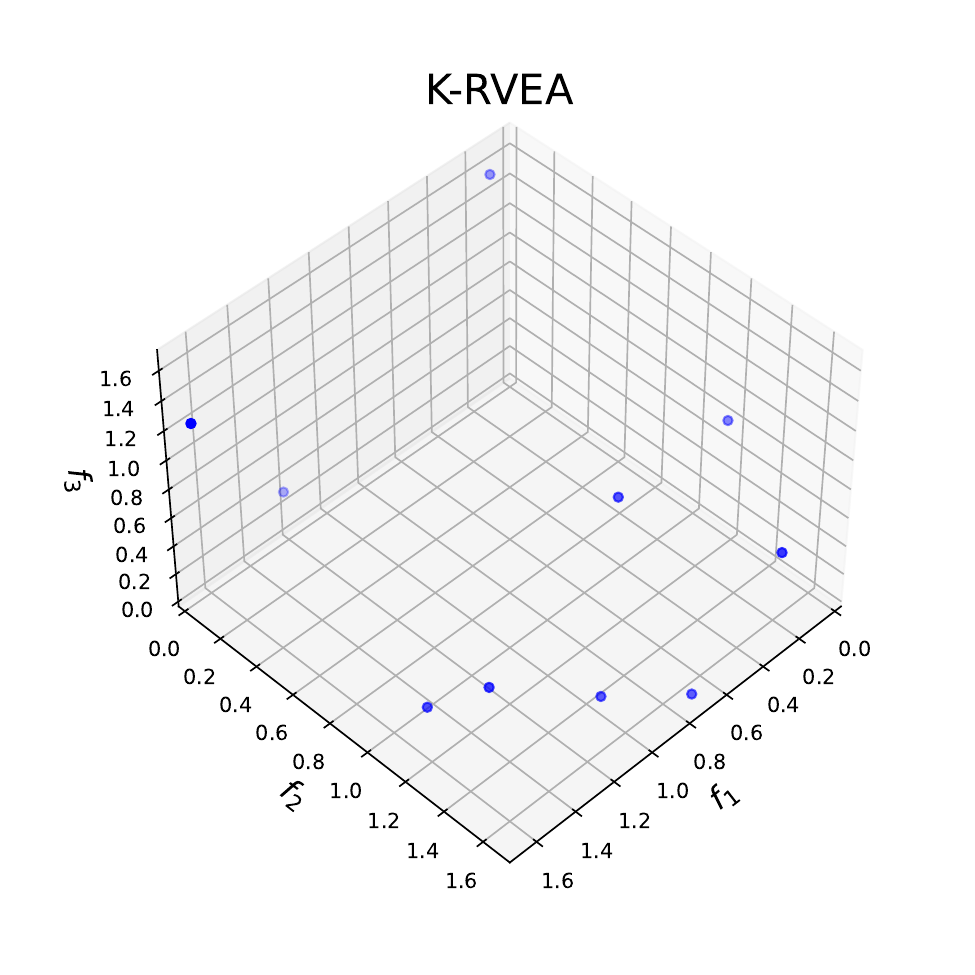}}
				\subfigure[]{\label{pareto15}\includegraphics[width=0.3\columnwidth]{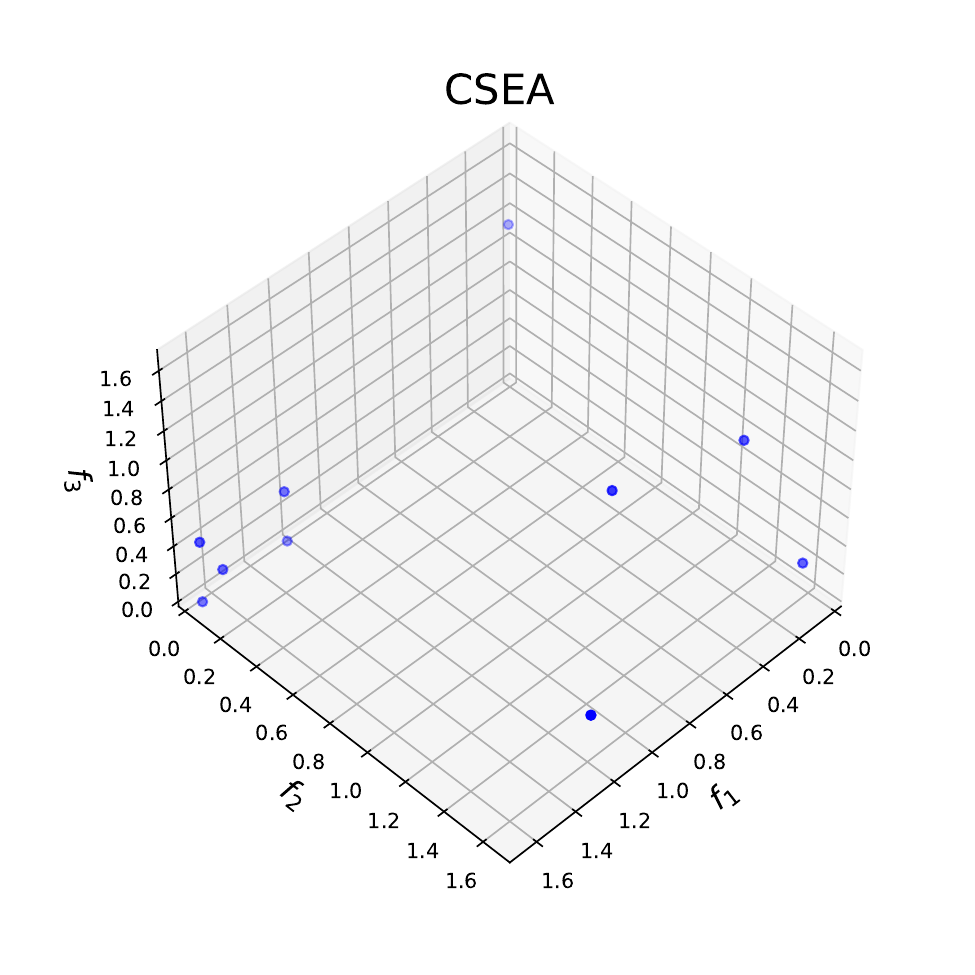}}
				\subfigure[]{\label{pareto17}\includegraphics[width=0.3\columnwidth]{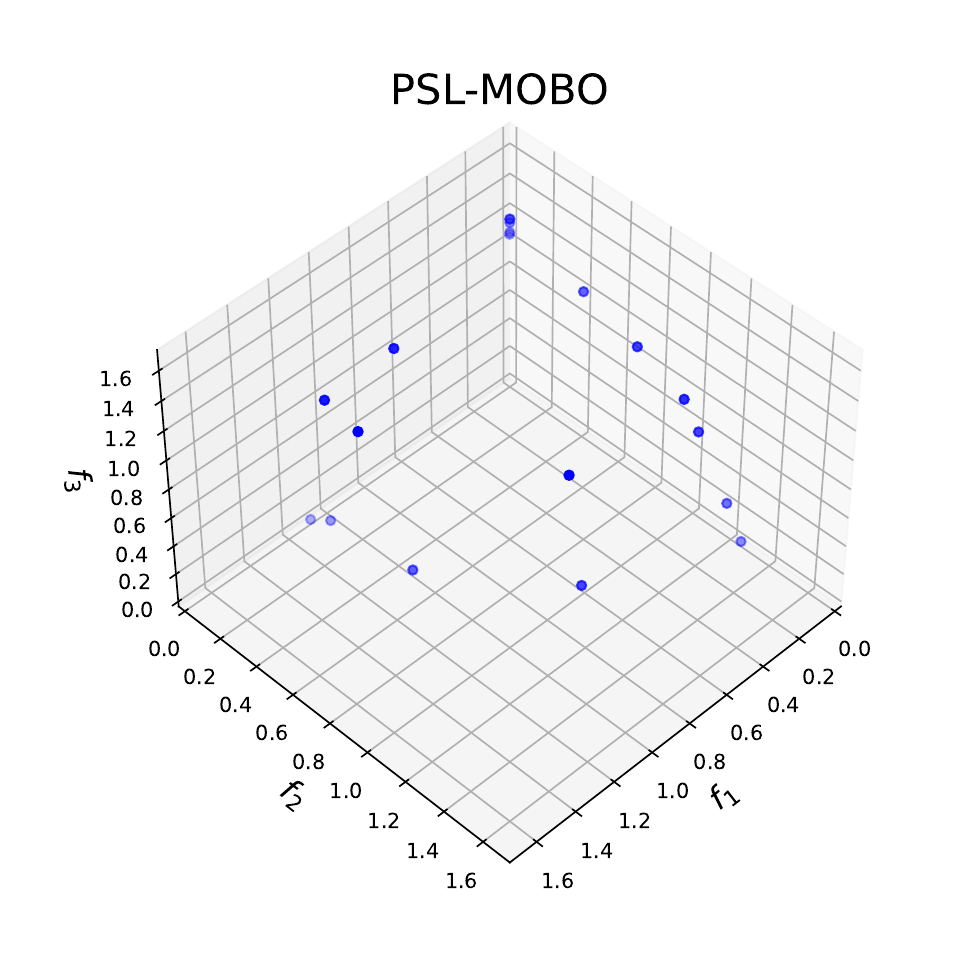}}
				\subfigure[]{\label{pareto17}\includegraphics[width=0.3\columnwidth]{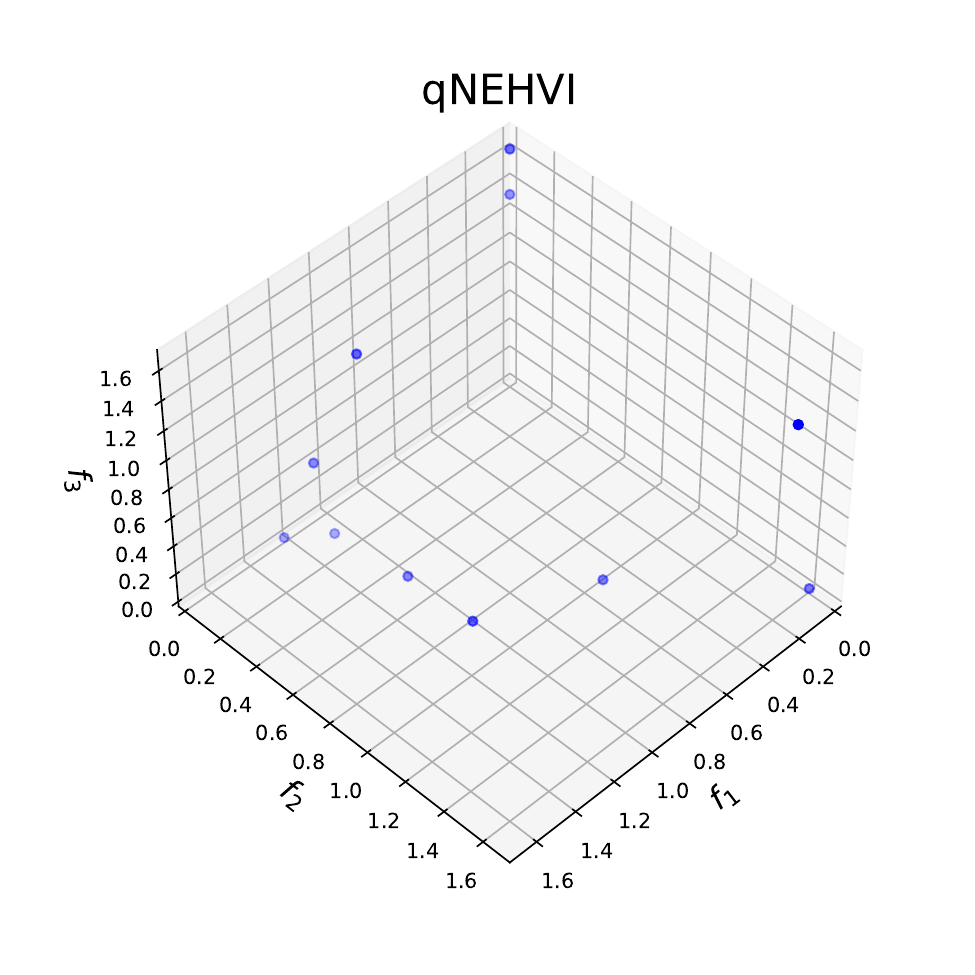}}
				\subfigure[]{\label{pareto11}\includegraphics[width=0.3\columnwidth]{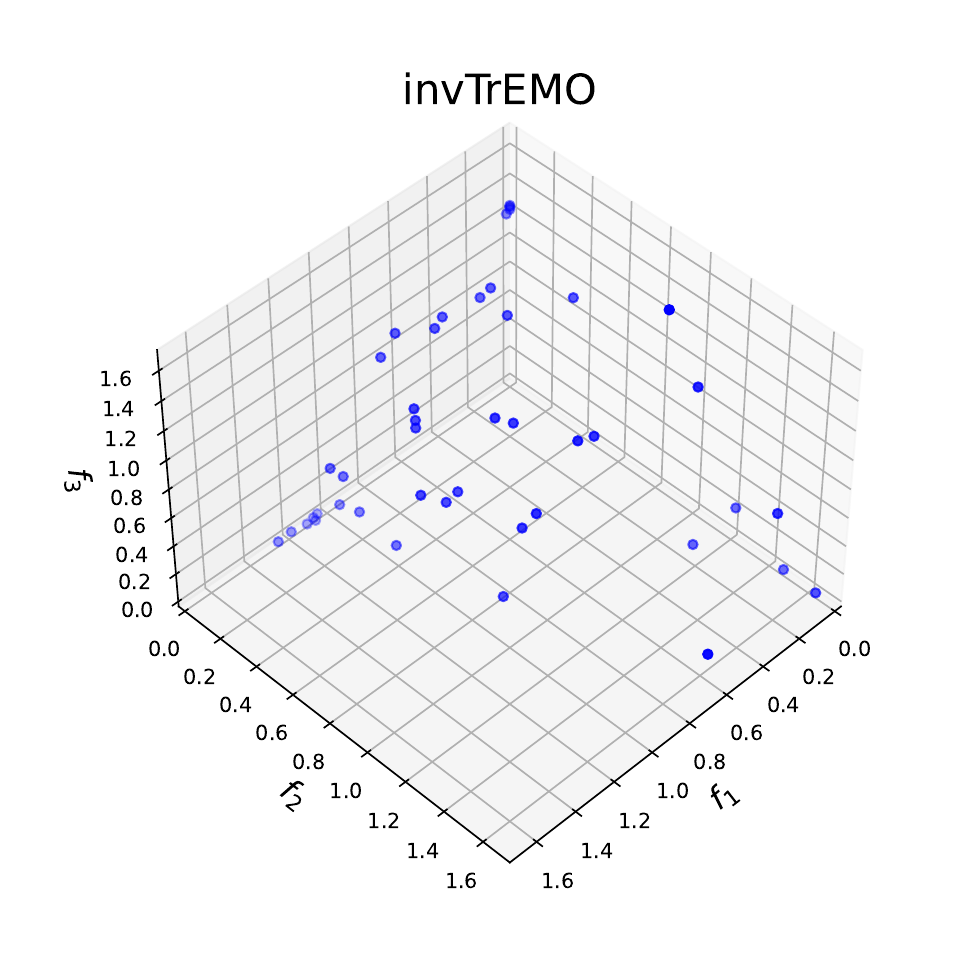}}
				\caption{Pareto front approximation by ParEGO-UCB, MOEA/D-EGO, K-RVEA, CSEA, qNEHVI, PSL-MOBO, and invTrEMO on mDTLZ3-($1,0$) over 100 evaluations. Note that, only the solutions with the objective function values in the region $[0,1.7] \times [0,1.7] \times [0,1.7]$ are shown in the figure. (a) ParEGO-UCB. (b) MOEA/D-EGO. (c) K-RVEA. (d) CSEA. (e) qNEHVI. (f) PSL-MOBO. (g) invTrEMO.}\label{Fig:PF}
			\end{center}
		\end{figure*}
		
		\begin{figure*}[!h]
			\begin{center}
				\subfigure[]{\label{pareto17}\includegraphics[width=0.3\columnwidth]{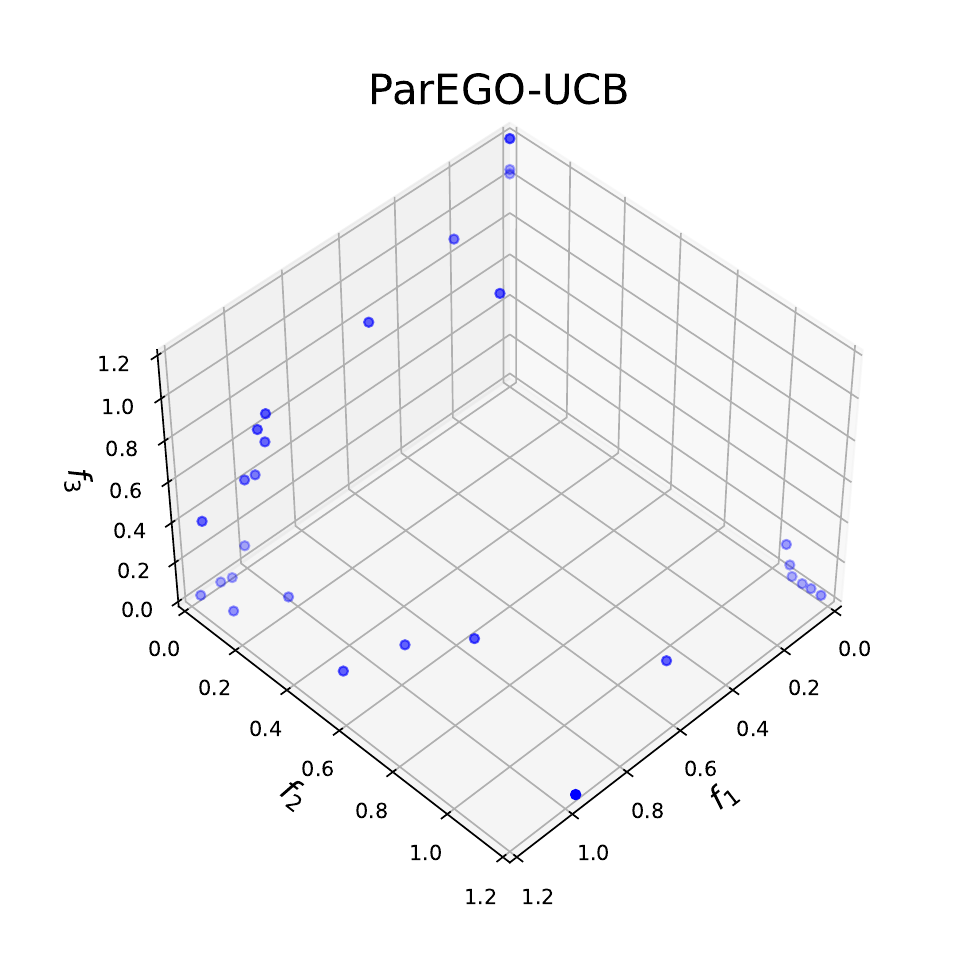}}
				\subfigure[]{\label{pareto16}\includegraphics[width=0.3\columnwidth]{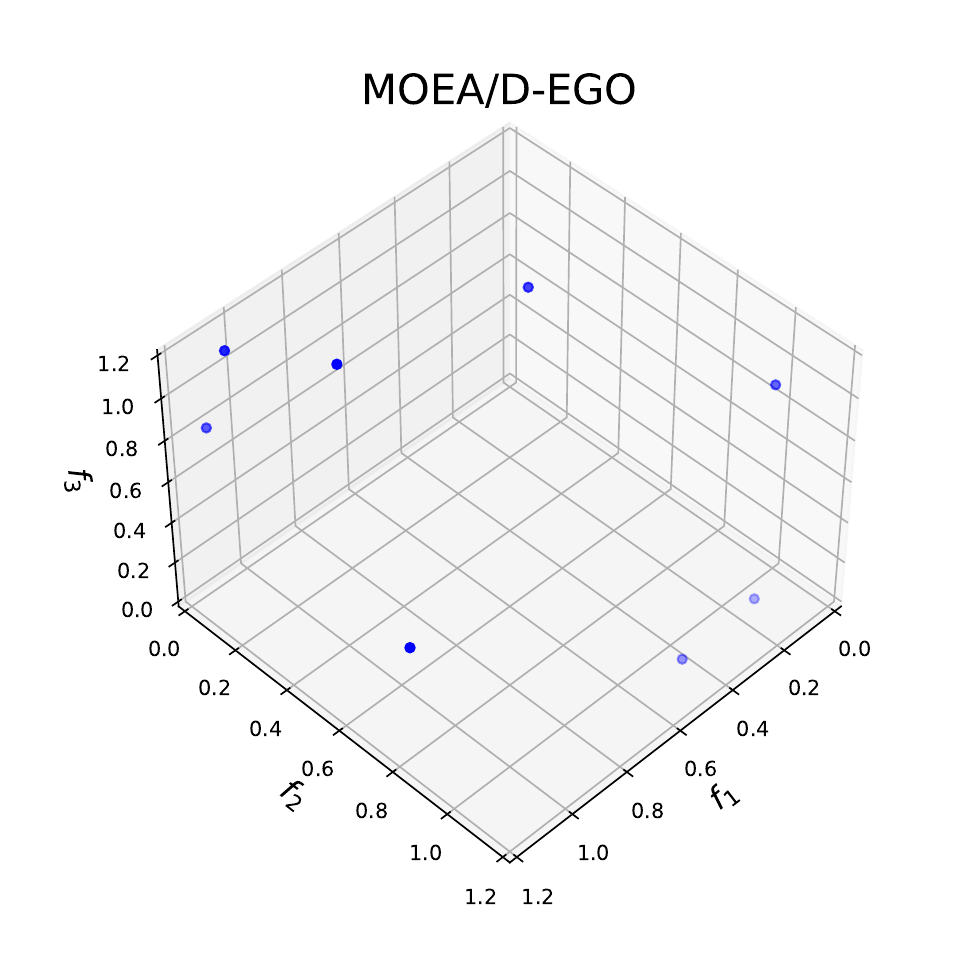}}
				\subfigure[]{\label{pareto16}\includegraphics[width=0.3\columnwidth]{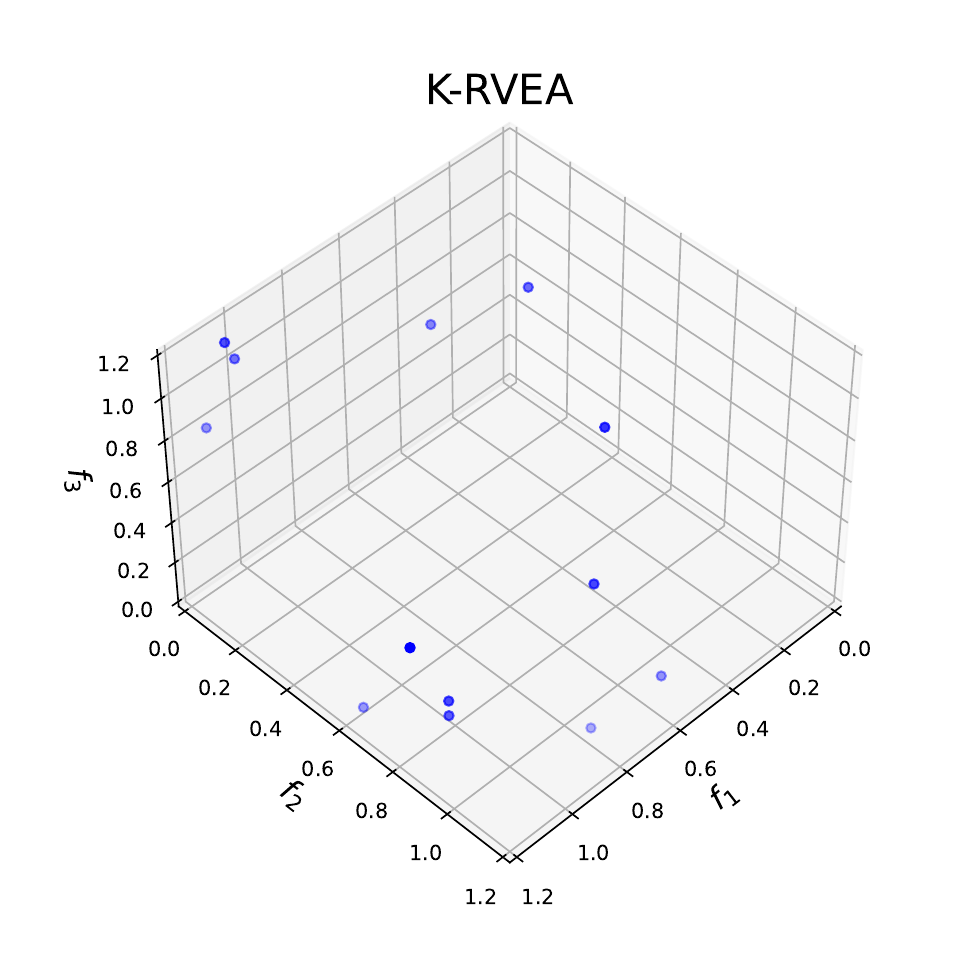}}
				\subfigure[]{\label{pareto15}\includegraphics[width=0.3\columnwidth]{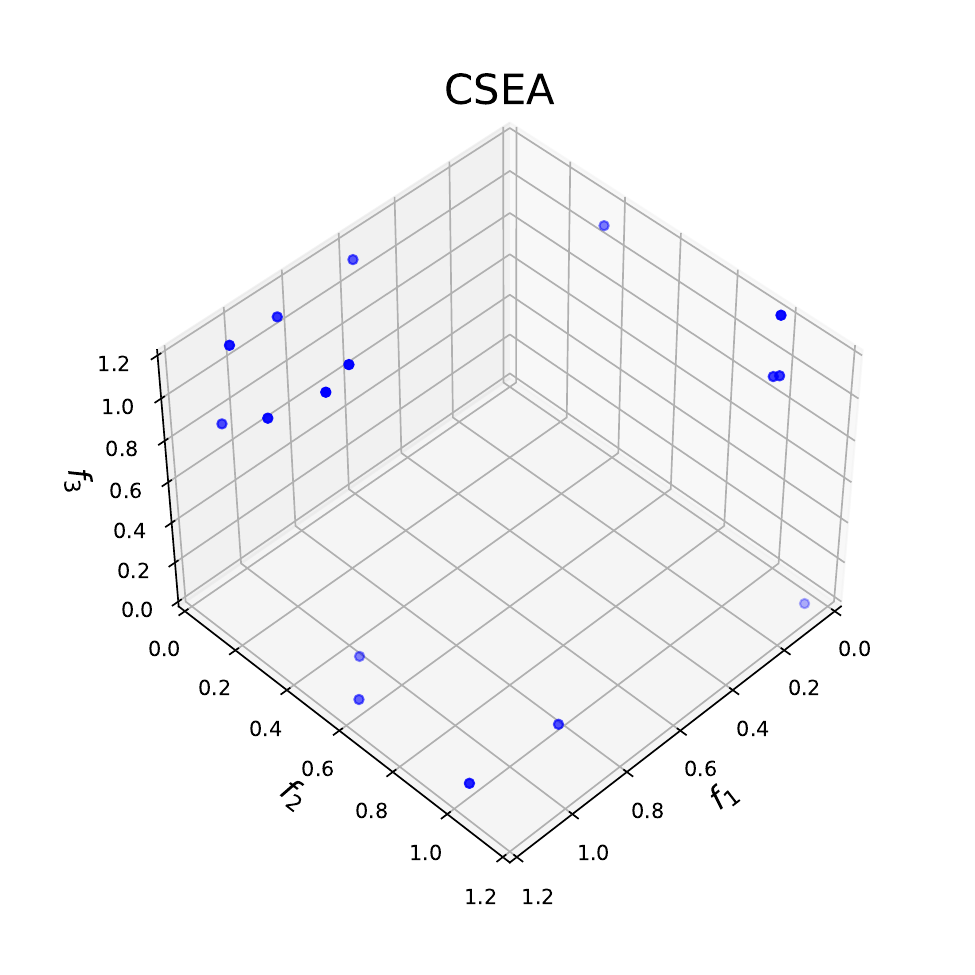}}
				\subfigure[]{\label{pareto17}\includegraphics[width=0.3\columnwidth]{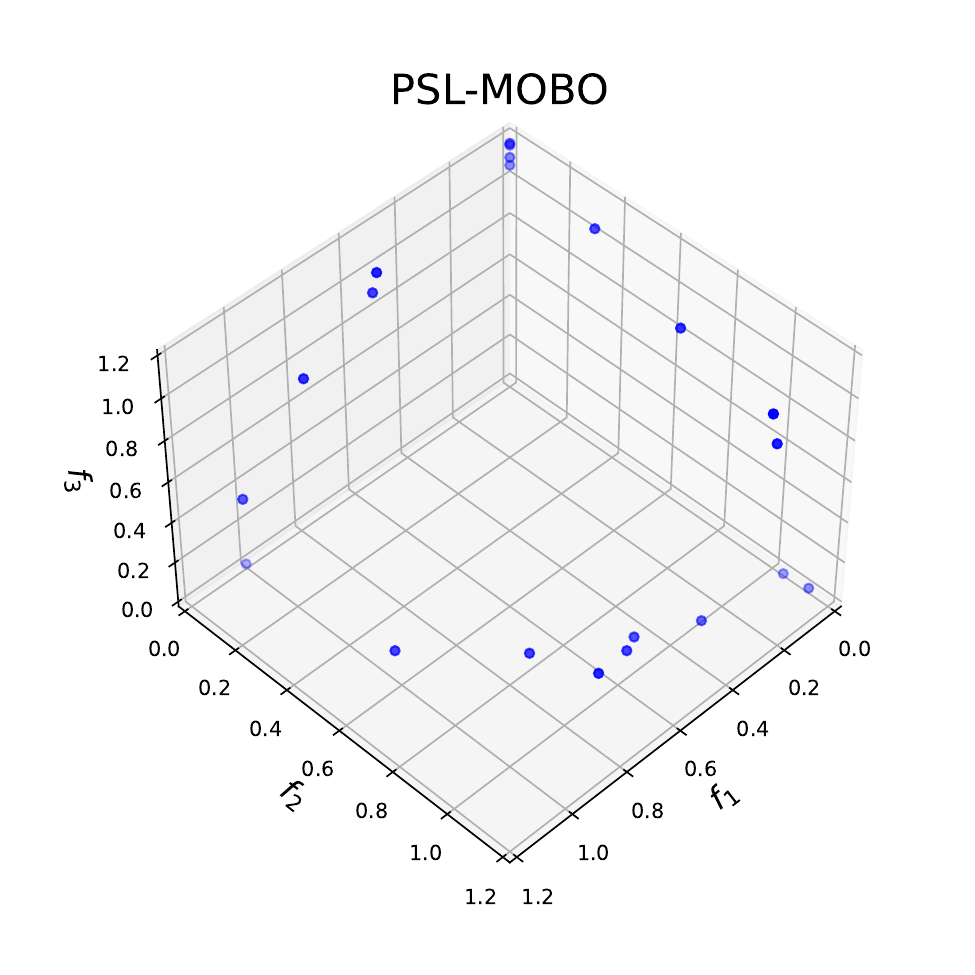}}
				\subfigure[]{\label{pareto17}\includegraphics[width=0.3\columnwidth]{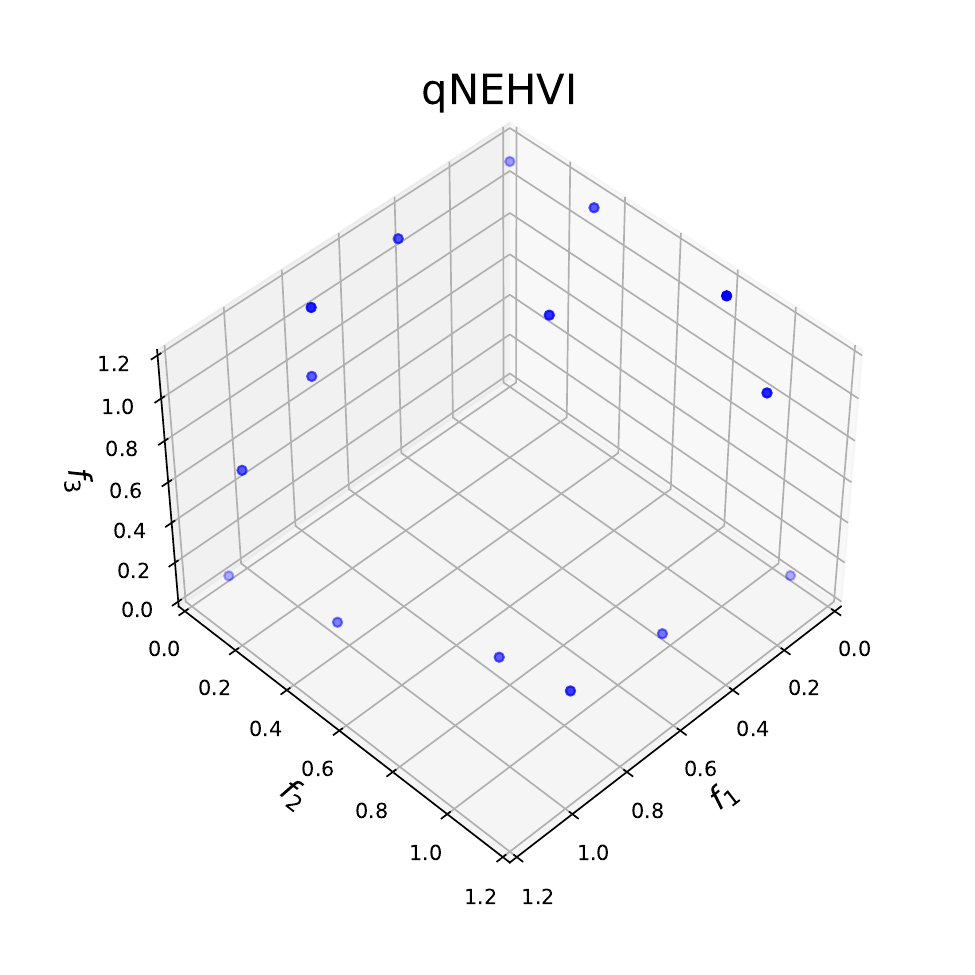}}
				\subfigure[]{\label{pareto11}\includegraphics[width=0.3\columnwidth]{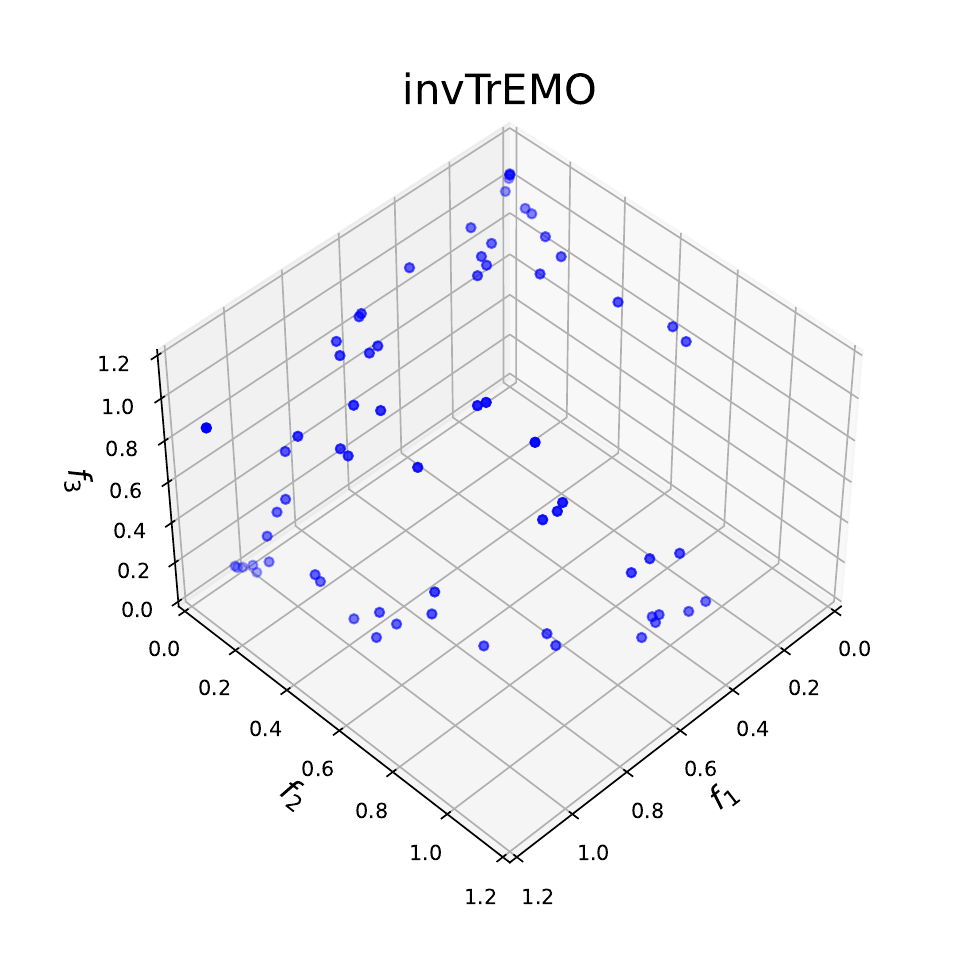}}
				\caption{Pareto front approximation by ParEGO-UCB, MOEA/D-EGO, K-RVEA, CSEA, qNEHVI, PSL-MOBO, and invTrEMO on mDTLZ4-($1,0$) over 100 evaluations. Note that, only the solutions with the objective function values in the region $[0,1.2] \times [0,1.2] \times [0,1.2]$ are shown in the figure. (a) ParEGO-UCB. (b) MOEA/D-EGO. (c) K-RVEA. (d) CSEA. (e) qNEHVI. (f) PSL-MOBO. (g) invTrEMO.}\label{Fig:PF}
			\end{center}
		\end{figure*}
		
		\clearpage
		
		
		
	}
	
\end{document}